%% file: main.tex
\documentclass[10pt,twocolumn,letterpaper]{article}

\usepackage[pagenumbers]{cvpr} 


\input{_constant}
\usepackage{graphicx}

\usepackage{multirow}
\usepackage{bm}

\definecolor{cvprblue}{rgb}{0.21,0.49,0.74}
\usepackage[pagebackref,breaklinks,colorlinks,allcolors=cvprblue]{hyperref}

\def\paperID{13706} 
\def\confName{CVPR}
\def\confYear{2025}

\title{Taming Feed-forward Reconstruction Models as Latent Encoders for \\3D Generative Models}

\author{Suttisak Wizadwongsa$^{1,2}$\thanks{Equal contribution} \qquad
    Jinfan Zhou$^1$\footnotemark[1] \qquad
    Edward Li$^1$ \qquad Jeong Joon Park$^1$ \\
    $^1$University of Michigan, Ann Arbor  \qquad $^2$ Vidyasirimedhi Institute of Science and Technology\\
    {\tt\small suttisak.w\_s19@vistec.ac.th} \qquad {\tt\small \{zjf, edwarli, jjparkcv\}@umich.edu} \\
    \href{https://triflow.github.io/}{https://triflow.github.io/}
}

\begin{document}
\twocolumn[{
\maketitle


\begin{center}
    \vspace{-1cm}
    \input{img/main2/main_fig}
\label{fig:teaser}
\end{center}
}]
\let\thefootnote\relax\footnotetext{\textsuperscript{*}Equal Contribution}

\vspace{-1cm}
\input{sec/0_abstract}    
\input{sec/1_intro}

\input{sec/2_related}
\input{sec/3_preliminaries}

\input{img/triplane_vis/triplane_vis_fig}

\input{sec/4_method}
\input{sec/4_exps}

\input{sec/5_conclusion}
{
    \small
    \bibliographystyle{ieeenat_fullname}
    \bibliography{main}
}
 \input{sec/X_suppl}

\end{document}

%% file: img/main2/main_fig.tex
    \centering
    \begin{tabular}{c  c}
        \includegraphics[width=0.12\textwidth]{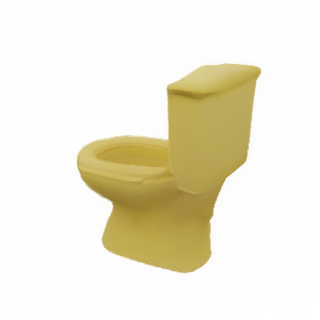} 
        \includegraphics[width=0.12\textwidth]{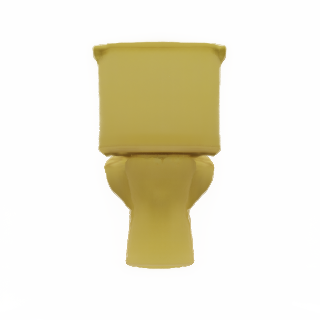}
        \includegraphics[width=0.12\textwidth]{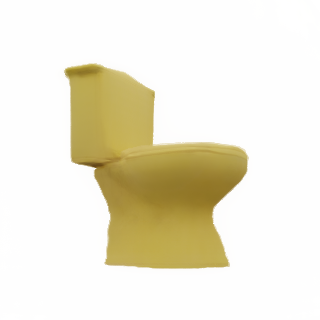}
        \includegraphics[width=0.12\textwidth]{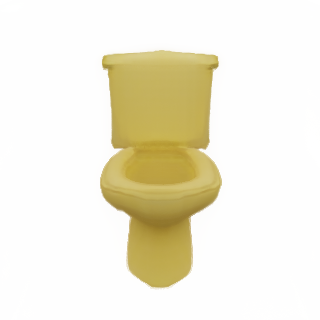} &
        \includegraphics[width=0.12\textwidth]{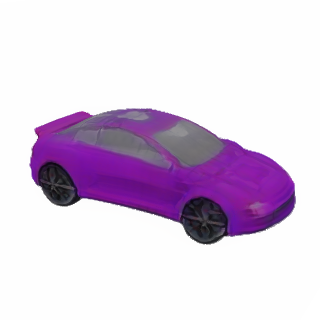} 
        \includegraphics[width=0.12\textwidth]{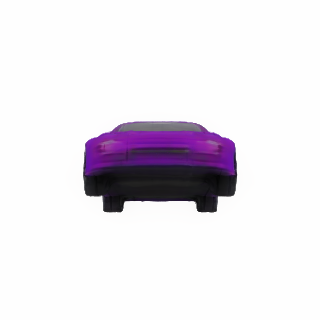} 
        \includegraphics[width=0.12\textwidth]{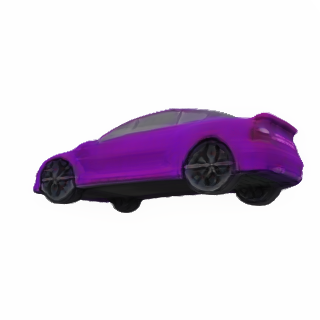} 
        \includegraphics[width=0.12\textwidth]{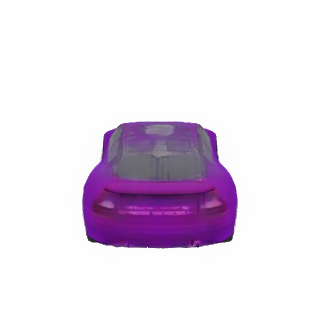}  \\
        ``a golden yellow toilet'' & ``a purple car with modern futuristic design'' \\
        
        \includegraphics[width=0.12\textwidth]{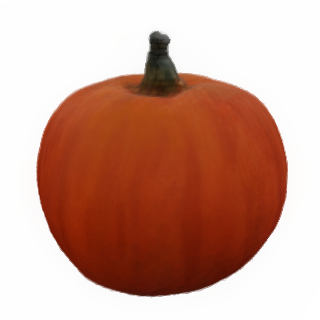} 
        \includegraphics[width=0.12\textwidth]{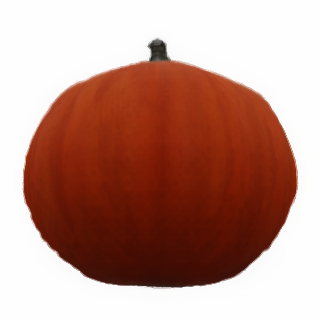}
        \includegraphics[width=0.12\textwidth]{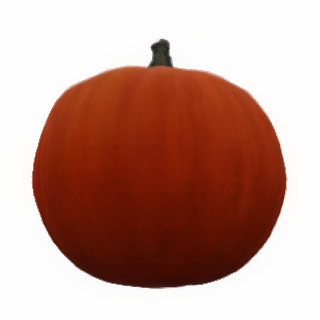}
        \includegraphics[width=0.12\textwidth]{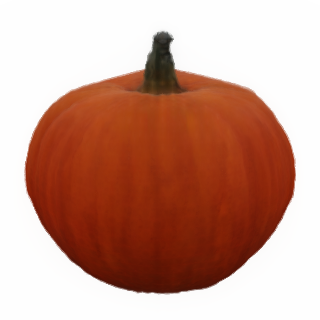} &
        \includegraphics[width=0.12\textwidth]{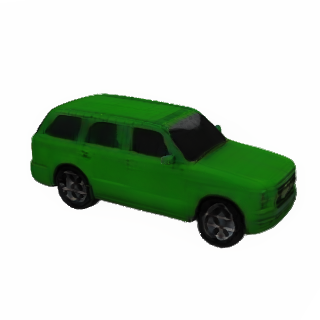} 
        \includegraphics[width=0.12\textwidth]{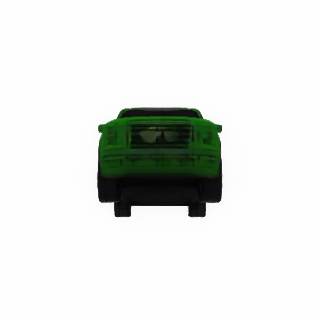} 
        \includegraphics[width=0.12\textwidth]{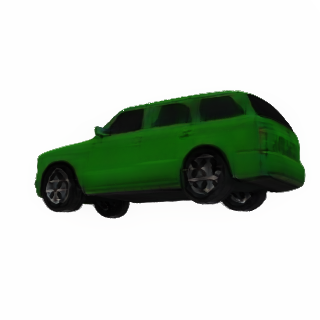} 
        \includegraphics[width=0.12\textwidth]{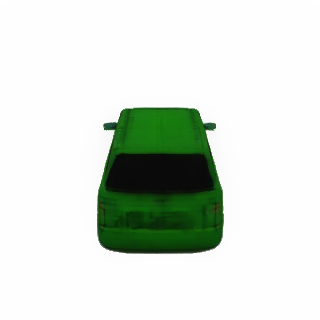}  \\
        ``a large pumpkin'' & ``a green suv car'' \\
        
        \includegraphics[width=0.12\textwidth]{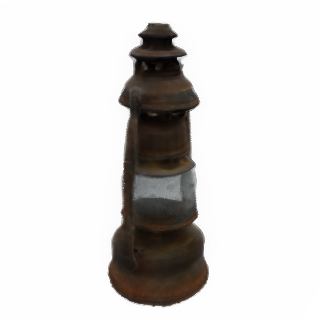} 
        \includegraphics[width=0.12\textwidth]{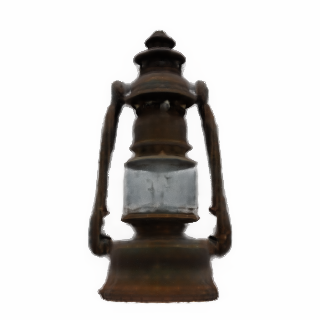}
        \includegraphics[width=0.12\textwidth]{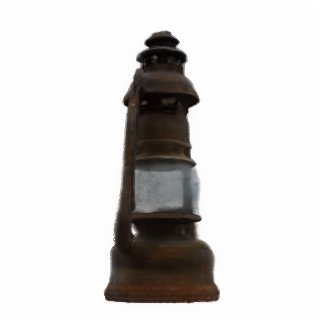}
        \includegraphics[width=0.12\textwidth]{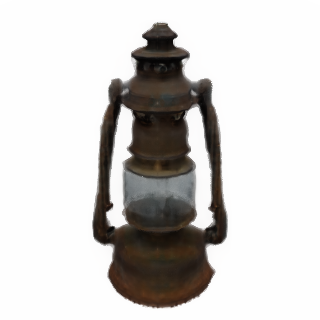} &
        \includegraphics[width=0.12\textwidth]{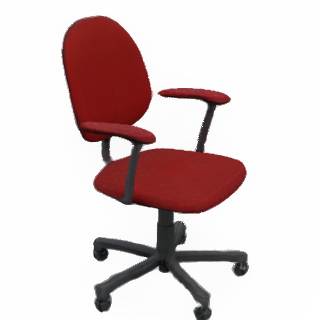} 
        \includegraphics[width=0.12\textwidth]{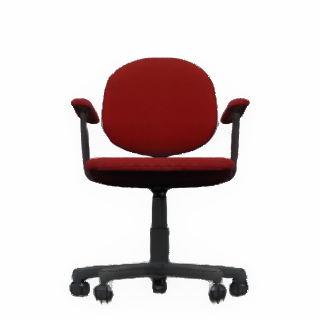} 
        \includegraphics[width=0.12\textwidth]{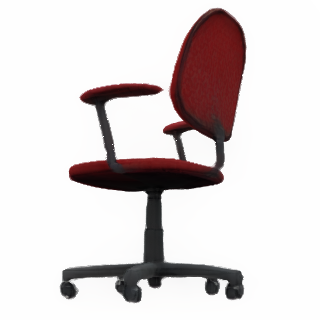} 
        \includegraphics[width=0.12\textwidth]{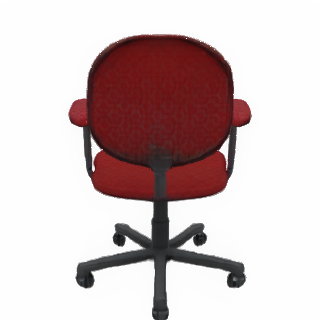}  \\
        ``an old-fashioned oil lamp'' & ``a red office chair with wheels'' \\
        
        \includegraphics[width=0.12\textwidth]{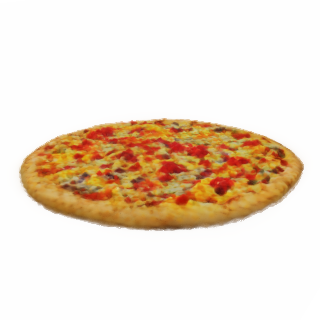} 
        \includegraphics[width=0.12\textwidth]{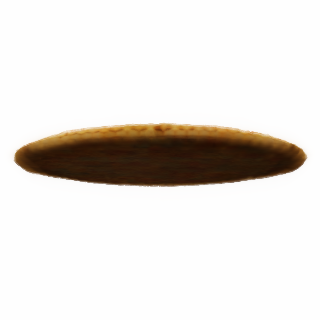}
        \includegraphics[width=0.12\textwidth]{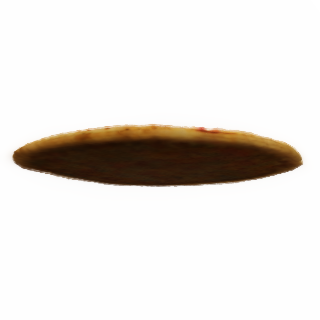}
        \includegraphics[width=0.12\textwidth]{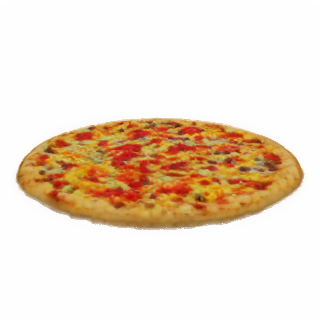} &
        \includegraphics[width=0.12\textwidth]{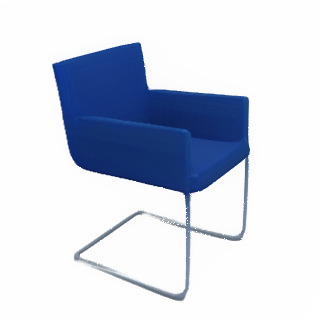} 
        \includegraphics[width=0.12\textwidth]{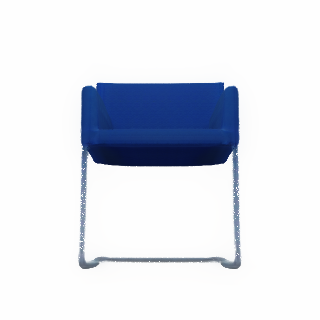} 
        \includegraphics[width=0.12\textwidth]{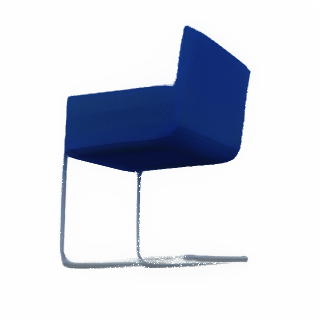} 
        \includegraphics[width=0.12\textwidth]{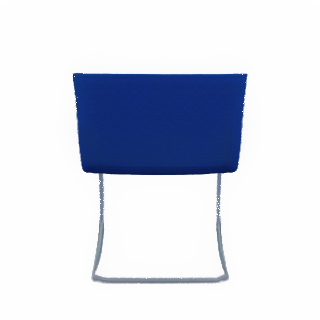}  \\
        ``a pizza with cheese and pepperoni topping'' & ``a blue living room chair'' \\
    \end{tabular}
    \captionof{figure}{\textbf{Text-to-3D generation.} 
    Our TriFlow model is trained on triplanes from a pretrained feed-forward reconstruction model \cite{xu2024instantmesh} and can generate a high-quality 3D model in a few seconds. Left column: samples of a model trained on Objaverse \cite{deitke2024objaverse} LVIS. Right: samples of models fine-tuned on ShapeNet \cite{chang2015shapenet} chairs and cars.}
    \label{fig:main2_comparison}
    \vspace{-0.1cm}

%% file: sec/0_abstract.tex
{\centering\large\bfseries Abstract\par}
\vspace{-0.cm}
{\itshape Recent AI-based 3D content creation has largely evolved along two paths: feed-forward image-to-3D reconstruction approaches and 3D generative models trained with 2D or 3D supervision. In this work, we show that existing feed-forward reconstruction methods can serve as effective latent encoders for training 3D generative models, thereby bridging these two paradigms.
By reusing powerful pre-trained reconstruction models, we avoid computationally expensive encoder network training and obtain rich 3D latent features for generative modeling for free.
However, the latent spaces of reconstruction models are not well-suited for generative modeling due to their unstructured nature. 
To enable flow-based model training on these latent features, we develop post-processing pipelines, including protocols to standardize the features and spatial weighting to concentrate on important regions. We further incorporate a 2D image space perceptual rendering loss to handle the high-dimensional latent spaces. Finally, we propose a multi-stream transformer-based rectified flow architecture to achieve linear scaling and high-quality text-conditioned 3D generation. Our framework leverages the advancements of feed-forward reconstruction models to enhance the scalability of 3D generative modeling, achieving both high computational efficiency and state-of-the-art performance in text-to-3D generation.
}


%% file: sec/1_intro.tex
\section{Introduction}
\label{sec:intro}

The demand for digital content across gaming, AR/VR, and interactive visualization continues to drive innovation in automatic 3D asset generation. Recent advances in AI-driven 3D content creation have evolved along two primary trajectories: 3D generative models trained with 2D or 3D supervision \cite{poole2022dreamfusion,lan2025ln3diff,vahdat2022lion}, and feed-forward image-to-3D reconstruction approaches \cite{hong2023lrm,xu2024instantmesh,boss2024sf3d}. While both paths have demonstrated compelling results, they have largely evolved independently. We propose bridging these paradigms by repurposing pre-trained reconstruction models as latent encoders for training scalable 3D generative models.

Diffusion models have revolutionized 2D image synthesis by leveraging VAE-based latent representations for efficient training on massive datasets. However, 3D generation faces unique challenges due to limited data availability and the complexity of multi-view representations. The absence of effective 3D latent encoders has led researchers to pursue either computationally intensive optimization-based approaches for obtaining 3D latents \cite{shue20233d,chen2023single,liu2024direct, cao2023large} or methods that generate 3D content directly from 2D priors \cite{poole2022dreamfusion,bahmani20244d}. 

Recent approaches like \cite{lan2025ln3diff} have demonstrated the potential of feed-forward VAEs for generating triplane latents, offering processing speeds comparable to 2D VAEs. However, these methods require training dataset-specific image-to-triplane VAEs with specialized loss functions—a computationally intensive process that rivals the training time of the diffusion model itself, typically requiring 7 days with several powerful GPUs.

Instead, We propose leveraging existing state-of-the-art reconstruction models as alternatives to specialized VAE training. Recent advances in feed-forward methods \cite{hong2023lrm,xu2024instantmesh} have produced highly accurate triplane and Gaussian Splatting representations from single or multi-view images. These models, whose training requires significant computational resources (100s of GPUs over multiple days), can handle diverse scenes effectively. Rather than duplicating this computational effort with specialized encoders, we show that flow-based models can be trained directly in the latent spaces of these pre-trained reconstruction models.

However, these feed-forward reconstruction models were not designed to serve as encoders for generative modeling, presenting several challenges in their direct application. First, unlike VAEs which are regularized towards producing standardized isotropic Gaussian distributions, their latent features lack consistent statistical properties. Second, their unregularized latent spaces contain high-frequency artifacts that don't meaningfully impact the final rendered output but degrade training convergence. Third, the high dimensionality of these latent spaces creates a significant discrepancy between latent space and rendering-space losses, complicating diffusion-like training approaches. Finally, training flow-based models on high-dimensional triplane representations is more computationally demanding than conventional 2D image generation.

To address these challenges, we develop a comprehensive pipeline for generative training from reconstructed latents. First, we introduce a feature standardization protocol that transforms the latent distributions to match VAE-like properties. Second, we implement spatial importance weighting that focuses computational resources on regions most relevant to the final rendering, effectively suppressing noise in less significant areas. To bridge the gap between latent and rendering spaces, we incorporate a perceptual rendering loss that ensures generated features produce visually realistic images. Finally, we propose a multi-stream transformer architecture, dubbed TriFlow, that processes these high-dimensional features efficiently through parallel attention mechanisms, achieving linear computational scaling with respect to the number of triplane tokens. This combination of techniques enables effective training of flow-based generative models on the reconstruction latent features.

Our extensive experiments demonstrate that our approach matches or exceeds state-of-the-art 3D generation methods in unconditional generation and outperforms in text-conditional generation, despite relying on off-the-shelf reconstruction models rather than specially trained encoders. These results validate that existing reconstruction models can serve as effective latent encoders for high-quality 3D generation, presenting a practical path toward scalable 3D content creation.

%% file: sec/2_related.tex
\vspace{-0.1cm}
\section{Related Work}
\label{sec:related_work}
\vspace{-0.1cm}

\textbf{3D Generative Models.} With the advancement generative models including VAE~\cite{kingma2013auto}, generative adversarial networks (GANs)~\cite{goodfellow2014generative, kang2023scaling}, and diffusion models~\cite{ho2020denoising, song2020score, karras2022elucidating}, we have seen substantial progress in content generation. Similarly, the rapid evolution of 3D scene modeling~\cite{mildenhall2021nerf, kerbl20233d, park2019deepsdf} has driven significant growth in 3D generative models. Early efforts utilizing 3D-aware GANs~\cite{or2022stylesdf, chan2022efficient, schwarz2020graf, chan2021pi, gao2022get3d} successfully generate high-quality 3D models but still suffered from the intrinsic issues of GANs such as mode collapse, training instability, and limited scalability~\cite{brock1809large, kang2023scaling}. As diffusion models~\cite{ho2020denoising, song2020score, karras2022elucidating} have demonstrated superior performance in 2D image generation~\cite{dhariwal2021diffusion}, the 3D generation community has also begun exploring 3D generative models with diffusion models~\cite{shue20233d, muller2023diffrf, chen2023single, liu2024direct, cao2023large, lan2025ln3diff, poole2022dreamfusion, wang2023score, wang2024prolificdreamer, gupta20233dgen}. One line of work~\cite{poole2022dreamfusion, wang2023score, wang2024prolificdreamer} involves distilling scores from pretrained 2D diffusion models, which requires costly per-scene optimization and still faces challenges like mode collapse~\cite{poole2022dreamfusion} and the Janus problem~\cite{armandpour2023re}. Another line of work deploys a two-stage training process~\cite{shue20233d, muller2023diffrf, cao2023large, lan2025ln3diff}. An auto-decoder~\cite{shue20233d, muller2023diffrf} or VAE~\cite{lan2025ln3diff} is pre-trained to encode the latent feature space, forming a dataset from which diffusion models learn the latent feature distribution. However, the pre-training phase in these methods can be time-consuming. SSDNeRF~\cite{chen2023single} and its follow-up work Direct-3D~\cite{liu2024direct} streamline this by integrating decoder and diffusion model training into a single-stage process, but still demands high computation cost to train the model on a large scale dataset. In this paper, we propose a novel approach that directly leverages the latent space learned by 3D reconstruction models and then trains a diffusion model to learn the distribution.

\noindent \textbf{Feed-forward Reconstruction Models.}  
Recent advances in 3D reconstruction models, such as LRM~\cite{hong2023lrm}, have shown that fast and reliable 3D generation is possible with large 3D datasets~\cite{deitke2023objaverse, deitke2024objaverse}, allowing efficient sparse-view reconstruction of large-scale, unseen objects. Follow-up approaches, including TripoSR~\cite{tochilkin2024triposr}, InstantMesh~\cite{xu2024instantmesh}, and SF3D~\cite{boss2024sf3d}, further build on these models, aiming to improve both speed and quality in reconstruction. Additionally, methods like Instant3D~\cite{li2023instant3d} incorporate priors from 2D diffusion models and use multi-view diffusion techniques to reduce reconstruction uncertainty. LN3Diff~\cite{lan2025ln3diff} not only develops an enhanced version of feed-forward models but also leverages these models to create datasets specifically for training diffusion models, facilitating broader use in 3D generation tasks. However, they argue that existing feed-forward models cannot be effectively used for 3D diffusion training, contrasting with what we propose.

\noindent \textbf{Latent Diffusion Models.}  
Latent diffusion models (LDM)~\cite{rombach2022high} utilize VAEs to compress high-resolution images into a low-dimensional latent space, balancing reconstruction fidelity with computational efficiency. In the 2D domain, state-of-the-art models such as Stable Diffusion~\cite{esser2024scaling} follow this approach by training within a pre-trained latent space. In the 3D domain, models like 3DGen~\cite{gupta20233dgen} and L3NDiff~\cite{lan2025ln3diff} employ pre-trained feed-forward networks with 3D supervision, leveraging point clouds and depth information to model the latent feature space of 3D objects effectively.

%% file: sec/3_preliminaries.tex
\begin{figure*}
    \centering
    \includegraphics[width=1.0\textwidth]{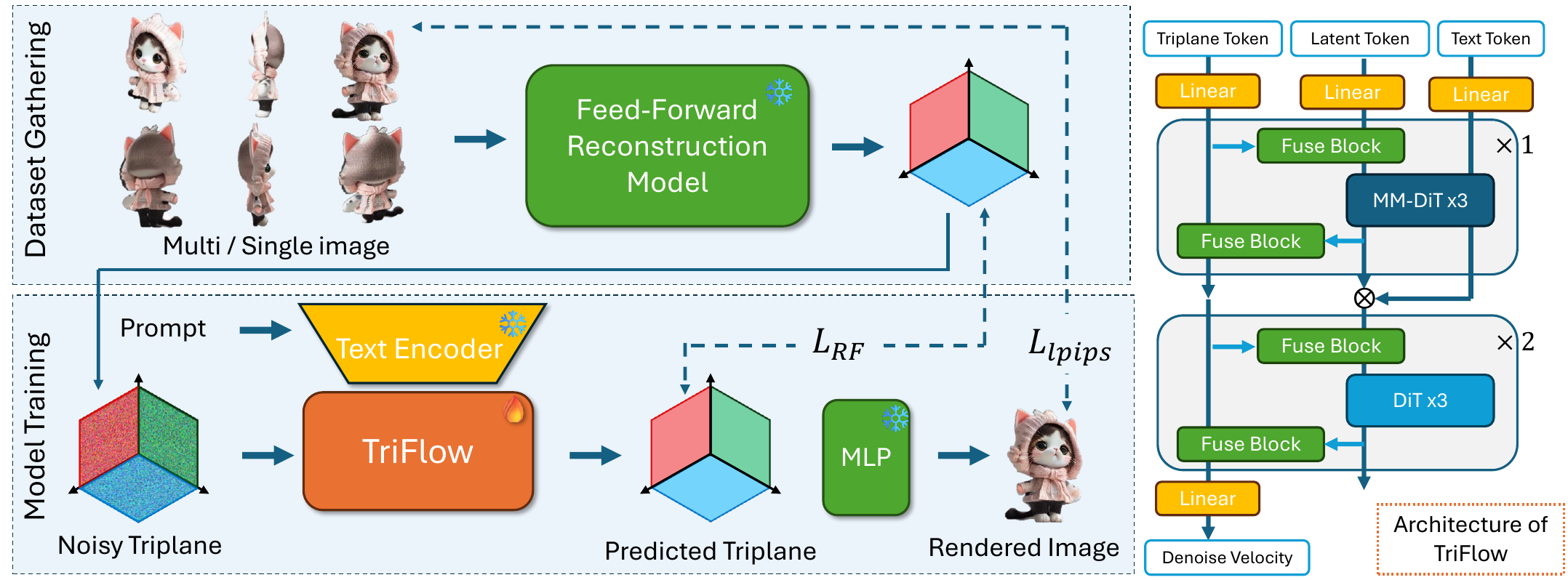}\\
    \caption{\textbf{Overview of our image-to-3D generation pipeline and architecture.} Our framework includes two main components: (1) a dataset preparation stage, where single-view or multi-view images are processed through a feed-forward image-to-triplane model to generate triplanes, and (2) TriFlow, a text-conditioned generative model trained on these triplanes using rectified-flow-based loss ($L_{RF}$) and perceptual loss ($L_{lpips}$) compared against the original images. On the right, we show our model architecture, a multi-stream transformer incorporating a combination of MM-DiT and DiT blocks. Note that $\otimes$ denotes concatenation between token streams.}
    \label{fig:overview}
\end{figure*}

\section{Methods}

\subsection{Overview}
Training diffusion models in the latent space is a widely accepted concept in the 2D domain. In 3D, various methods have proposed training separate VAEs as latent encoders to create triplane latents for diffusion training \cite{cao2023large,lan2025ln3diff,shue20233d}
. However, no methods to date have successfully utilized {\em existing} feed-forward reconstruction models as latent encoders. As noted in \cite{lan2025ln3diff}, this is largely due to the unstructured distributions of the triplanes created by the feed-forward models, which are not designed for generative modeling in their latent spaces. In this section, we describe a series of processes we developed to make the reconstructed triplanes more amenable to generative training and to improve computational efficiency in training generative models leveraging the powerful pre-trained reconstruction models.


\subsection{Reconstructing 3D Latent Features}

We aim to use existing, powerful feed-forward reconstruction methods to obtain the latent features for training diffusion-like models, removing the need for separate VAE training on different datasets.

Among various pre-trained feed-forward models, we focused on those using triplane intermediate representations, which are particularly suited for generation tasks. We selected InstantMesh~\cite{xu2024instantmesh}, a state-of-the-art reconstruction model, as our feed-forward latent encoder. InstantMesh transforms single or multi-view images into a triplane, which can then be rendered as images using Mip-NeRF-style ray marching. Similar to other feed-forward methods \cite{hong2023lrm}, it produces 64x64 resolution triplanes with C-dimensional features (hence, $3\times C\times H\times W$) that can be decoded into RGB and density values through a four-layer MLP decoder. We emphasize that our proposed processing pipelines and training insights are general and not restricted to a particular feed-forward model, and should apply to other latent spaces.


\subsection{Standardizing the 3D Latent Features}\label{sec:standard}

Crucially, directly using triplanes from the feed-forward models introduces challenges for training a generative model due to their high-variance, unstructured latent spaces. These characteristics are noted by various methods exploring latent diffusion for 3D generation \cite{shue20233d,lan2025ln3diff}. Diffusion models tend to favor training on zero-mean Gaussian data as shown in \cite{rombach2022high}. One can train a VAE or VQGAN with KL-reguliarzation to both standardize and compress the latent space in 3D. However, we found that the VAE option incurs significant computational overheads and complicates later fine-tuning stage described in Sec.~\ref{sec:recon}.

Consequently, instead of training a VAE, we opt for a simpler standardization technique that involves estimating the mean ($\mu$) and covariance matrix ($\Sigma$) of all pixels across the entire dataset. We compute an eigendecomposition on the full covariance matrix to obtain $\Sigma = E D E^T$, where $E$ is the matrix of eigenvectors and $D$ is a diagonal matrix of eigenvalues. We can then standardize the triplane latent space so that they statistically follow a standard Normal distribution:
\begin{align}
Z = E D^{-1/2} E^T (X - \mu),
\end{align}
where \( X \in \mathcal{R}^C\) and \( Z \in \mathcal{R}^C\) represent the triplane before and after standardization. 

Importantly, we note that it is empirically beneficial to compute the generative loss in the original feature space $X$, so we revert the standardization before computing the loss. The original pixel values can be recovered as follows:
\begin{align}
X =  E D^{1/2} E^T Z + \mu.
\end{align}
This approach stabilizes the feature space for learning while preserving accuracy in the original representation.

\subsection{Rectified Flow with Spatial Weighting}
We adopt the state-of-the-art image generative algorithm, Rectified Flow \cite{liu2022flow}, which is adopted by the newest Stable Diffusion version to replace the diffusion model \cite{ho2020denoising}. Rectified flow creates linear trajectories between Gaussian noise and structured data, enabling more effective denoising. 
The path between standardized data \( Z_0 \) and noise \( Z_1 \) is defined as:
 \( Z_t = (1 - t) Z_0 + t Z_1 \)  for any   \( t \in [0, 1] \). 
Rectified flow employs an ordinary differential equation (ODE), \( dZ_t = v_\theta(Z_t, t) \, dt \), to define a velocity field \( v_\theta(Z_t, t) \), which is approximated by a neural network parameterized by \( \theta \). The model is trained to align the velocity along the trajectory by minimizing the loss:  \( \lVert v_\theta(Z_t, t) - (Z_1 - Z_0) \rVert^2 \).

\paragraph{Triplane Empty Space Masking.} When training a triplane-based generative model, a unique challenge arises: a large proportion of the triplane pixels describe empty space, which does not significantly contribute to the final object rendering. Crucially, these pixel features still present significant variations, as visualized using linear projection in Fig.~\ref{fig:triplane_vis}. Therefore, when naively training a diffusion-like model on these triplanes, a significant amount of the network capacity is wasted for modeling the meaningless variations in the empty space, producing low-quality results.

To address this, we create triplane weighting masks to highlight the object regions. For each triplane, we sample triplane features across all possible 3D coordinates and use the renderer’s MLP to evaluate their density. We then identify and mask regions that are empty versus those containing object details. As shown in Figure 1, typically only 15–30 percent of the triplane space contains object information.

To encourage the network to focus on the object regions during the rectified flow training, we apply a binary pixel mask $M$  with fixed weightings of 1 for the object pixels and 0.25 for the empty pixels. The mask $M$ is multiplied to the final loss to de-weight the empty pixels as follows:
\begin{align*}
L_{\text{RF}} = \left\| M \odot \left(E D^{1/2} E^T \left( v_\theta(Z_t, t) - (Z_1 - Z_0) \right) \right) \right\|^2,   
\end{align*} 

where $\odot$ represents the element-wise (Hadamard) product. Note that we're reverting the standardization before computing the final norm as explained in Sec~\ref{sec:standard}.
This targeted approach helps the model focus on meaningful areas, significantly accelerating and improving the generative training.

\subsection{Perceptual Rendering Loss}\label{sec:recon}
Training generative models in triplane space presents another unique challenge: triplanes have higher channel dimensionality ($C=80$) compared to 2D image latent features ($C=4$), and the importance of each dimension varies significantly in the final rendering. Consequently, equal magnitudes of error in triplane space can result in vastly different perceptual errors in rendered images. This dimensional mismatch necessitates an additional loss function to ensure perceptual quality.

To address this challenge, we augment the flow loss with a reconstruction loss that better aligns with rendered image quality. After obtaining the initial velocity prediction, we compute an approximation of the clean triplane:
\begin{align}
\hat{Z}_0 = Z_t - t \cdot v_\theta (Z_t, t).
\end{align}
We then render an image patch $Y(\hat{Z}_0, c)$ using a randomly sampled camera view $c$ and compare it to the corresponding ground-truth image $\bar{Y}_c$ using perceptual similarity:
\begin{align}
    L_{lpips} = \text{LPIPS}(Y(\hat{Z}_0, c), \bar{Y}_c).
\end{align}
This LPIPS loss \cite{zhang2018unreasonable} guides the model to learn the relative importance of triplane dimensions and potentially exceed the quality achievable by the triplane encoder alone.

However, since patch rendering is computationally expensive and less critical in early training, we implement a two-stage training strategy. Stage I focuses solely on the denoising objective $L_{RF}$, while Stage II incorporates both denoising and reconstruction objectives $L_{RF} + L_{lpips}$. This approach optimally balances training efficiency with rendered image quality.


\subsection{Efficient Multi-Stream Architecture}

Our proposed architecture, TriFlow, addresses text-conditioned triplane generation through a novel multi-stream design that achieves both efficiency and generation quality. At its core, we use MM-DiT modules from Stable Diffusion 3.0 \cite{esser2024scaling} that enable parallel processing of text and noised latent representations. While MM-DiT excels at text-latent alignment through pretrained encoders, its high parameter count poses computational challenges with triplanes. Following insights from AuraFlow \cite{cloneofsimo2024}, 
we strategically combine MM-DiT with DiT blocks \cite{peebles2023scalable} to reduce computation while maintaining quality.

Inspired by PointInfinity \cite{huang2024pointinfinity}, we achieve linear scaling through a multi-stream transformer architecture. It includes a triplane stream processing spatial information and two streams handling text conditioning and feature transformation. The triplane streams communicate with others through Fuse blocks: the first enables triplane-informed latent processing, while the second updates triplane representations based on latent computations (see Fig.~\ref{fig:overview}).

By restricting self-attention to the smaller latent streams and strategically placing MM-DiT for text conditioning, our architecture maintains linear scaling with triplane size. This control of information flow between streams allows for efficient computation, enabling robust text-conditioned triplane generation at scale.

%% file: img/triplane_vis/triplane_vis_fig.tex
\begin{figure}[t]
    \centering
        \includegraphics[width=0.75\linewidth]{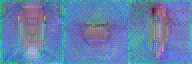} 
         \\
        \includegraphics[width=0.75\linewidth]{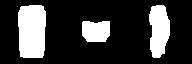} 
    \caption{\textbf{Visualization of triplane features} produced by InstantMesh \cite{xu2024instantmesh} and the mask used for the weighted loss during training. Note the severe noise of the features in the empty spaces.}
    \label{fig:triplane_vis}
    \vspace{-0.3cm}
\end{figure}

%% file: sec/4_exps.tex
\section{Experiments}

\input{img/shapenet_con/fig}
We begin by training our main model on the Objaverse dataset (LVIS) \cite{deitke2023objaverse}, a large 3D dataset comprising over 100k objects across various categories. After filtering out lower-quality objects and excluding those that InstantMesh cannot reliably reconstruct, we reduce the dataset to approximately 80k objects. For text captions, we randomly select captions from two sources, 3DTopia \cite{hong20243dtopia} and Cap3D \cite{luo2024scalable}, to maximize caption availability. Model training spans several weeks and is conducted using 4 L40S GPUs across both Stage I and Stage II of our method. This final checkpoint is used for evaluation and also serves as the starting point for some future experiments.

\begin{table*}[h]
\centering
\begin{tabular}{llllll}
\hline
                                            & FID$\downarrow$  & CMMD$\downarrow$  & Prec$\uparrow$   & Rec$\uparrow$ & CLIP$\uparrow$ \\ \hline
a) Stage I with single-stream transformer   &35.07 & 0.624 & 0.3801& 0.1125 & 24.63\\ 
b) Stage I train from scratch               &32.05 & 0.579 & 0.4090& 0.1172 & 24.80\\ 
c) Stage I from scratch and no mask         &36.77 & 0.673 & 0.3441& 0.0997 & 24.63\\ 
d) Stage I train from main checkpoint  &24.54 & 0.363 & 0.7460& 0.0544 & 24.73\\ 
e) Stage I+I: extended stage I              &22.09 & 0.358 & \textbf{0.7520}& 0.0776 & 24.80\\
f) Stage I+II: complete model               &\textbf{14.80} &\textbf{0.297} & {0.6829}& \textbf{0.1601} & \textbf{24.81}\\ \hline
g) Triplane Encoded (InstantMesh)           & 3.418 & 0.077 & 0.9438& 0.9438 & 23.81 \\ 
\end{tabular} 
\caption{\textbf{Ablation study} of our model measured by the generated image quality (a-f). This study is performed for text-conditional generation on ShapeNet Chair. As expected, our full model with multi-stream transformer, mask weighting, and perceptual loss fine-tuning leads to the best outcome. We also provide the performance of the InstantMesh's image-to-3D reconstruction shown in (g) as a reference. }
\vspace{-0.25cm}
\label{tab:abalation}
\end{table*}

\subsection{Ablation Studies} \label{sec:ablation}
We conducted an ablation study to validate our algorithm’s design choices, summarized in Table \ref{tab:abalation}. All experiments used the ShapeNet Chair \cite{chang2015shapenet} dataset and text prompts.
Each model was used to recreate the Chair dataset based on all training text prompts, sampling each instance with 30 steps and a guidance scale of 4. Each triplane was rendered into six views around each object using InstantMesh input camera views.
We evaluated the renderings using several metrics: the FID score (lower is better), CLIP Maximum Mean Discrepancy (CMMD, lower is better), improved Precision and Recall (Prec/Rec, higher is better), and CLIP score using ViT-L/14 (higher is better), which indicates alignment between the generated and target representations.

The study examines several design choices. First, we compare a standard single-stream transformer (a) with our multi-stream transformer (b), both trained with \(L_{RF}\) loss for 100k steps, showing the advantages of the multi-stream design. We then analyze the impact of the triplane mask in \(L_{RF}\) by removing it (c), which results in reduced quality compared to (b).
Next, we assess the effect of pre-trained weights by initializing from a checkpoint extensively trained on LVIS, showing quality improvements over (b) in (d). To test our staged training approach, we perform Stage II training from checkpoint (b) with \(L_{RF} + L_{lpips}\) losses over 80k steps, yielding further quality gains in (f). As a control, we extend Stage I training on (b) for an additional 80k steps, shown in (e). Lastly, we compare our triplane dataset rendering to the original images in (g), noting that the trained model’s CLIP score already surpasses that of the dataset in configuration (g). 

\begin{figure}[t]
    \centering
        \includegraphics[width=0.85\linewidth]{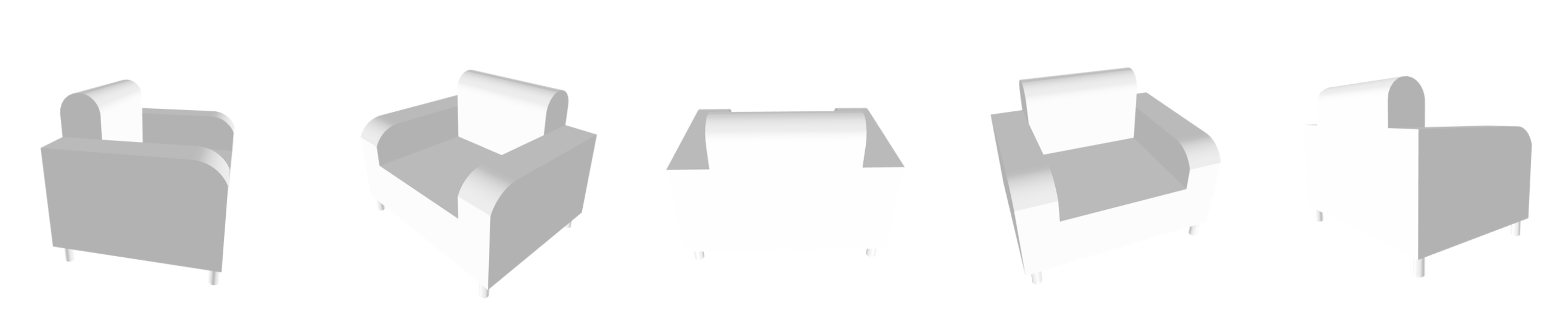} \\
        \vspace{-0.1cm}
        {\small (a) Bad Input Data}
        \vspace{-0.1cm}
        \includegraphics[width=0.87\linewidth]{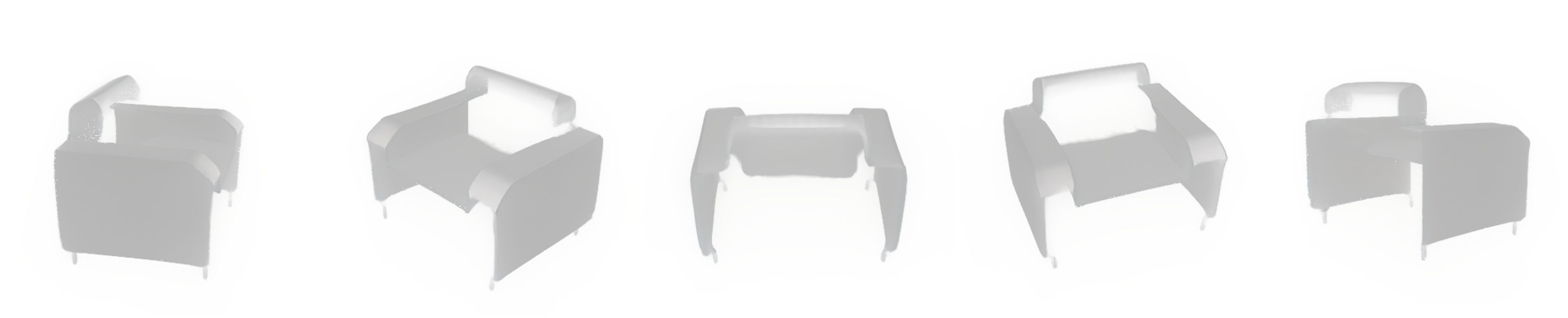} \\
        {\small (b) Reconstruction Renderings}
        \vspace{-0.1cm}
        \includegraphics[width=0.87\linewidth]{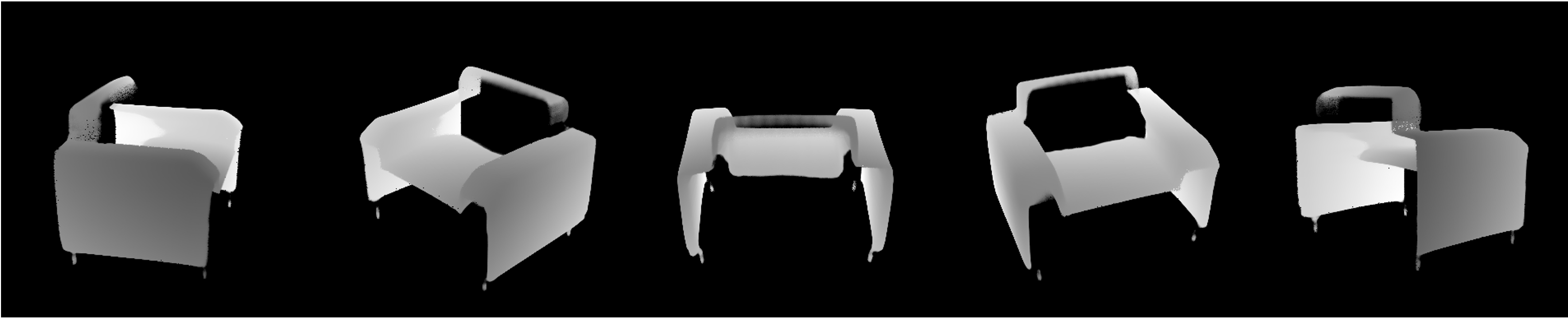}  \\
        {\small (c) Reconstruction Depths}
        \vspace{-0.1cm}
    \caption{Some ShapeNet objects cause artifacts when rendered without textures following existing works (a), which damages the geometry inference of feed-forward reconstruction methods (b,c).}
    \label{fig:bad_render}
    \vspace{-0.3cm}
\end{figure}

\subsection{Text Conditional Triplane Generation}\label{sec: text}
The goal of this experiment is to evaluate our text-conditioned approach for text-to-3D generation. As shown in Fig. \ref{fig:main2_comparison}, our method can generate triplane representations based on text inputs.
We compared our model with two recent SOTA text-to-triplane generative models: LN3Diff \cite{lan2025ln3diff} and Direct3D \cite{liu2024direct}. Both models were trained on a portion of the Objaverse dataset, though with slightly different text captions. For a fair comparison, we used 150 text captions focused on common daily objects, randomly generated by ChatGPT, for all models under identical conditions. We then generated six rendered views per prompt and assessed text-to-image alignment using two metrics: CLIP Score \cite{hessel2021clipscore} and CLIP R-Precision \cite{park2021rprecision} (higher is better for both), as shown in Tab. \ref{tab:text_con_compare_main}. For both metrics, we evaluate using ViT-B/32 and ViT-L/14 CLIP models \cite{Radford2021LearningTV}.

In Fig. \ref{fig:text_con_compare_main}, LN3Diff \cite{lan2025ln3diff} demonstrates some degree of text understanding but struggles with unseen prompts, resulting in reduced quality. Direct3D \cite{liu2024direct} provides two checkpoints: one $(T=0.002)$, trained on a smaller dataset, which generates high-quality triplanes but often lacks relevance to the text prompt; the other $(T=0.07)$, trained on a larger dataset, shows improved prompt alignment but at a lower output quality. In contrast, our model achieves superior text alignment and vocabulary range, marking a significant improvement in high-quality, text-controllable 3D generation.

\begin{table}[h]
    \centering
    \begin{tabular}{c|cc|cc}
     & \multicolumn{2}{c|}{CLIP score$\uparrow$} & \multicolumn{2}{c}{CLIP R-precision$\uparrow$} \\
     & ViT-B & ViT-L & ViT-B & ViT-L \\
    \hline
    LN3Diff~\cite{lan2025ln3diff} & 25.75 & 20.29 & 34.0\% & 35.33\% \\
    \shortstack{Direct3D$_{T=.002}$\cite{liu2024direct}} & 26.08 & 20.40 & 40.0\% & 43.33\% \\
    \shortstack{Direct3D$_{T=.07}$\cite{liu2024direct}} & 26.84 & 21.41 & 44.7\% & 50.67\% \\
    \textbf{Ours} & \textbf{27.61} & \textbf{22.21} & \textbf{52.0\%} & \textbf{54.67\%} \\
    \end{tabular}
    \caption{\textbf{Quantitative comparisons on Objaverse} for text-conditional triplane generation using CLIP~\cite{Radford2021LearningTV} metrics.}
    \label{tab:text_con_compare_main}
\end{table}
\input{img/vs_ln3diff/main_fig2}

\noindent  \textbf{Text-Conditioned Single-Class 3D Generation.} 
LN3Diff \cite{lan2025ln3diff} uses ShapeNet with simpler texture renderings than those in our ablation studies. To align with LN3Diff, we fine-tune our main model on the ShapeNet Car, Chair, and Plane classes.
Following the evaluation methods and metrics outlined in Sec. \ref{sec:ablation}, we perform conditional sampling on both models under identical input conditions and camera views. We then evaluate the models using FID \cite{heusel2017gans}, CMMD \cite{jayasumana2024rethinking}, and CLIP score metrics. Results in Tab. \ref{tab:comparison} show that our model significantly outperforms LN3Diff across all categories and metrics.

Notably, LN3Diff’s side-view renders exhibit artifacts under objects (see Fig. \ref{fig:text_con_compare}), likely due to its training with an upper-hemisphere camera distribution that emphasizes top views. In contrast, our model demonstrates robust performance even when rendering from side angles.

\begin{table}[h]
\centering
\begin{tabular}{ll|rrrrr}
 & Method & FID$\downarrow$ & CMMD$\downarrow$  & CLIP$\uparrow$ \\
\hline
\multirow{2}{*}{Car} & LN3Diff~\cite{lan2025ln3diff} & 45.56 & 0.606  & 25.21 \\
& \textbf{Ours} & \textbf{15.64} & \textbf{0.116}  & \textbf{25.56} \\
\hline
\multirow{2}{*}{Plane} & LN3Diff~\cite{lan2025ln3diff} & 44.98 & 0.438  & 23.21 \\
& \textbf{Ours} & \textbf{12.92} & \textbf{0.087} & \textbf{24.70} \\
\hline
\multirow{2}{*}{Chair} & LN3Diff~\cite{lan2025ln3diff} & 42.52 & 0.602  & 21.71 \\
& \textbf{Ours} & \textbf{10.95} & \textbf{0.086} & \textbf{22.41} \\
\end{tabular}
\caption{\textbf{Quantitative comparisons on ShapeNet} for text-conditioned generation.}
\label{tab:comparison}
\end{table}

\subsection{Unconditional tripalne generation}

We evaluate our TriFlow models on ShapeNet Car and ShapeNet Chair datasets \cite{chang2015shapenet}, containing 3,514 and 6,778 instances respectively. For each object, we extract triplane representations using InstantMesh \cite{xu2024instantmesh} from 6 random camera views. Although our model is designed for text-conditioned generation, we evaluate unconditional generation performance using empty text prompts. Detailed training and sampling protocols are provided in the supplementary material.

As shown in Tab.~\ref{tab:uncond_comp}, our approach achieves competitive results even in unconditional generation, outperforming most existing GAN-based and diffusion-based methods. Notably, we achieve the best FID score on ShapeNet Car. On ShapeNet Chair, while our performance is slightly below \cite{lan2025ln3diff}, which uses category-specific VAEs, this comparison requires important context.

Following standard evaluation protocols from previous works \cite{lan2025ln3diff,chan2022efficient,gao2022get3d,cao2023large}, objects are rendered without textures. However, this creates a significant challenge for off-the-shelf reconstruction methods \cite{xu2024instantmesh} when object colors are insufficiently differentiated from the background (example shown in Fig.\ref{fig:bad_render}). In contrast, specialized methods like LN3Diff \cite{lan2025ln3diff} avoid this issue by incorporating depth information during VAE training. Despite these dataset-specific challenges, our method demonstrates strong performance in unconditional generation while excelling at text-to-3D generation tasks, as detailed in Sec.\ref{sec: text}.

\begin{table}[]
\begin{tabular}{lcccc}
\hline
\multirow{2}{*}{Method} & \multicolumn{2}{c}{Car} & \multicolumn{2}{c}{Chair} \\ \cline{2-3} \cline{4-5} 
                        & FID$\downarrow$       & KID(\%)$\downarrow$     & FID$\downarrow$        & KID(\%)$\downarrow$      \\ \hline
EG3D~\cite{chan2022efficient}                  & 33.33     & 1.4         & 26.09      & \underline{1.1}          \\
GET3D~\cite{gao2022get3d}                   & 41.41     & 1.8         & 35.33      & 1.5          \\
DiffRF~\cite{muller2023diffrf}                  & 75.09     & 5.1         & 99.37      & 4.9          \\
RenderDiffusion~\cite{anciukevivcius2023renderdiffusion}         & 46.5      & 4.1         & 53.3       & 6.4          \\
DiffTF~\cite{cao2023large}                 & 36.68     & 1.6         & 35.16      & \underline{1.1}          \\
LN3Diff~\cite{lan2025ln3diff}                & \underline{17.6}      & \textbf{0.49}        & \textbf{16.9}       & \textbf{0.47}         \\ \hline
\textbf{Ours}           & \textbf{16.36 }    & \underline{0.94}        &     \underline{19.22}       &    \underline{1.1}      \\ \hline
\end{tabular}
\caption{\textbf{Unconditional generation comparisons on ShapeNet}. Overall, our method performs competitively but is likely hampered by the rendering artifacts shown in Fig.~\ref{fig:bad_render}, which negatively affect off-the-shelf reconstruction models.}
\vspace{-0.3cm}
\label{tab:uncond_comp}
\end{table}

%% file: img/shapenet_con/fig.tex
\begin{figure*}[htbp]
    \centering
    \begin{tabular}{@{}c@{\hspace{2pt}}c@{\hspace{2pt}}c@{\hspace{2pt}}c@{}}
        \raisebox{0.cm}{\rotatebox{90}{\Large LN3Diff \cite{lan2025ln3diff}}}  &
        \includegraphics[width=0.16\textwidth,viewport={0 0 512 512},clip]{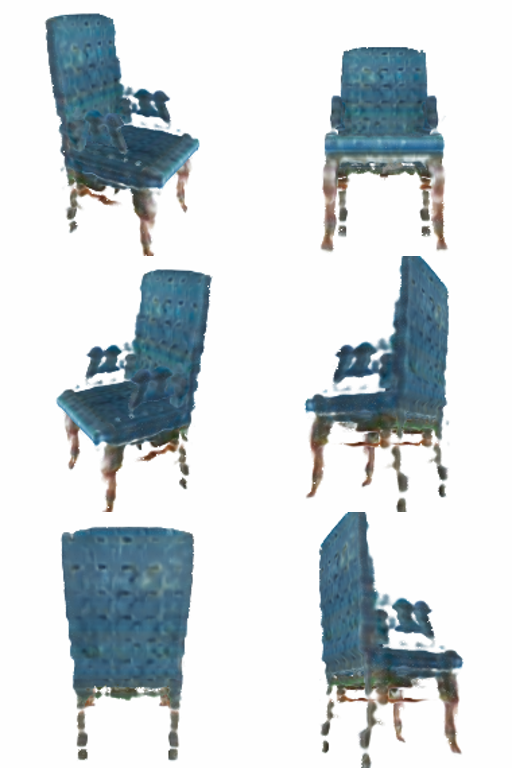}   
        \includegraphics[width=0.16\textwidth,viewport={0 0 512 512},clip]{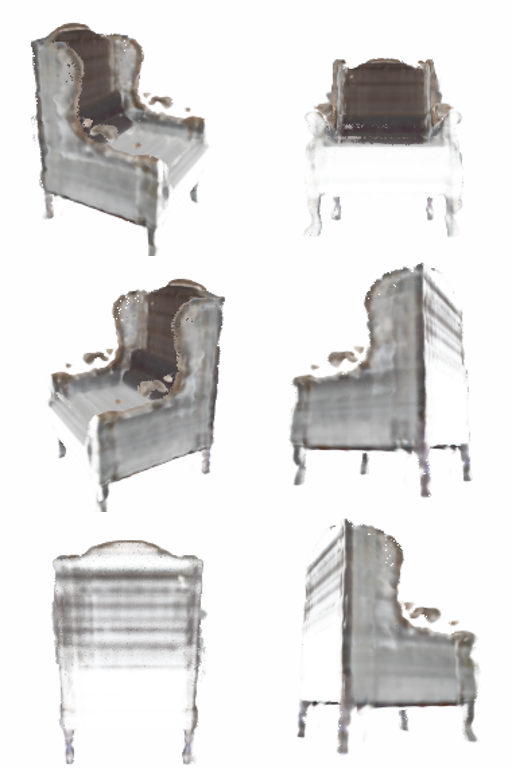}   &  
        \includegraphics[width=0.16\textwidth,viewport={3 3 512 512},clip]{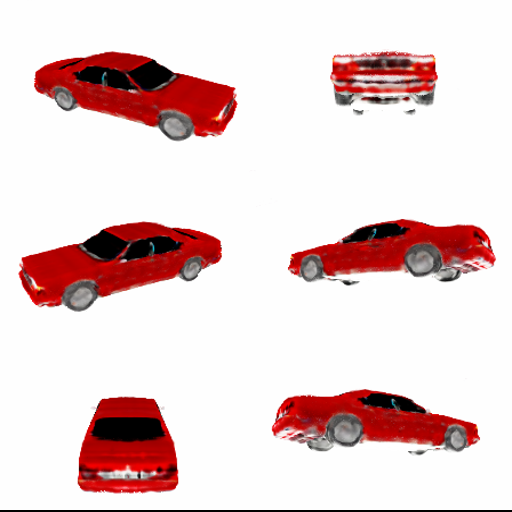} 
        \includegraphics[width=0.16\textwidth,viewport={3 3 512 512},clip]{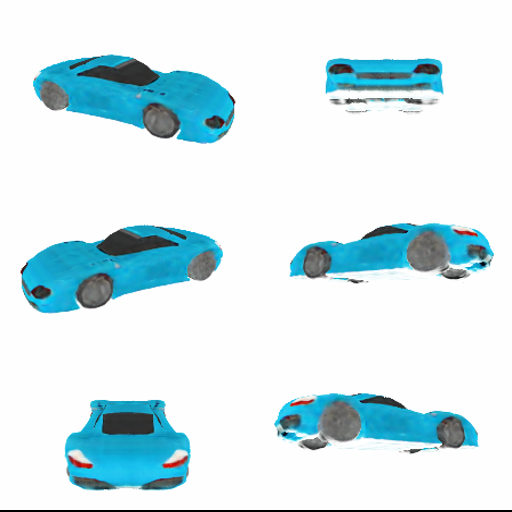}  &
        \includegraphics[width=0.16\textwidth,viewport={3 3 512 512},clip]{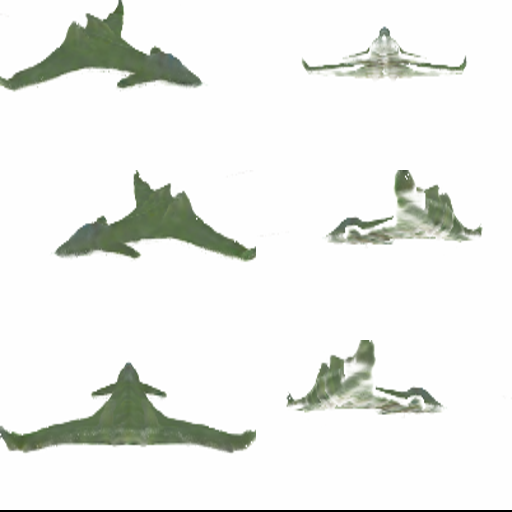} 
        \includegraphics[width=0.16\textwidth,viewport={3 3 512 512},clip]{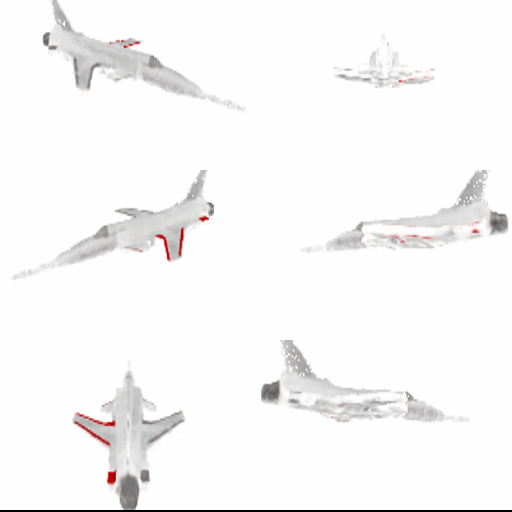}  
        \\
        \raisebox{0.5cm}{\rotatebox{90}{\centering \Large \textbf{Ours}}} &
        \includegraphics[width=0.16\textwidth,viewport={0 0 512 512},clip]{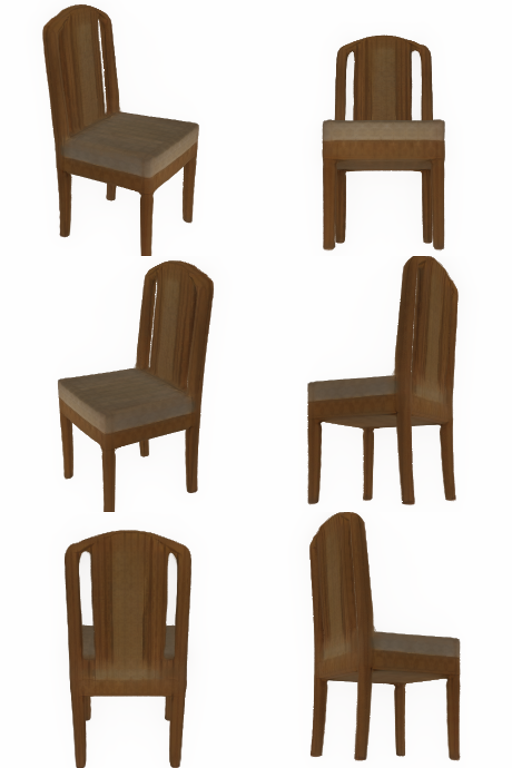}   
        \includegraphics[width=0.16\textwidth,viewport={0 0 512 512},clip]{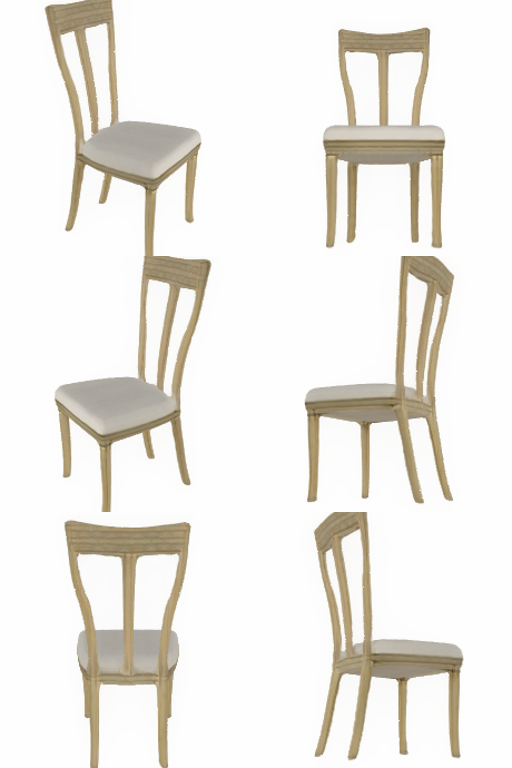}   & 
        \includegraphics[width=0.16\textwidth,viewport={4 4 512 512},clip]{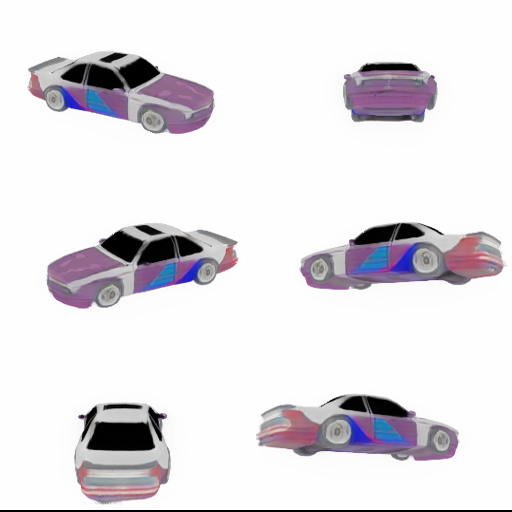} 
        \includegraphics[width=0.16\textwidth,viewport={4 4 512 512},clip]{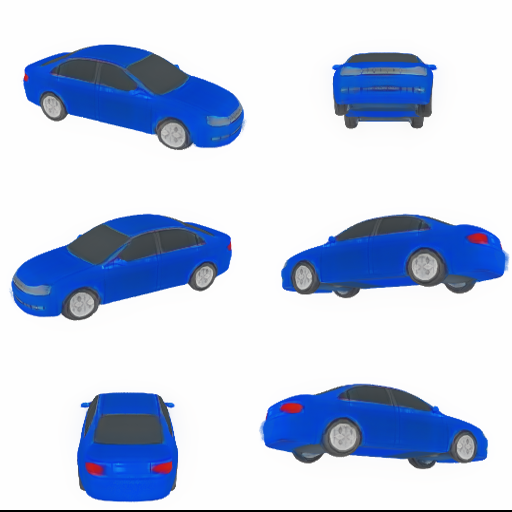}  &
        \includegraphics[width=0.16\textwidth,viewport={4 4 512 512},clip]{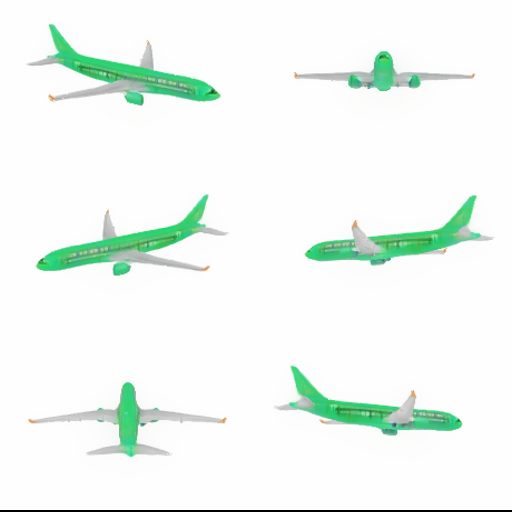} 
        \includegraphics[width=0.16\textwidth,viewport={4 4 512 512},clip]{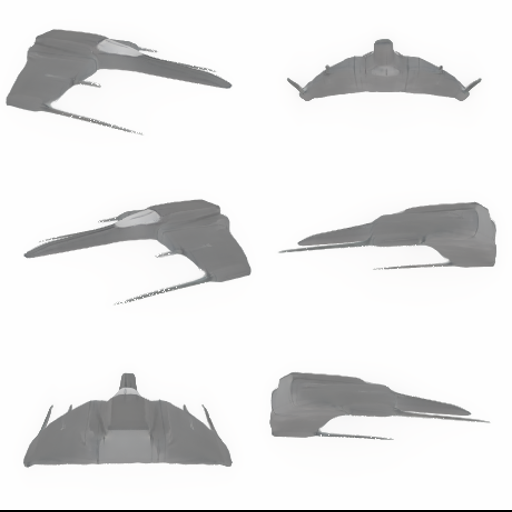}  
        \\
        
         &
         \small \shortstack{``a four legged \\ chair with cushion''} \shortstack{``A fancy dining\\ room chair with \\sculpted wooden legs''} &
         \small \shortstack{``BMW E30 \\ race car'' } \quad \quad  \quad``a blue car '' &
         \small \shortstack{``a green airplane \\  in the air''}  \quad \quad \shortstack{``a star wars \\ fighter jet''}
    
    \end{tabular}
    \caption{\textbf{Text-conditional generation on ShapeNet}. While  LN3Diff\cite{lan2025ln3diff}, a leading text-to-3D method, shows poor visual results when rendered from the side or bottom views, our results are of higher quality with fewer artifacts and adhere better to the input prompts unseen during training. Zoom in for the best view.}
    \vspace{-0.1cm}
    \label{fig:text_con_compare}
\end{figure*}
  

%% file: img/vs_ln3diff/main_fig2.tex
\begin{figure*}[htbp]
    \centering
    \begin{tabular}{c p{3.8cm} p{3.8cm} p{3.8cm} p{3.8cm}}
        \raisebox{0.5cm}{\rotatebox{90}{ LN3Diff \cite{lan2025ln3diff}}}  &
        \includegraphics[width=0.21\textwidth,viewport={0 0 512 512},clip]{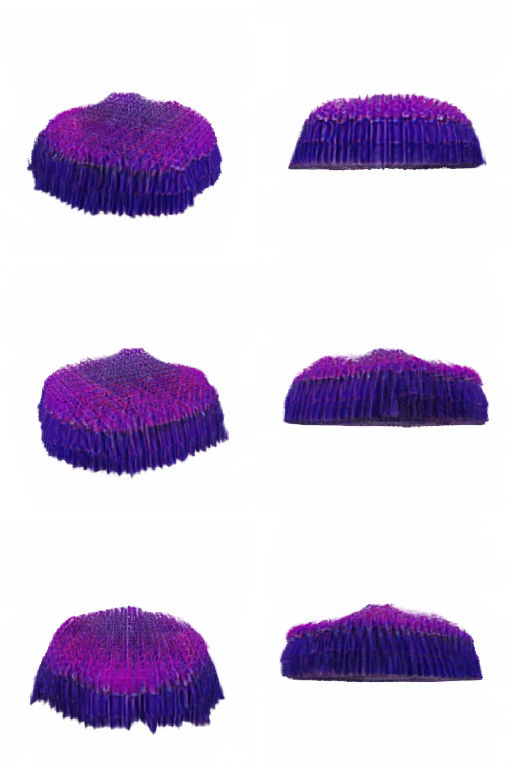}   &
        \includegraphics[width=0.21\textwidth,viewport={0 0 512 512},clip]{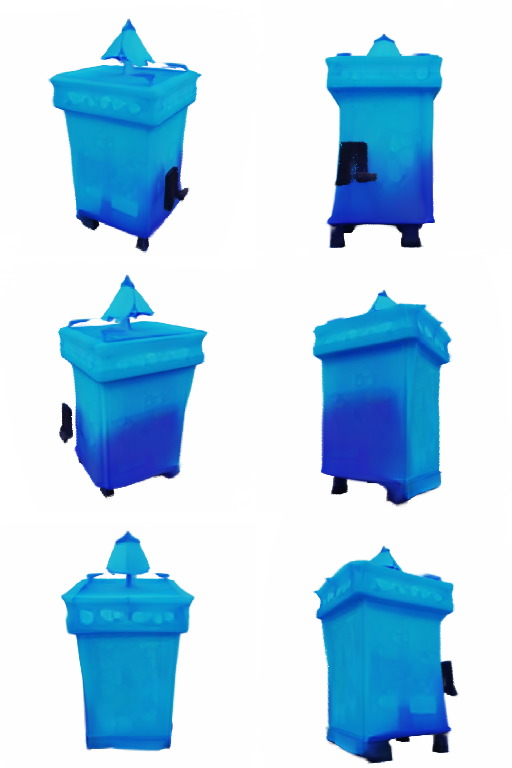}   &
        \includegraphics[width=0.21\textwidth,viewport={0 0 512 512},clip]{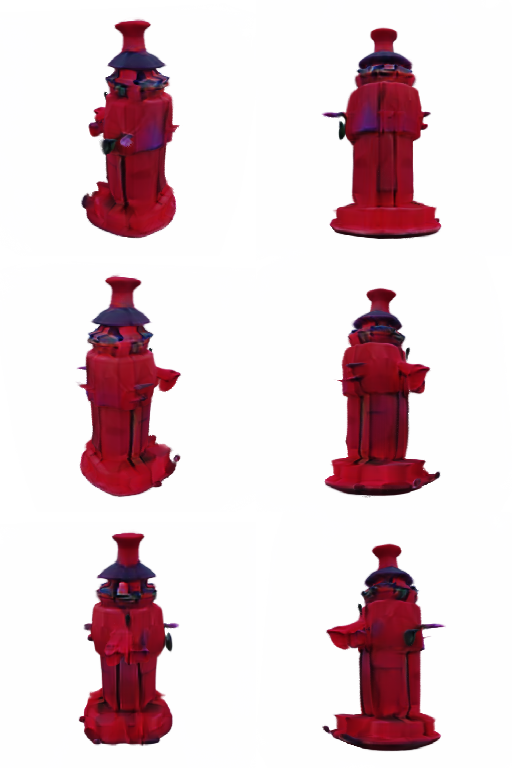}   &
        \includegraphics[width=0.21\textwidth,viewport={0 0 512 512},clip]{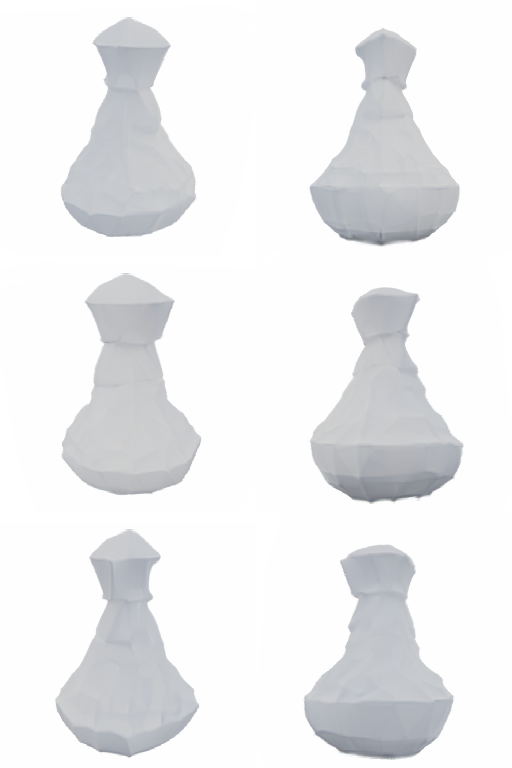}    \\
        
        \raisebox{0.5cm}{\rotatebox{90}{ \shortstack{ Direct3D \cite{liu2024direct} \\ (T=0.002)}}}  &
        \includegraphics[width=0.21\textwidth,viewport={0 0 350 350},clip]{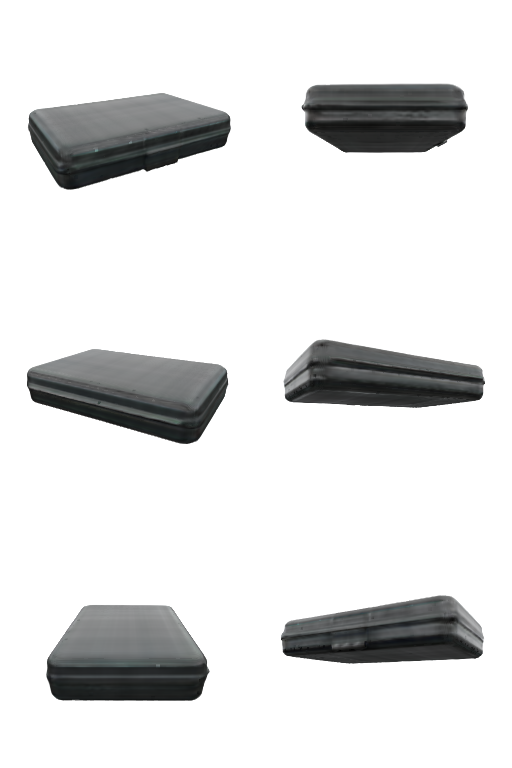}   & 
        \includegraphics[width=0.21\textwidth,viewport={0 0 350 350},clip]{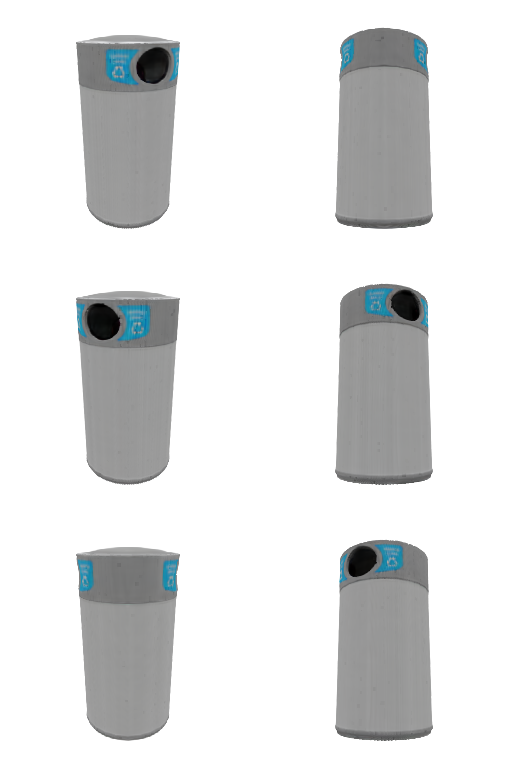}   &
        \includegraphics[width=0.21\textwidth,viewport={0 0 350 350},clip]{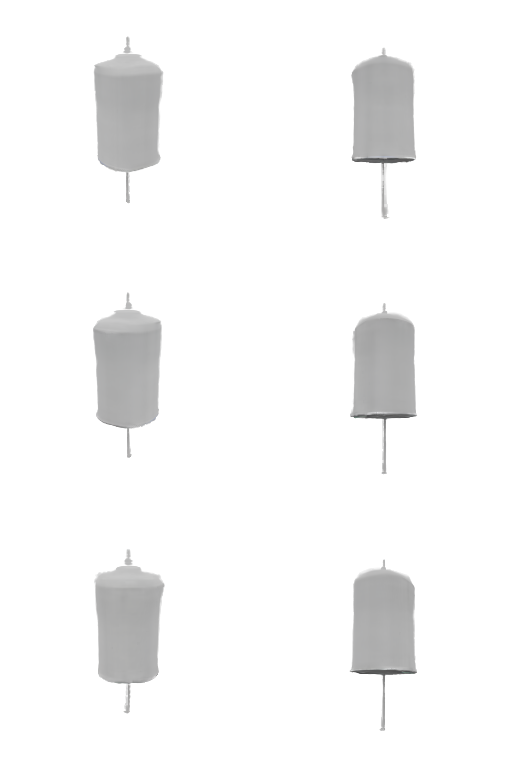}   &
        \includegraphics[width=0.21\textwidth,viewport={0 0 350 350},clip]{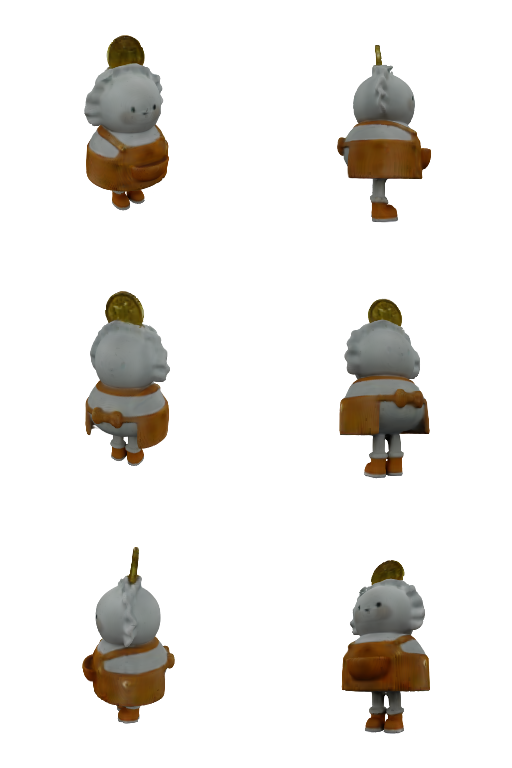}    
            \\
        \raisebox{0.5cm}{\rotatebox{90}{ \shortstack{ Direct3D \cite{liu2024direct}  \\(T=0.07)}}}  &
        \includegraphics[width=0.21\textwidth,viewport={0 0 350 350},clip]{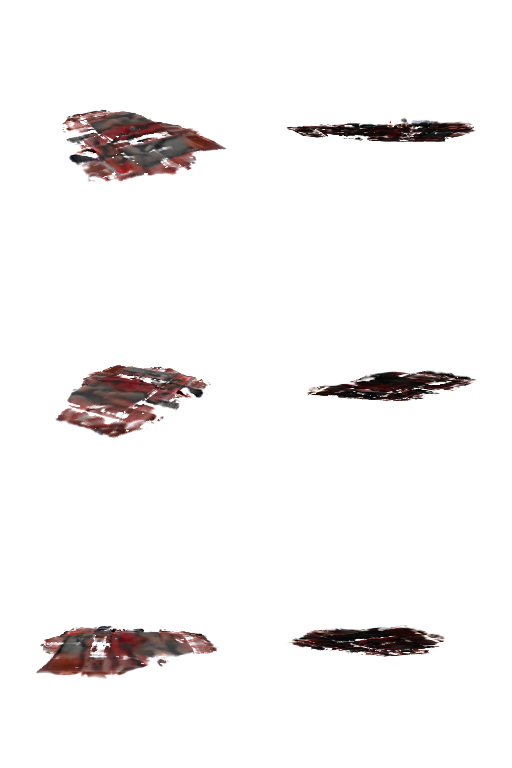}   & 
        \includegraphics[width=0.21\textwidth,viewport={0 0 350 350},clip]{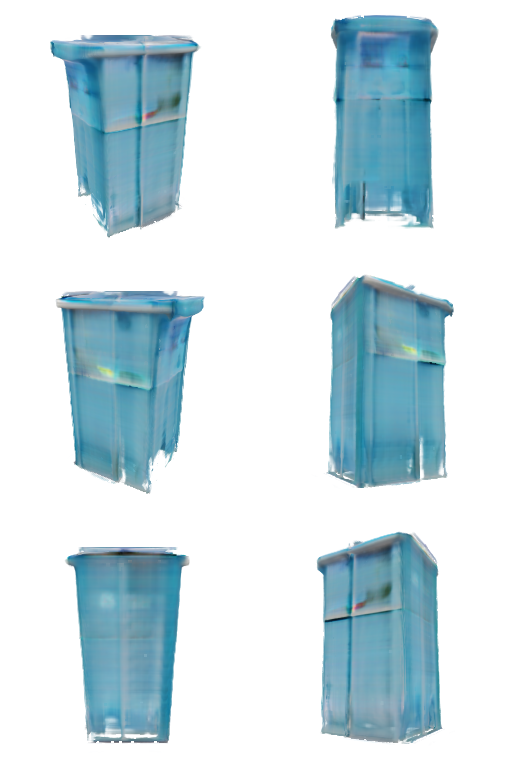}   &
        \includegraphics[width=0.21\textwidth,viewport={0 0 350 350},clip]{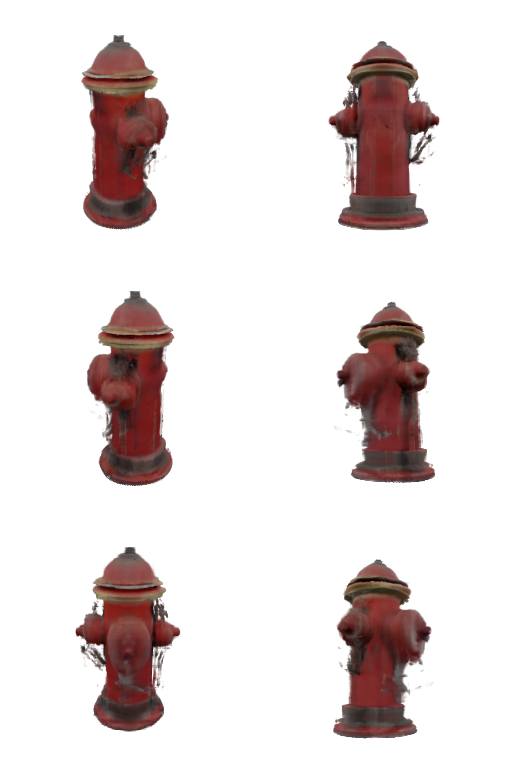}   &
        \includegraphics[width=0.21\textwidth,viewport={0 0 350 350},clip]{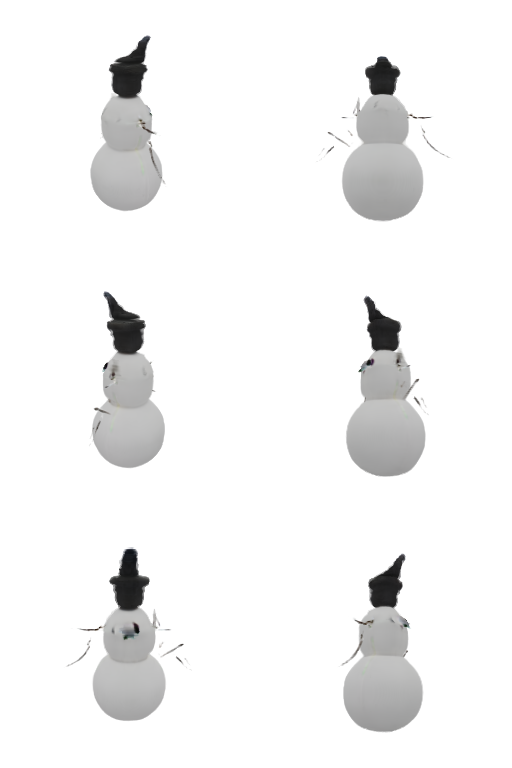}    
            \\

        \raisebox{1.5cm}{\rotatebox{90}{ \textbf{Ours}}}  &
        \includegraphics[width=0.21\textwidth,viewport={0 0 512 512},clip]{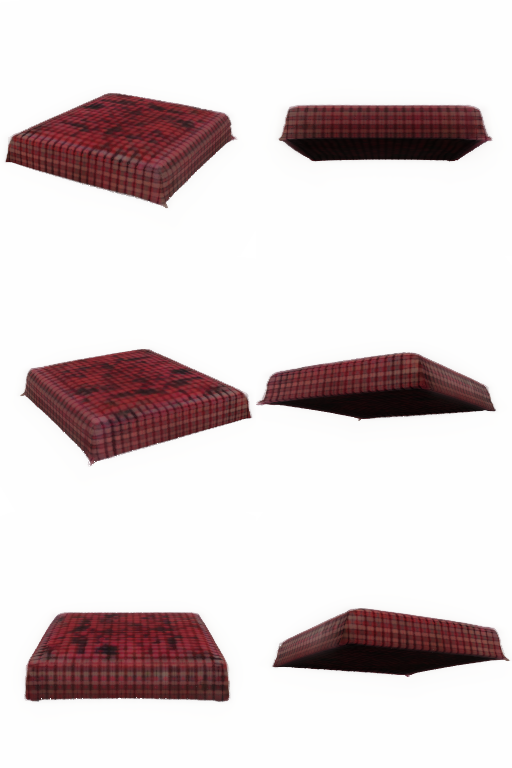}   &
        \includegraphics[width=0.21\textwidth,viewport={0 0 512 512},clip]{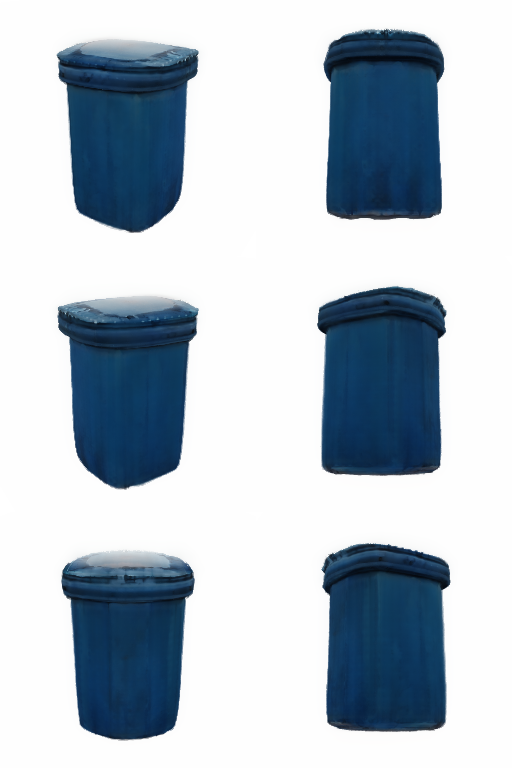}   &
        \includegraphics[width=0.21\textwidth,viewport={0 0 512 512},clip]{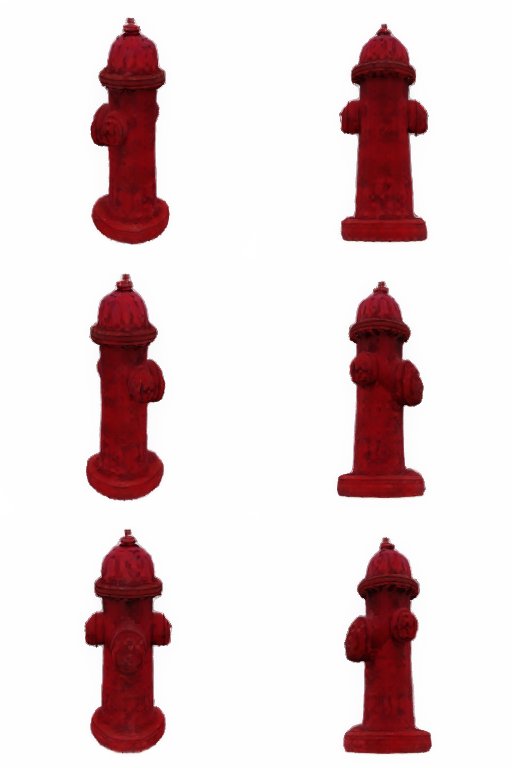}   &
        \includegraphics[width=0.21\textwidth,viewport={0 0 512 512},clip]{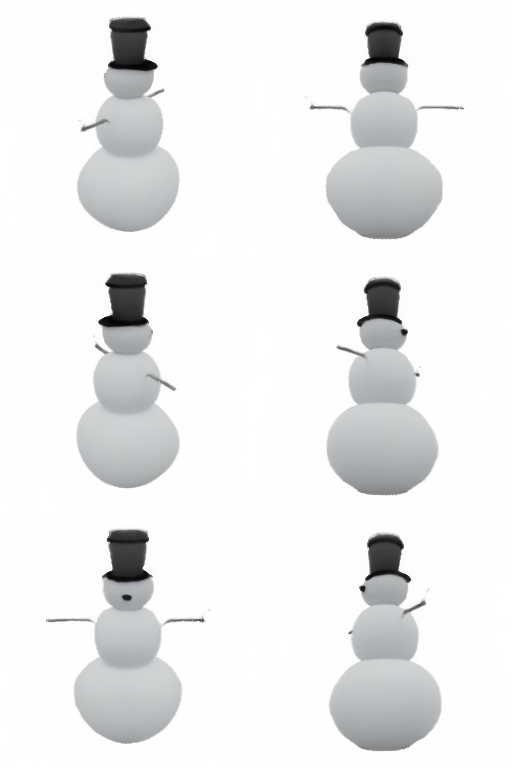}    \\

         & 
        ``A red and black checkered picnic blanket'' & 
        ``A small blue recycling bin'' & 
        ``A classic red fire hydrant'' & 
        ``A white snowman with a black top hat'' \\
    \end{tabular}
    \caption{\textbf{Qualitative comparisons on Objaverse}. Compared with other SOTA text conditional 3D generation methods, we demonstrate a better balance between realistic rendering and text alignment with fewer visual artifacts, evidenced by superior CLIP metrics (Tab.~\ref{tab:comparison}). Zoom in for the best view.}
    \label{fig:text_con_compare_main}
\end{figure*}

%% file: sec/5_conclusion.tex
\section{Dicussion}
\paragraph{Limitations.} While our current work does not scale beyond the 100K Objaverse LVIS dataset, the framework shows promise for extension to natural images through existing and future image-to-3D reconstruction models. This scalability to natural image domains represents an exciting direction for future research. Moreover, even with our architectural innovations, the current TriFlow network requires more than one week of training on Objaverse, indicating significant opportunities for further computational efficiency improvements, which we continue to explore.

\paragraph{Conclusion.} In this work, we show that directly training generative models on triplane features is not only feasible but also effective. By reusing powerful pre-trained reconstruction models, we avoid the computational burden of training custom encoders while leveraging rich 3D latent features for generative modeling. We propose well-motivated protocols for taming triplane latent features for generative modeling. We introduce TriFlow, a novel approach to text-conditioned 3D generation that uses a multi-stream transformer architecture that efficiently processes the triplane latent features. Our approach achieves high-quality, controllable 3D object generation, demonstrating superior performance in text alignment and rendering quality while addressing limitations in multi-view fidelity and generation robustness found in previous works. 

%% file: sec/X_suppl.tex
\clearpage
\setcounter{page}{1}
\maketitlesupplementary
\appendix

\section{Document Overview}
This material contains  \texttt{supplementary.pdf} document and \texttt{supplementary.html} webpage.
The webpage features video renderings of our generated objects, including the results presented in both main and supplementary documents. We strongly encourage readers to view the {\it offline} webpage in a browser.

In Sec.~\ref{supp1}, we provide implementation and experiment details that supplement the descriptions in the main paper, including the network architecture and training specifics. In Sec.~\ref{supp2}, we provide additional visual and numerical results of our method including further analysis of the text-controllability and generalizability behavior of the baseline generative models.

\section{Implementation and Experiment Details}\label{supp1}

\subsection{Model Architecture Design}

We provide further details of our TriFlow model architecture which is illustrated in Fig. \ref{fig:overview} of the main paper. We employ a multi-stream transformer architecture that minimizes computational complexity, particularly when handling large numbers of triplane tokens. This design builds upon the two-stream transformer approach first introduced in PointInfinity \cite{huang2024pointinfinity} and later validated in SF3D \cite{boss2024sf3d}.

Our architecture comprises three primary streams: a (1) \textbf{triplane stream} processing variable-sized triplane tokens, a (2) \textbf{latent stream} performing computations on fixed-size latent representations, and a (3) \textbf{text stream} providing conditional guidance. This multi-stream design achieves linear computational complexity with respect to triplane token size, avoiding the quadratic costs typically associated with self-attention operations on large token sets.
Each stream is initialized through fully connected (FC) layers. Within each transformer block set, the triplane stream first interacts with the latent stream through cross-attention in the initial Fuse block. Here, triplane tokens generate keys (K) and values (V), while latent tokens produce queries (Q). The resulting latent stream then processes this information through a sequence of three MM-DiT or DiT blocks, which handle the core computations.
The processed latent stream then feeds back to the triplane stream via cross-attention in the second Fuse block, where triplane tokens generate queries (Q) and latent tokens produce keys (K) and values (V). This bidirectional cross-attention pattern repeats across multiple block sets. The final outputs from both streams pass through a terminal FC layer to produce the model's denoising velocity prediction.

Our implementation incorporates both MM-DiT and DiT blocks within the multi-stream transformer block sets, adapted from the Flux implementation\footnote{https://github.com/black-forest-labs/flux/tree/main}. The architecture begins with a set of three MM-DiT blocks, followed by two sets of three DiT blocks each, all utilizing 768 hidden channels. This combination, demonstrated effective in Flux \cite{blackforestlabs2024} and AuraFlow \cite{cloneofsimo2024}, provides robust text alignment while maintaining computational efficiency. While Flux employs 2D rotary position embedding (RoPE) \cite{su2024roformer}, its effectiveness for 3D triplane data remains unverified. Following AuraFlow's approach, we instead implement Learnable Positional Embedding (PE), which better suits our triplane context. Our triplane tokens are generated by flattening the triplane data into 2$\times$2 patches before applying the learnable PE.

\subsection{A Tip for Reducing InstantMesh's Triplane Channel Dimension}

We propose a method to reduce the 80-dimensional channel of a triplane while preserving rendering quality.
This reduction leverages the rendering MLP decoder's structure. Let \((T_{xy}, T_{yz}, T_{xz})\) denote the triplane, and (W) represent the weight matrix of the MLP's first layer. The matrix (W) can be decomposed into \([W_{xy}, W_{yz}, W_{xz}]\), where each component corresponds to features from its respective triplane projection.

By applying (W) to its corresponding triplane components, we obtain a compressed representation \((W_{xy} T_{xy}, W_{yz} T_{yz}, W_{xz} T_{xz})\) with 64 dimensions. This reduced-dimensional space can serve as the latent representation for noise addition and model training. The original 80-dimensional space can be reconstructed using pseudo-inverse multiplication.
While this process yields triplanes that differ from the original representation, both versions produce identical outputs after passing through the rendering MLP.

\subsection{ShapeNet Unconditional Training Details}

For unconditional ShapeNet generation experiments, we trained our models from scratch using the base architecture on 6 NVIDIA L40S GPUs with a batch size of 132. Training utilized the AdamW optimizer~\cite{loshchilov2017decoupled} with a learning rate of \(5 \times 10^{-5}\).
We conducted experiments on the ShapeNet Car and ShapeNet Chair datasets, comprising 3,514 and 6,778 instances respectively. The dataset was prepared by randomly sampling 6 views per object from 36 camera positions on the upper hemisphere. These views were rendered and processed through InstantMesh to generate triplane representations.
The triplane features were reduced from 80 to 64 dimensions using a rendering MLP, followed by standardization as detailed in Section~\ref{sec:standard}. We then applied center cropping to reduce the spatial resolution from \(64 \times 64\) to \(48 \times 48\), removing only regions corresponding to empty space in the final render. The resulting triplane representations had dimensions of \(3 \times 64 \times 48 \times 48\).
Following LN3Diff's~\cite{lan2025ln3diff} evaluation protocol, we computed Fréchet Inception Distance (FID) and Kernel Inception Distance (KID) scores using 50K generated samples for 2D evaluation. Additional results are presented in Fig.~\ref{fig:uncond_figs}.

\subsection{Camera Viewpoint Distributions}

To create a dataset of triplanes for Objaverse and ShapeNet, we rendered objects using six default input camera views tailored for InstantMesh, using the same settings as Zero123++. The elevation angles were set to \(\{20, -10, 20, -10, 20, -10\}\), the azimuth angles to \(\{30, 90, 150, 210, 270, 330\}\), and the radius to 3.5.  

For experiments comparing with LN3Diff~\cite{lan2025ln3diff} (Tables \ref{tab:comparison} and \ref{tab:uncond_comp}), we used the rendered images provided by LN3Diff to ensure a fair comparison. These images included 50 views sampled from the upper hemisphere with a radius of 1.2. To focus on views that captured more geometric detail, we filtered out poses with elevation angles greater than \(45^\circ\).  
To reduce computational costs, we randomly selected half of the remaining poses. From these, we evaluated all possible combinations of six poses by projecting them onto the equatorial plane. We then selected the combination that maximized the minimum angular distance between the poses for use in our experiments.  

\section{Additional Experiment Results}\label{supp2}

\subsection{Full Quantitative Results on Text-Conditional Generation on ShapeNet} 
We present an extended version of Tab.~\ref{tab:comparison} from the main paper in Tab.~\ref{tab:more_comparison}. We evaluate our method's text-conditioned generation capabilities on the ShapeNet dataset, comparing against LN3Diff \cite{lan2025ln3diff} across cars, planes, and chairs categories using multiple evaluation metrics.

FID and sFID assess the similarity between generated and real sample distributions, with lower values indicating higher realism. While FID evaluates overall similarity in Inception-V3 feature space, sFID focuses on spatial relationships, favoring structurally coherent samples. CMMD, introduced by Jayasumana et al. \cite{jayasumana2024rethinking}, measures distribution similarity in CLIP ViT-L feature space, aligning with human perception and following the same lower-is-better interpretation as FID metrics.

The Inception Score (IS) evaluates both sample diversity and quality, with higher values indicating more visually distinct and classifiable outputs. Improved Precision/Recall metrics \cite{kynkaanniemi2019improved} measure fidelity and diversity respectively, where higher precision indicates better sample fidelity (more model samples within the data manifold), and higher recall reflects better diversity (more real samples within the sample manifold). We also include CLIP Scores \cite{dhariwal2021diffusion} for VIT-B/32 and VIT-L/14 models, where higher scores indicate better text-image alignment.
Our evaluation uses OpenAI's official implementation\footnote{https://github.com/openai/guided-diffusion/tree/main/evaluations} for FID, sFID, IS, Precision, and Recall metrics, and a PyTorch implementation\footnote{https://github.com/sayakpaul/cmmd-pytorch} for CMMD calculations.

Table \ref{tab:more_comparison} demonstrates our method's superior performance compared to LN3Diff \cite{lan2025ln3diff}, the state-of-the-art text-to-3D generation, across most metrics. The lower FID, sFID, and CMMD values indicate enhanced realism, spatial coherence, and perceptual quality. Our competitive IS scores and improved Precision/Recall metrics suggest better sample quality and diversity. Furthermore, higher CLIP scores confirm stronger alignment between generated shapes and input text prompts, validating our method's effectiveness in text-guided 3D shape generation.

\begin{table*}[h]
\centering
\begin{tabular}{ll|cccccccc}
Dataset& Methods & FID$\downarrow$ & sFID$\downarrow$ & {CMMD}$\downarrow$ & {IS}$\downarrow$ & {Pre}$\uparrow$ & {Recall}$\uparrow$ & \shortstack{CLIP score\\VIT-B/32 }$\uparrow$ & \shortstack{CLIP score\\VIT-L/14 }$\uparrow$  \\
\midrule
\multirow{2}{*}{car} & LN3Diff\cite{lan2025ln3diff}  & 45.57 & 45.57 & 0.606 & 4.071 & 0.2981 & 0.1285 & \textbf{30.60} & 25.21 \\
 & Ours & \textbf{15.65} & \textbf{19.99} & \textbf{0.116} & \textbf{3.967} & \textbf{0.9200} & \textbf{0.1718} & \textbf{30.60} & \textbf{25.56} \\
\midrule
\multirow{2}{*}{plane}& LN3Diff\cite{lan2025ln3diff}  & 44.98 & 41.42 & 0.438 & \textbf{2.672} & 0.3264 & \textbf{0.3147} & 27.78 & 23.21 \\
& Ours  & \textbf{12.92} & \textbf{11.92} & \textbf{0.087} & {2.845} & \textbf{0.8442} & 0.2061 & \textbf{29.03} & \textbf{24.70} \\
\midrule
\multirow{2}{*}{chair}& LN3Diff\cite{lan2025ln3diff}  & 42.52 & 52.37 & 0.602 & 4.525 & 0.2544 & 0.1793 & 28.21 & 21.71 \\
& Ours  & \textbf{10.95} & \textbf{17.57} & \textbf{0.086} & \textbf{3.748} & \textbf{0.7961} & \textbf{0.2293} & \textbf{28.25} & \textbf{22.41} \\
\end{tabular}
\caption{\textbf{Text-conditioned generation on ShapeNet comparisons}. This table is an extended version of Tab. \ref{tab:comparison} from the main paper, comparing our method against the state-of-the-art method on text-to-3D generation on ShapeNet, LN3Diff.}
\label{tab:more_comparison}
\end{table*}

\subsection{Additional Qualitative Results on Text-Conditional Generation}
In Fig. \ref{fig:sup_result}, we showcase a variety of outputs generated from diverse text prompts, highlighting the model's strong semantic understanding and ability to create visually accurate representations across multiple domains.

\paragraph{Stress testing with color} In Fig. \ref{fig:toilet}, we evaluate our model's text-to-3D triplane generation capabilities against state-of-the-art methods. Our test examines the model's ability to generate identical objects with unconventional color variations.

We conduct a systematic test generating toilets in five distinct colors: purple, blue, green, yellow, and orange. This experiment evaluates both color generalization capabilities and structural preservation during color transformations. Moreover, this test measures how well each model can be controlled in detail with text prompts.

Our model demonstrates robust prompt understanding and successful generalization for different colors, even for the unconventional object-color combinations. In comparison, LN3Diff \cite{lan2025ln3diff} shows limited success, exhibiting some color understanding but failing to consistently generate correctly colored objects.
Direct3D \cite{liu2024direct} shows varying performance across different temperature settings. At T=0.002, it struggles with both color and object representation, often producing outputs misaligned with text prompts. At T=0.07, it achieves marginal improvement, successfully generating yellow toilets but failing to generalize effectively across other color combinations.


\input{img/main2/sub_fig}
\input{img/toilet/fig}

 \input{img/uncond_figs/uncond_figs}

%% file: img/main2/sub_fig.tex
\begin{figure*}[htbp]
    \centering
    \begin{tabular}{ p{5.5cm} p{5.5cm} p{5.5cm}}
        \includegraphics[width=0.07\textwidth]{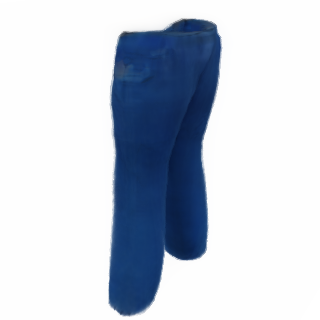} 
        \includegraphics[width=0.07\textwidth]{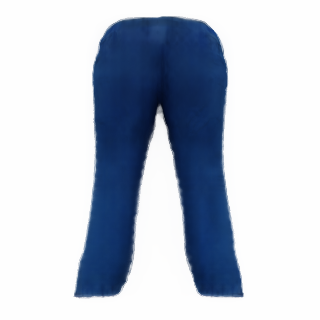}
        \includegraphics[width=0.07\textwidth]{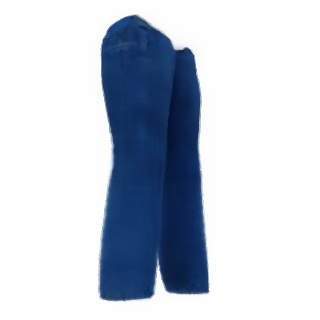}
        \includegraphics[width=0.07\textwidth]{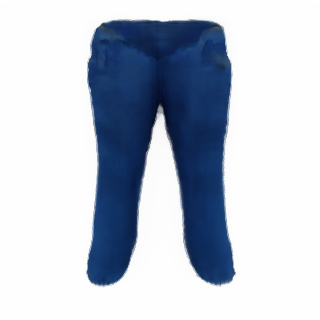} &
        \includegraphics[width=0.07\textwidth]{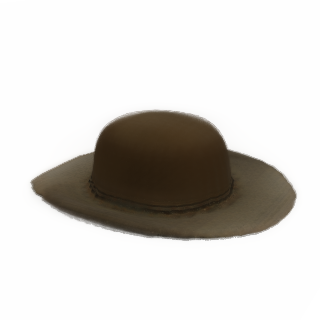} 
        \includegraphics[width=0.07\textwidth]{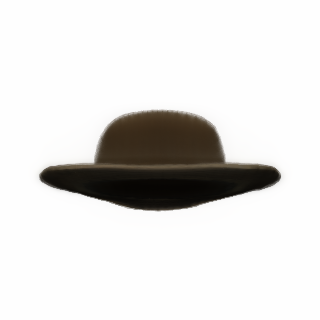} 
        \includegraphics[width=0.07\textwidth]{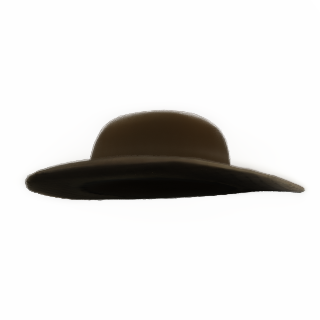} 
        \includegraphics[width=0.07\textwidth]{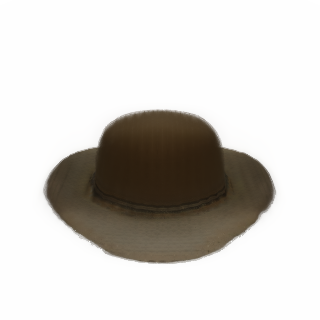}  &
        \includegraphics[width=0.07\textwidth]{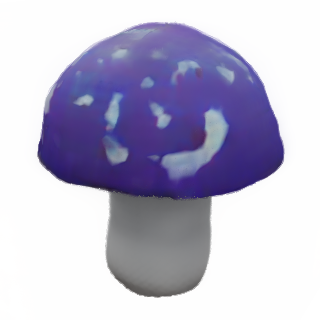} 
        \includegraphics[width=0.07\textwidth]{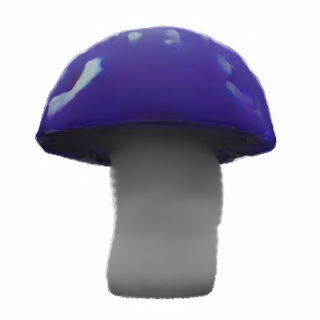}
        \includegraphics[width=0.07\textwidth]{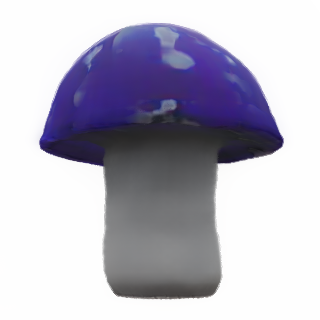}
        \includegraphics[width=0.07\textwidth]{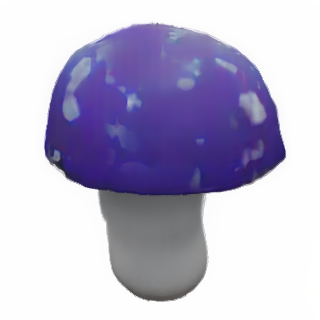} \\
        ``a pair of comfortable blue jeans'' & ``a cowboy hat'' & ``a purple cap mushroom''\\
        
        \includegraphics[width=0.07\textwidth]{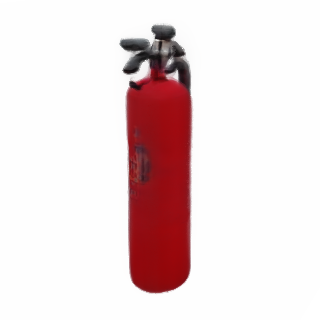} 
        \includegraphics[width=0.07\textwidth]{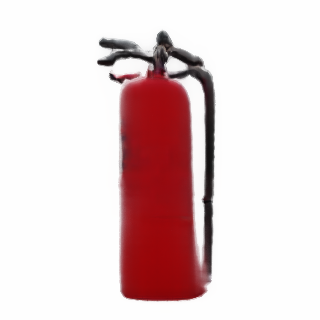}
        \includegraphics[width=0.07\textwidth]{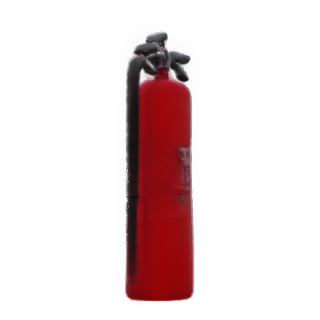}
        \includegraphics[width=0.07\textwidth]{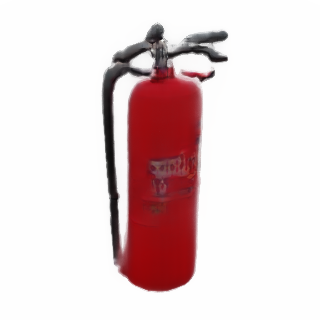} &
        \includegraphics[width=0.07\textwidth]{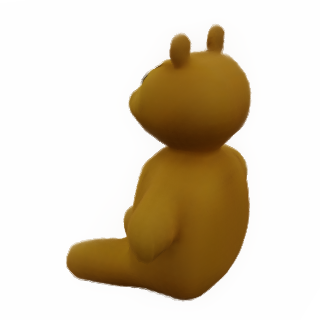} 
        \includegraphics[width=0.07\textwidth]{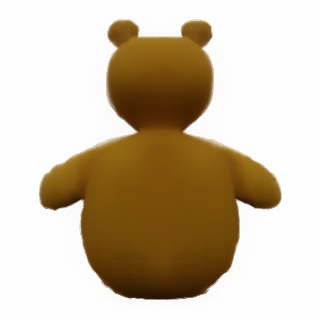} 
        \includegraphics[width=0.07\textwidth]{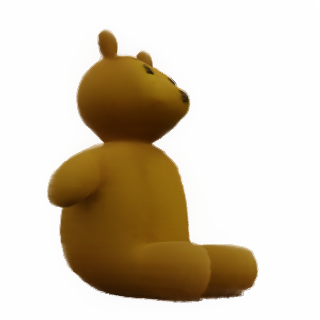} 
        \includegraphics[width=0.07\textwidth]{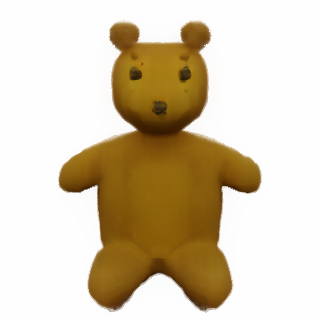}  &
        \includegraphics[width=0.07\textwidth]{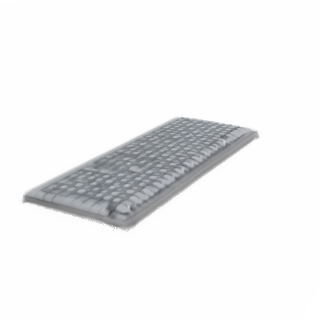} 
        \includegraphics[width=0.07\textwidth]{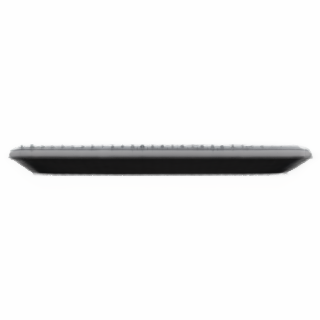}
        \includegraphics[width=0.07\textwidth]{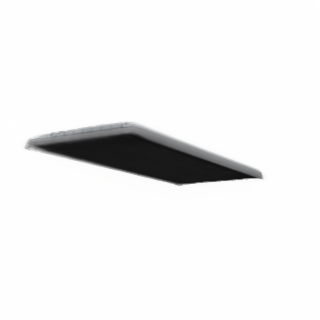}
        \includegraphics[width=0.07\textwidth]{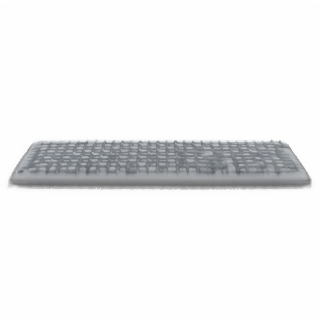} \\
        ``a red fire extinguisher with black handle'' & ``a yellow and brown teddy bear'' & ``a computer keyboard''\\
        
        \includegraphics[width=0.07\textwidth]{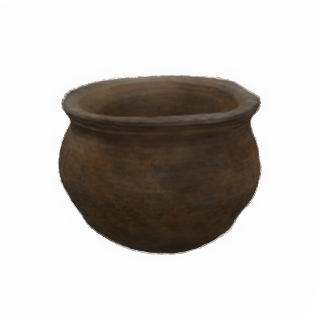} 
        \includegraphics[width=0.07\textwidth]{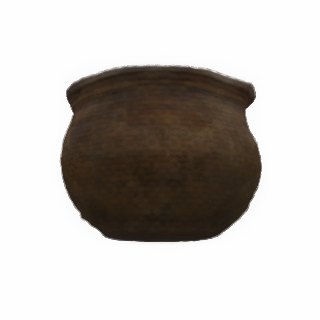}
        \includegraphics[width=0.07\textwidth]{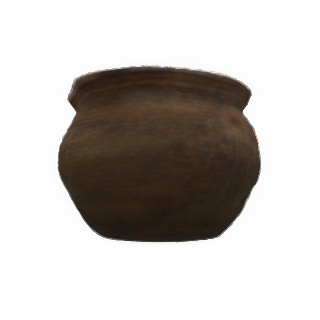}
        \includegraphics[width=0.07\textwidth]{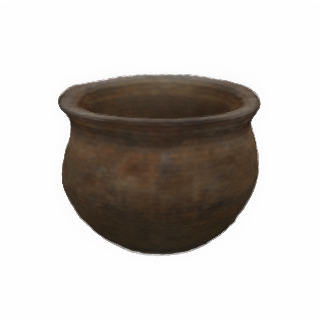} &
        \includegraphics[width=0.07\textwidth]{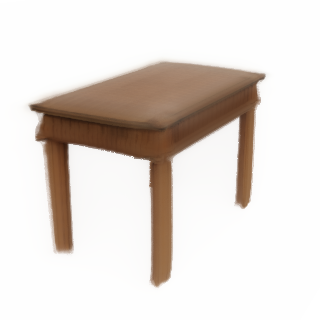} 
        \includegraphics[width=0.07\textwidth]{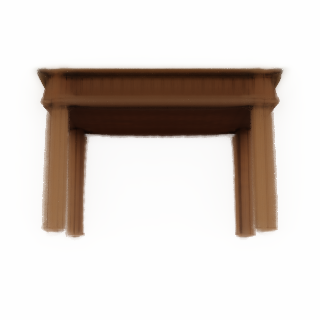} 
        \includegraphics[width=0.07\textwidth]{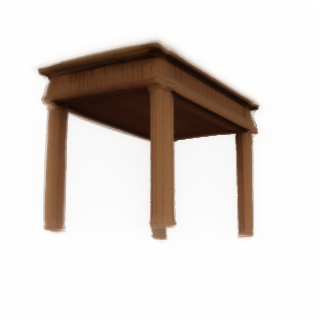} 
        \includegraphics[width=0.07\textwidth]{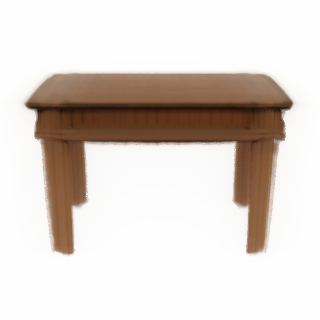}  &
        \includegraphics[width=0.07\textwidth]{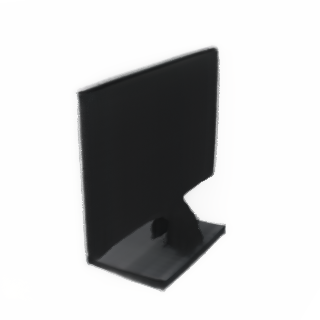} 
        \includegraphics[width=0.07\textwidth]{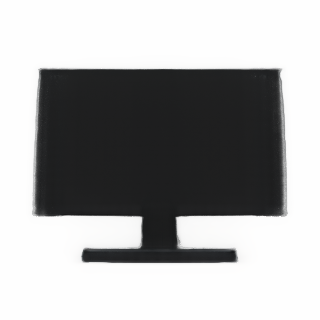}
        \includegraphics[width=0.07\textwidth]{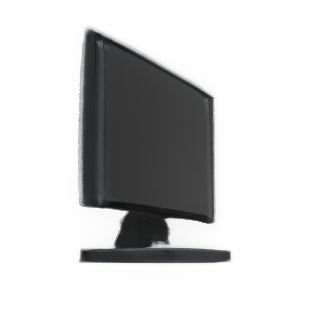}
        \includegraphics[width=0.07\textwidth]{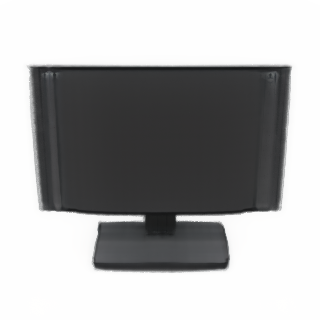} \\
        ``an ancient clay pot'' & ``a wooden desk'' & ``a computer mornitor with a flat screen''\\
        
        \includegraphics[width=0.07\textwidth]{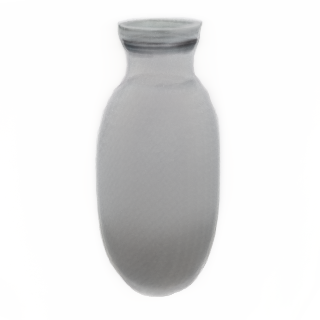} 
        \includegraphics[width=0.07\textwidth]{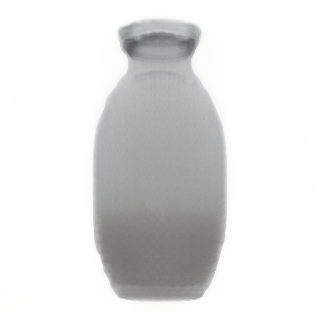}
        \includegraphics[width=0.07\textwidth]{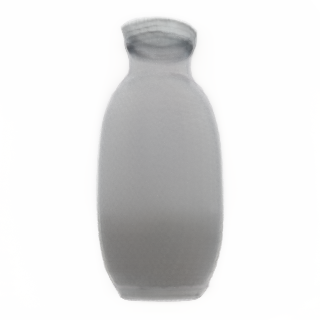}
        \includegraphics[width=0.07\textwidth]{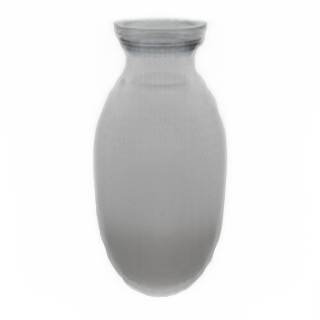} &
        \includegraphics[width=0.07\textwidth]{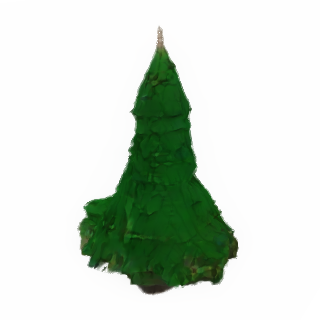} 
        \includegraphics[width=0.07\textwidth]{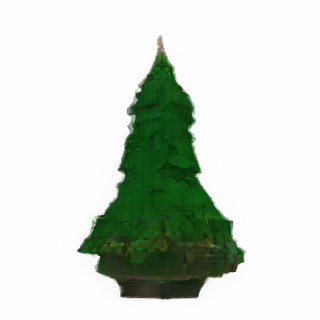} 
        \includegraphics[width=0.07\textwidth]{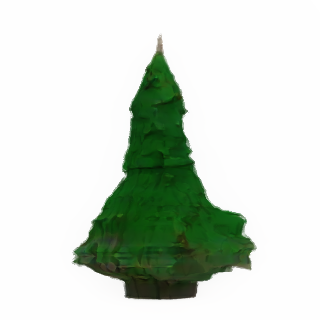} 
        \includegraphics[width=0.07\textwidth]{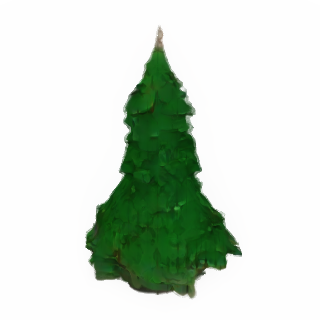}  &
        \includegraphics[width=0.07\textwidth]{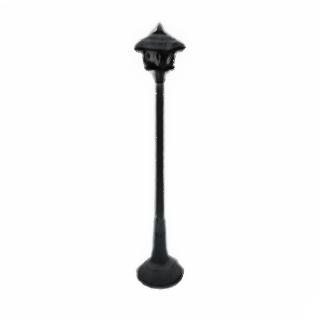} 
        \includegraphics[width=0.07\textwidth]{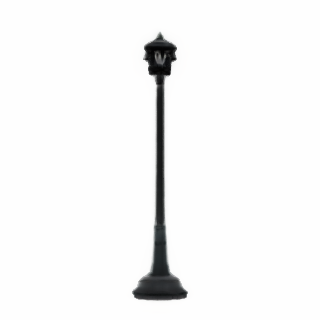}
        \includegraphics[width=0.07\textwidth]{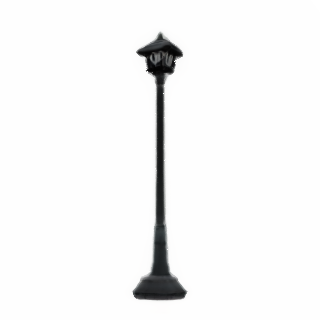}
        \includegraphics[width=0.07\textwidth]{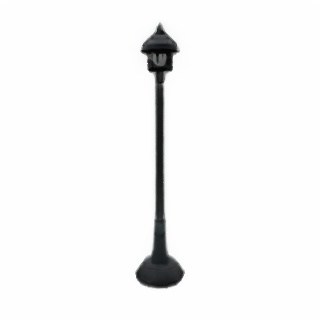} \\
        ``a glass jar'' & ``a big green Christmas tree'' & ``a metal lampost with a lantern on the top''\\
        
        \includegraphics[width=0.07\textwidth]{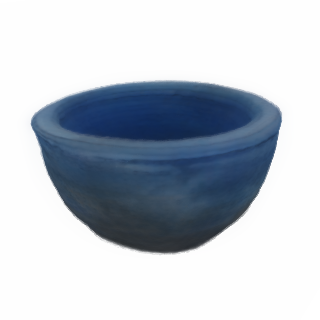} 
        \includegraphics[width=0.07\textwidth]{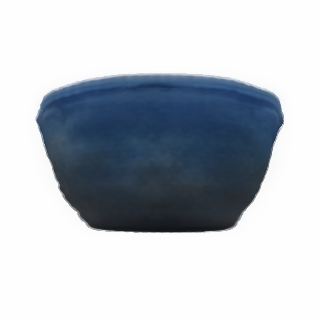}
        \includegraphics[width=0.07\textwidth]{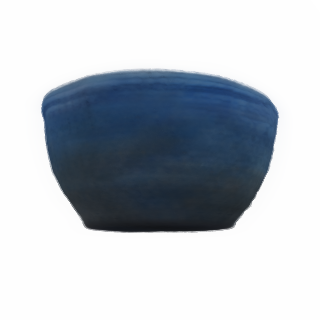}
        \includegraphics[width=0.07\textwidth]{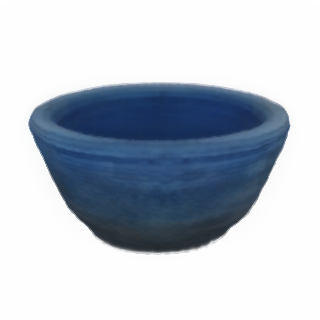} &
        \includegraphics[width=0.07\textwidth]{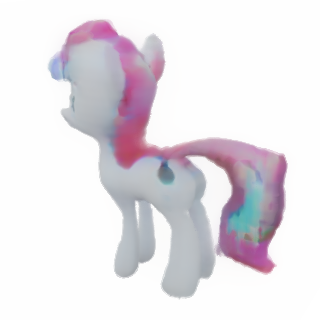} 
        \includegraphics[width=0.07\textwidth]{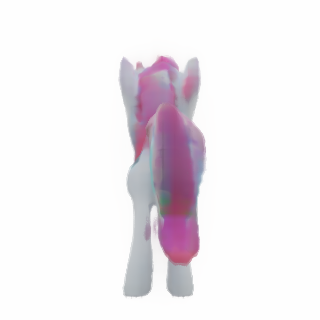} 
        \includegraphics[width=0.07\textwidth]{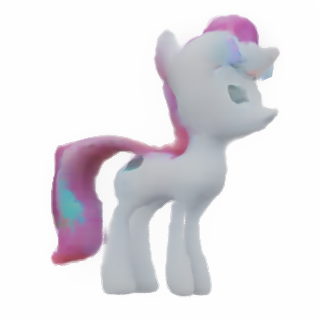} 
        \includegraphics[width=0.07\textwidth]{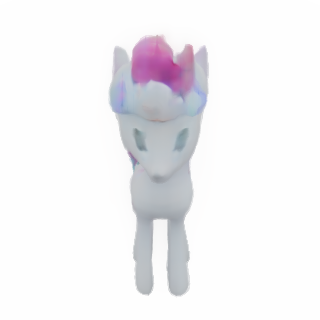}  &
        \includegraphics[width=0.07\textwidth]{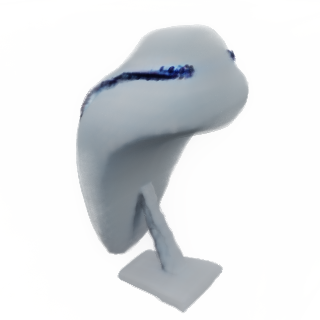} 
        \includegraphics[width=0.07\textwidth]{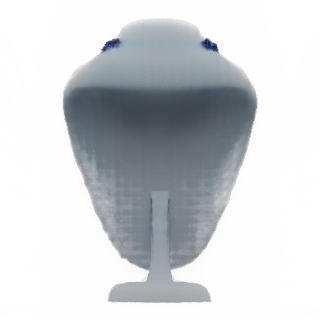}
        \includegraphics[width=0.07\textwidth]{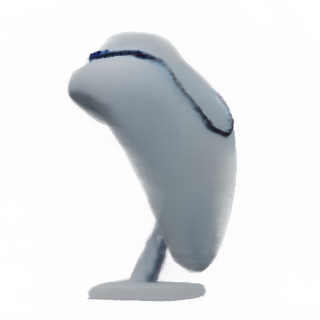}
        \includegraphics[width=0.07\textwidth]{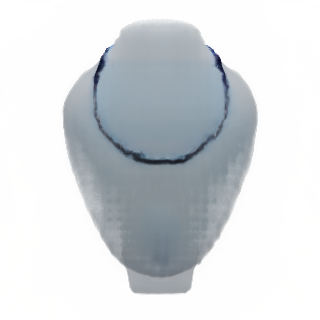} \\
        ``a blue ceramic bowl'' & ``My Little Pony'' & ``a blue jewelry necklace''\\
        
        \includegraphics[width=0.07\textwidth]{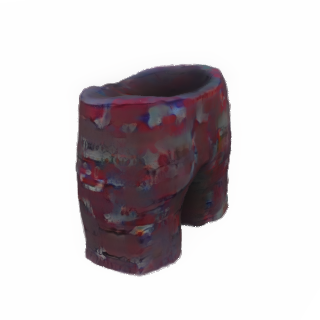} 
        \includegraphics[width=0.07\textwidth]{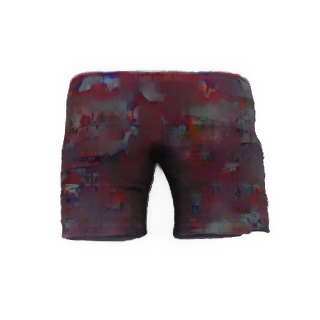}
        \includegraphics[width=0.07\textwidth]{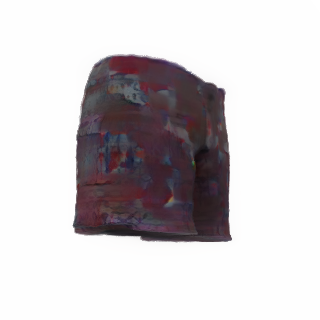}
        \includegraphics[width=0.07\textwidth]{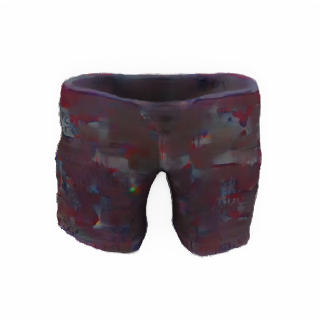} &
        \includegraphics[width=0.07\textwidth]{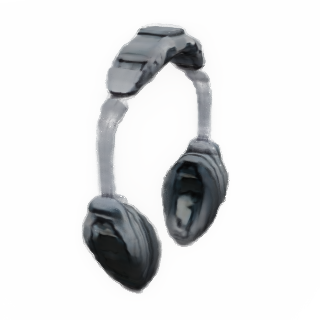} 
        \includegraphics[width=0.07\textwidth]{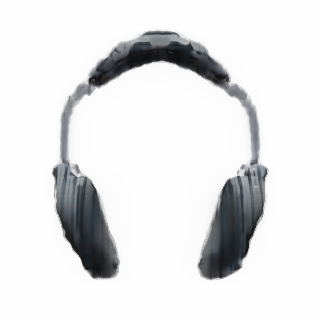} 
        \includegraphics[width=0.07\textwidth]{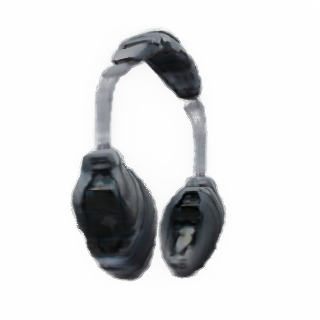} 
        \includegraphics[width=0.07\textwidth]{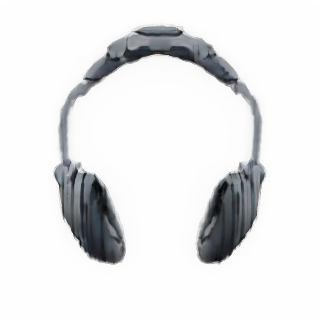}  &
        \includegraphics[width=0.07\textwidth]{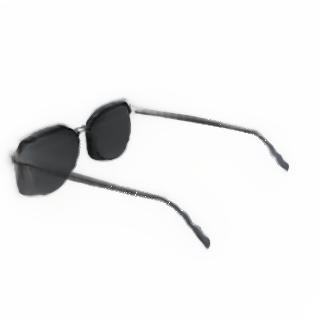} 
        \includegraphics[width=0.07\textwidth]{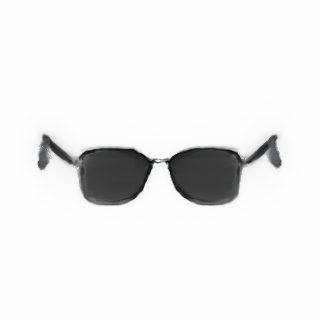}
        \includegraphics[width=0.07\textwidth]{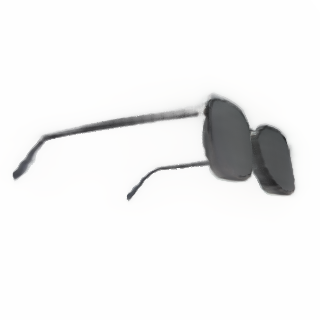}
        \includegraphics[width=0.07\textwidth]{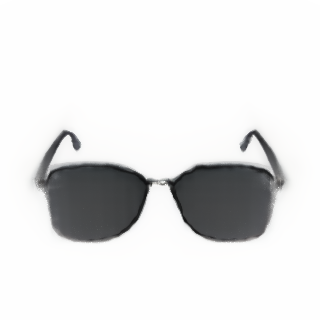} \\
        ``a pair of boxer shorts with color pattern'' & ``a pair of headphones with modern design'' & ``a pair of black sunglasses''\\
        
        \includegraphics[width=0.07\textwidth]{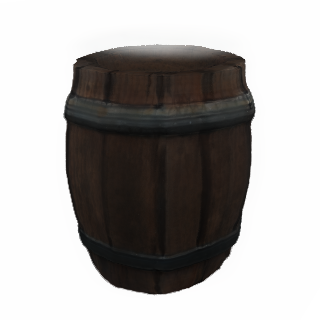} 
        \includegraphics[width=0.07\textwidth]{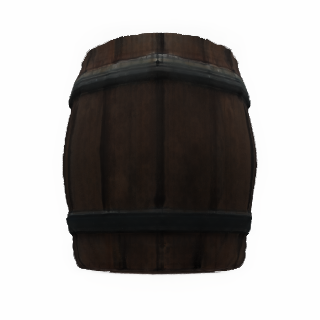}
        \includegraphics[width=0.07\textwidth]{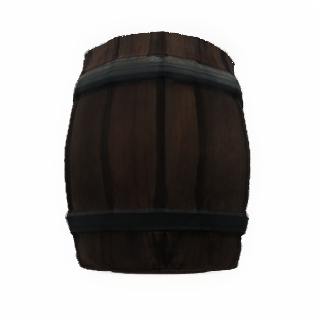}
        \includegraphics[width=0.07\textwidth]{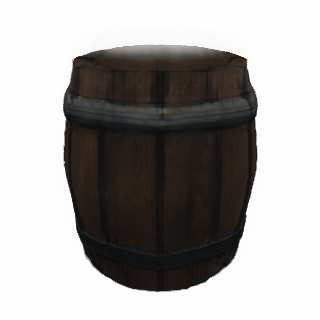} &
        \includegraphics[width=0.07\textwidth]{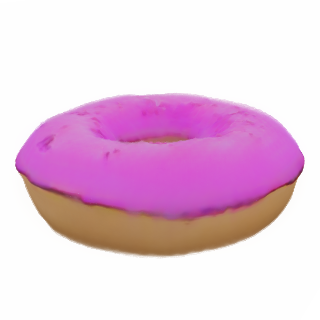} 
        \includegraphics[width=0.07\textwidth]{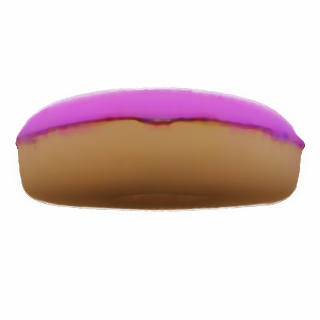} 
        \includegraphics[width=0.07\textwidth]{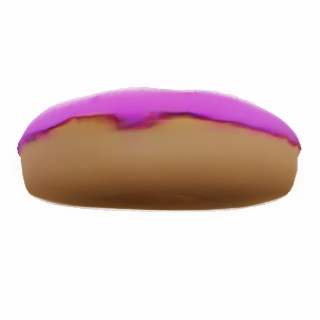} 
        \includegraphics[width=0.07\textwidth]{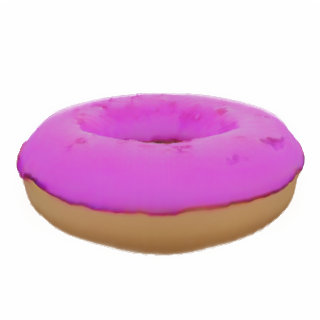}  &
        \includegraphics[width=0.07\textwidth]{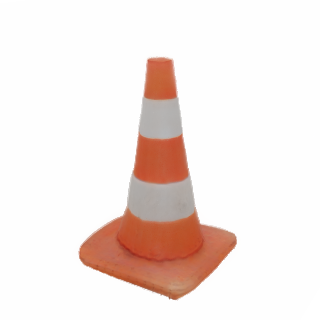} 
        \includegraphics[width=0.07\textwidth]{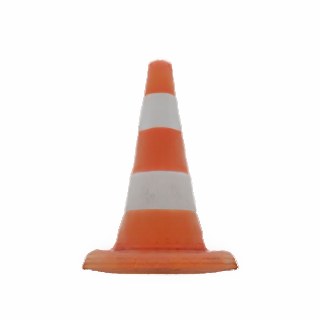}
        \includegraphics[width=0.07\textwidth]{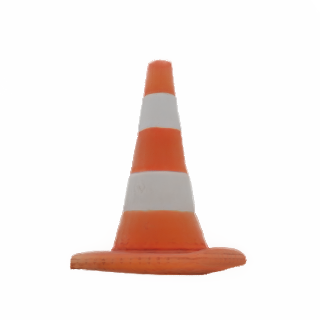}
        \includegraphics[width=0.07\textwidth]{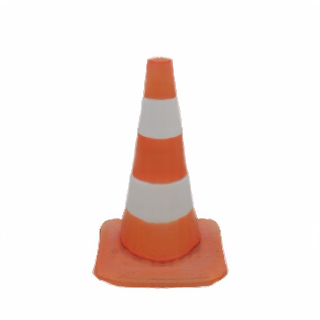} \\
        ``a rustic wooden barrel'' & ``a pink cream doughnut'' & ``a white and orange traffic cone''\\
        
        \includegraphics[width=0.07\textwidth]{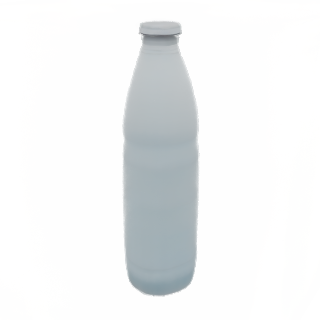} 
        \includegraphics[width=0.07\textwidth]{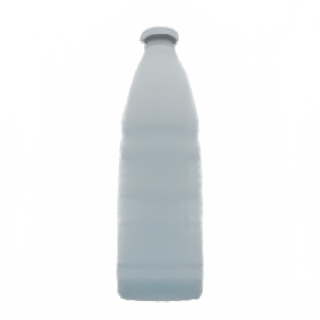}
        \includegraphics[width=0.07\textwidth]{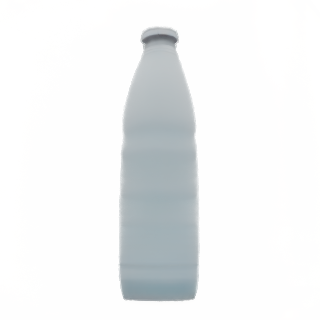}
        \includegraphics[width=0.07\textwidth]{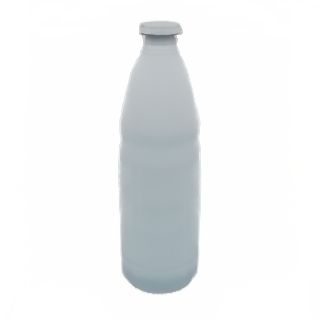} &
        \includegraphics[width=0.07\textwidth]{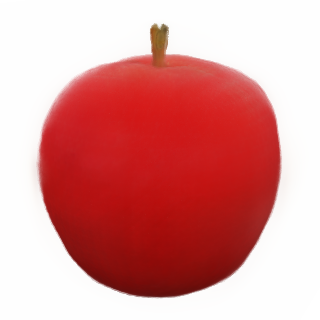} 
        \includegraphics[width=0.07\textwidth]{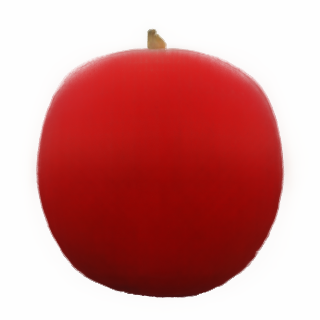} 
        \includegraphics[width=0.07\textwidth]{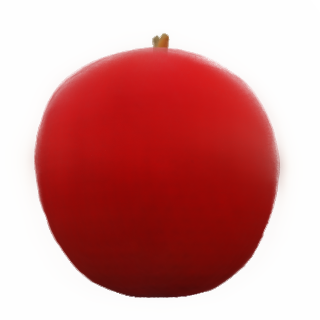} 
        \includegraphics[width=0.07\textwidth]{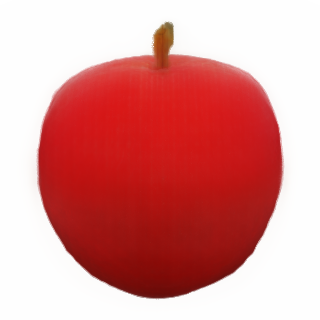}  &
        \includegraphics[width=0.07\textwidth]{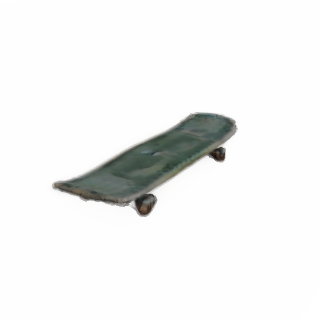} 
        \includegraphics[width=0.07\textwidth]{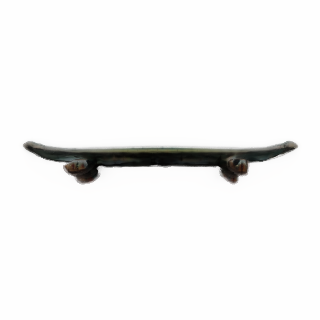}
        \includegraphics[width=0.07\textwidth]{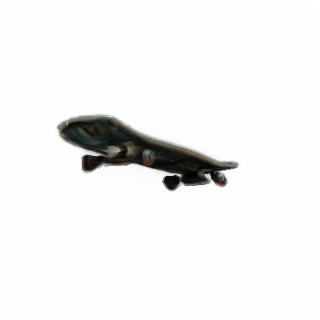}
        \includegraphics[width=0.07\textwidth]{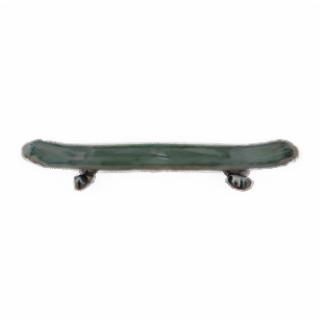} \\
        ``a plastic water bottle'' & ``a red and pink apple'' & ``a skateboard with green and white pattern design''\\
        
        \includegraphics[width=0.07\textwidth]{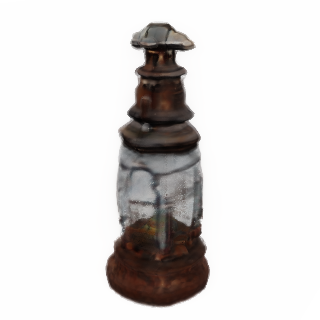} 
        \includegraphics[width=0.07\textwidth]{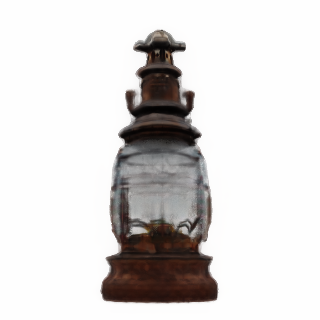}
        \includegraphics[width=0.07\textwidth]{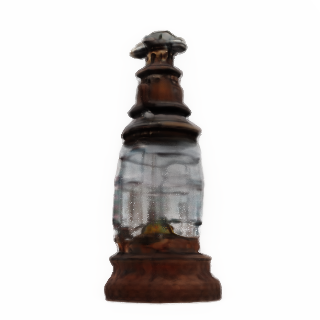}
        \includegraphics[width=0.07\textwidth]{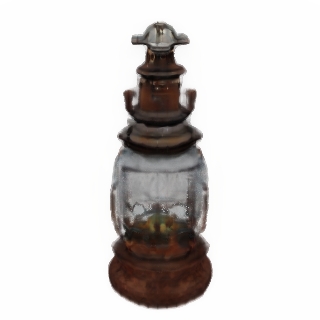} &
        \includegraphics[width=0.07\textwidth]{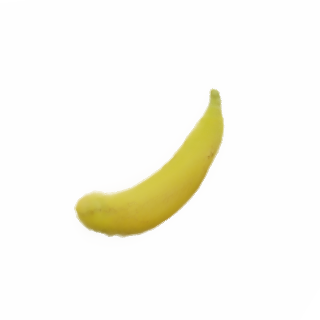} 
        \includegraphics[width=0.07\textwidth]{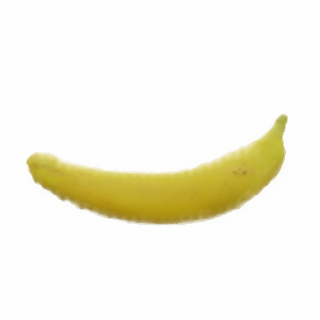} 
        \includegraphics[width=0.07\textwidth]{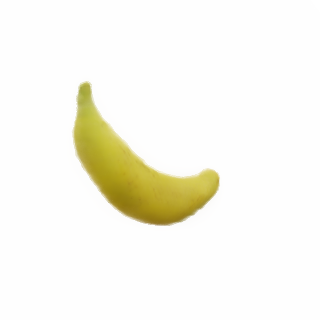} 
        \includegraphics[width=0.07\textwidth]{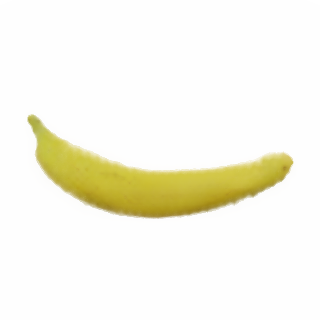}  &
        \includegraphics[width=0.07\textwidth]{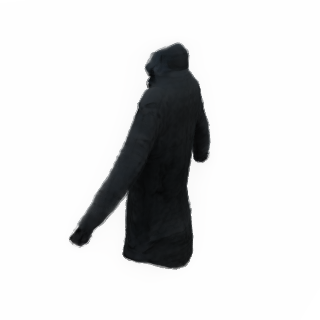} 
        \includegraphics[width=0.07\textwidth]{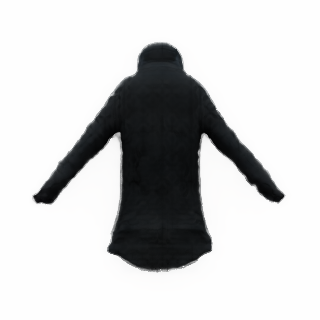}
        \includegraphics[width=0.07\textwidth]{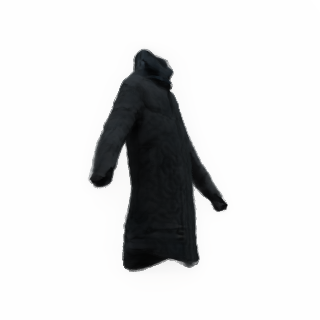}
        \includegraphics[width=0.07\textwidth]{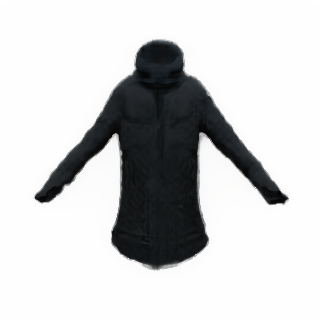} \\
        ``a small copper latern'' & ``a small yellow curved banana'' & ``a modern gray jacket''\\
        
        \includegraphics[width=0.07\textwidth]{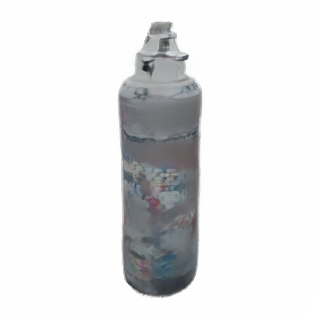} 
        \includegraphics[width=0.07\textwidth]{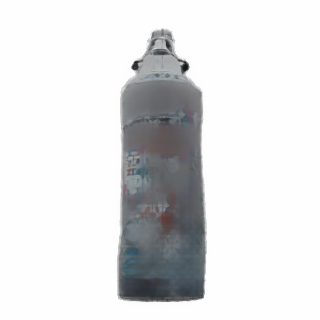}
        \includegraphics[width=0.07\textwidth]{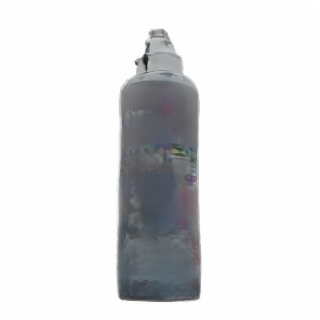}
        \includegraphics[width=0.07\textwidth]{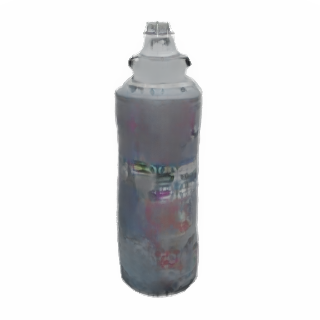} &
        \includegraphics[width=0.07\textwidth]{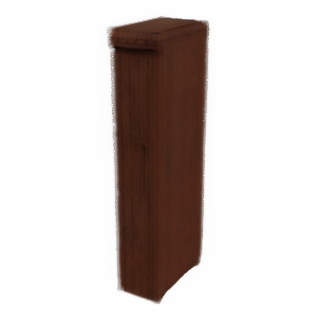} 
        \includegraphics[width=0.07\textwidth]{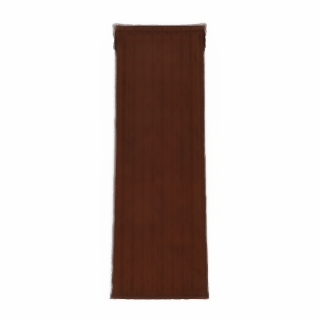} 
        \includegraphics[width=0.07\textwidth]{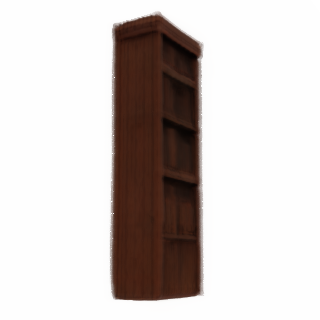} 
        \includegraphics[width=0.07\textwidth]{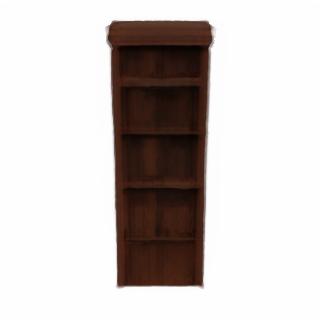}  &
        \includegraphics[width=0.07\textwidth]{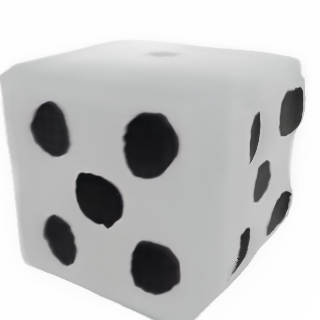} 
        \includegraphics[width=0.07\textwidth]{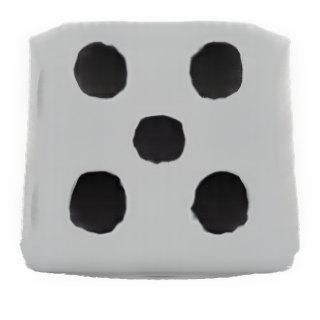}
        \includegraphics[width=0.07\textwidth]{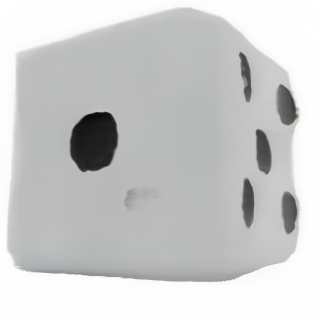}
        \includegraphics[width=0.07\textwidth]{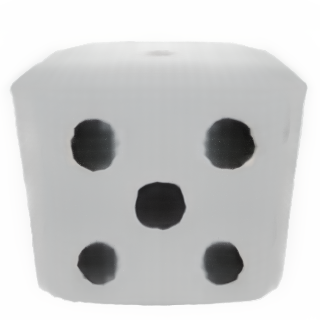} \\
        ``a spray paint can'' & ``a tall wooden bookcase'' & ``a white die with black dots''\\
        
        \includegraphics[width=0.07\textwidth]{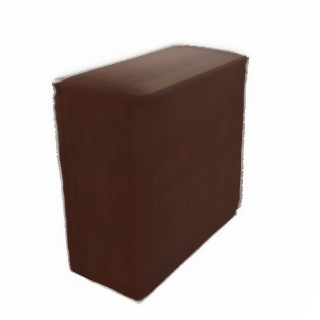} 
        \includegraphics[width=0.07\textwidth]{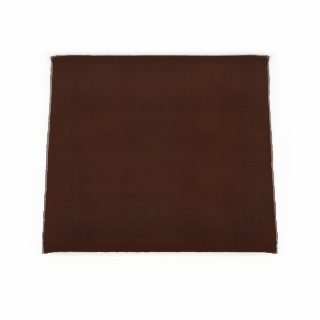}
        \includegraphics[width=0.07\textwidth]{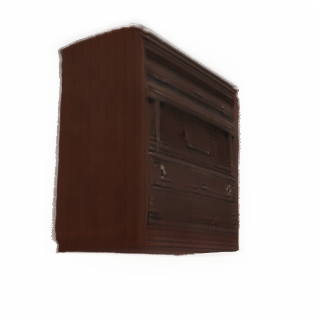}
        \includegraphics[width=0.07\textwidth]{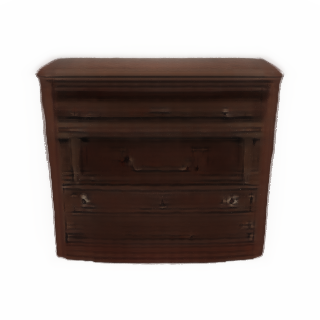} &
        \includegraphics[width=0.07\textwidth]{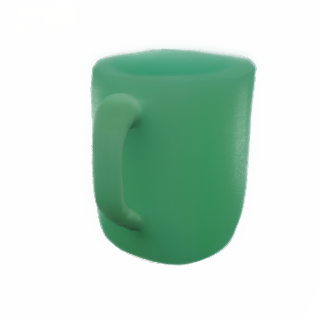} 
        \includegraphics[width=0.07\textwidth]{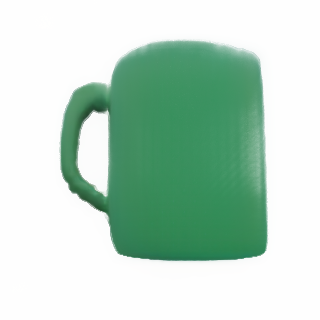} 
        \includegraphics[width=0.07\textwidth]{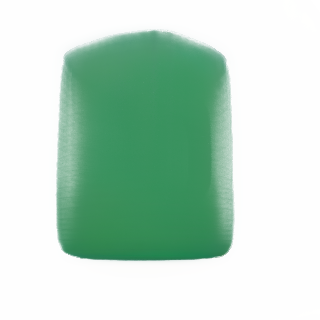} 
        \includegraphics[width=0.07\textwidth]{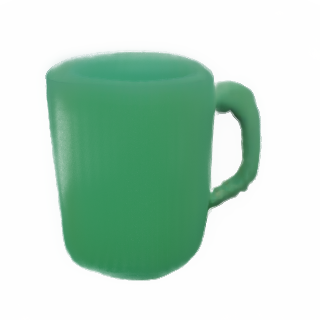}  &
        \includegraphics[width=0.07\textwidth]{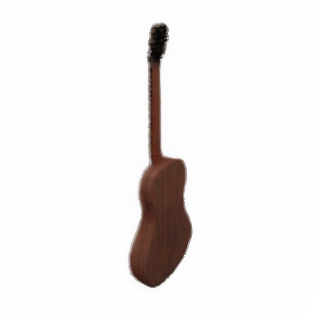} 
        \includegraphics[width=0.07\textwidth]{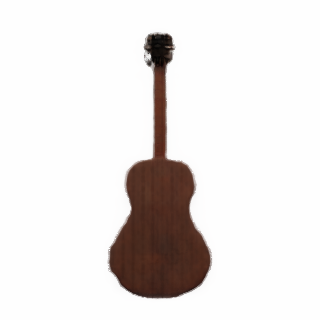}
        \includegraphics[width=0.07\textwidth]{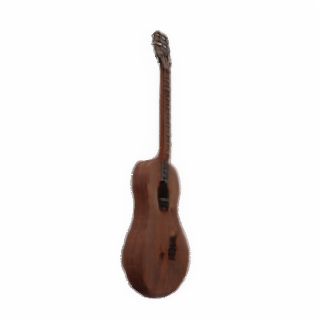}
        \includegraphics[width=0.07\textwidth]{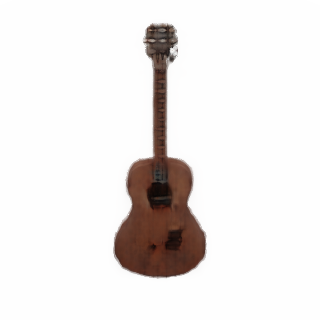} \\
        ``a wooden dresser with drawers and cabinets'' & ``a green mug with handle'' & ``a wooden classic guitar''\\

    \end{tabular}
    \caption{\textbf{Additional text-to-3D results on Objaverse.} We showcase text conditional samples generated by TriFlow. }
    \label{fig:sup_result}
\end{figure*}


%% file: img/toilet/fig.tex
\begin{figure*}[htbp]
    \centering
    \begin{tabular}{c p{2.8cm} p{2.8cm} p{2.8cm} p{2.8cm} p{2.8cm}}
        \raisebox{0.5cm}{\rotatebox{90}{ LN3Diff \cite{lan2025ln3diff}}}  &
        \includegraphics[width=0.18\textwidth,viewport={0 0 512 512},clip]{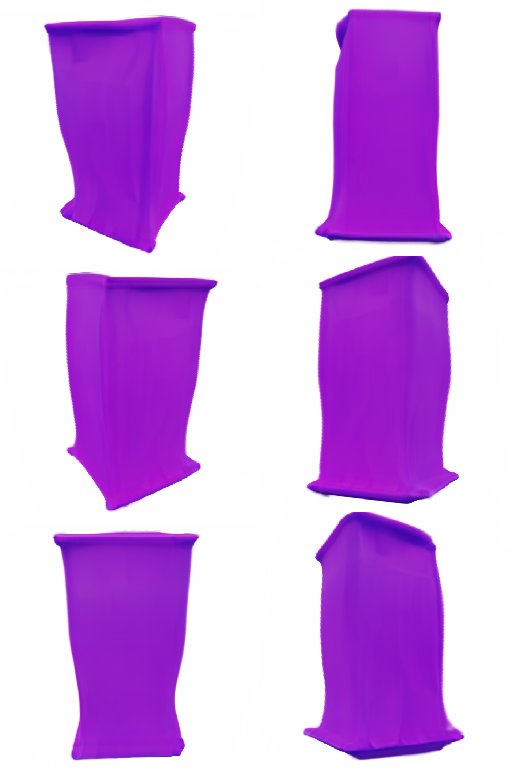}   &
        \includegraphics[width=0.18\textwidth,viewport={0 0 512 512},clip]{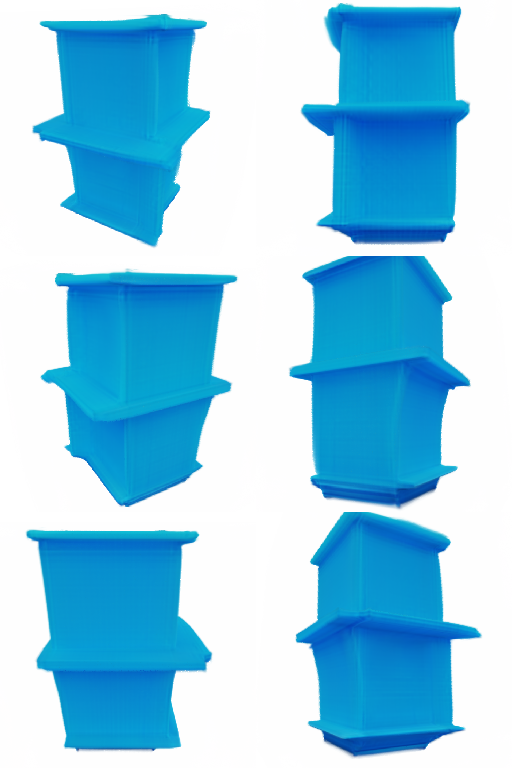}    &
        \includegraphics[width=0.18\textwidth,viewport={0 0 512 512},clip]{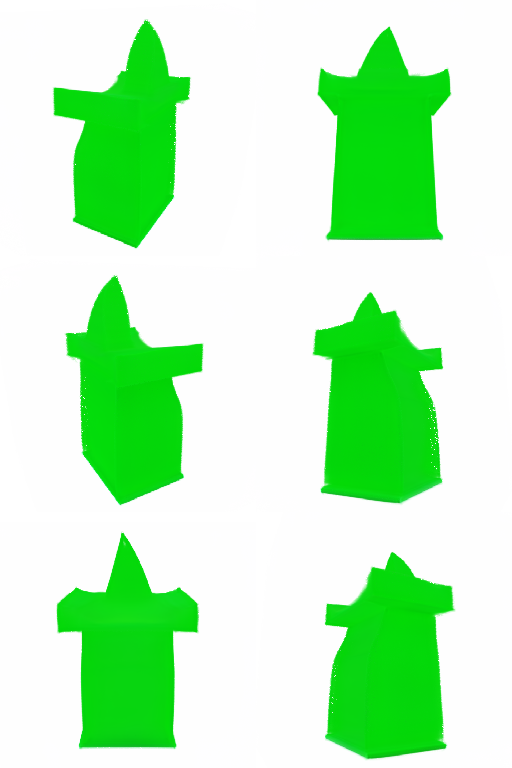}    &
        \includegraphics[width=0.18\textwidth,viewport={0 0 512 512},clip]{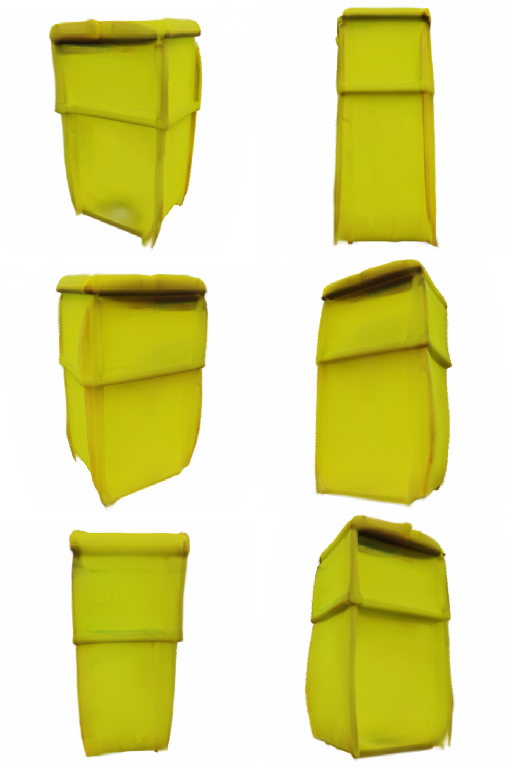}  &
        \includegraphics[width=0.18\textwidth,viewport={0 0 512 512},clip]{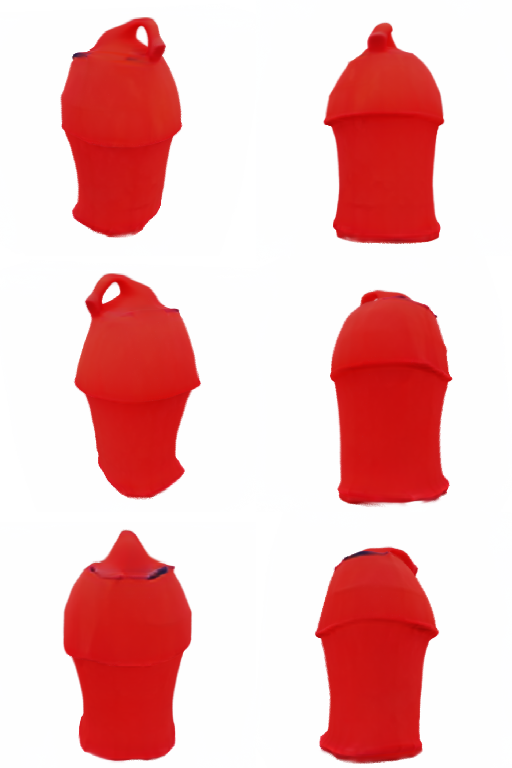}    \\
        
        \raisebox{0.5cm}{\rotatebox{90}{ \shortstack{ Direct3D \cite{liu2024direct} \\ (T=0.002)}}}  &
        \includegraphics[width=0.18\textwidth,viewport={0 0 350 350},clip]{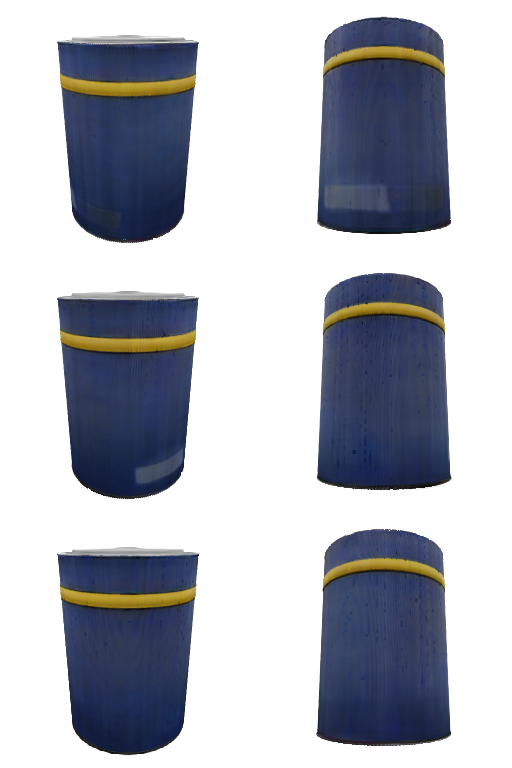}    & 
        \includegraphics[width=0.18\textwidth,viewport={0 0 350 350},clip]{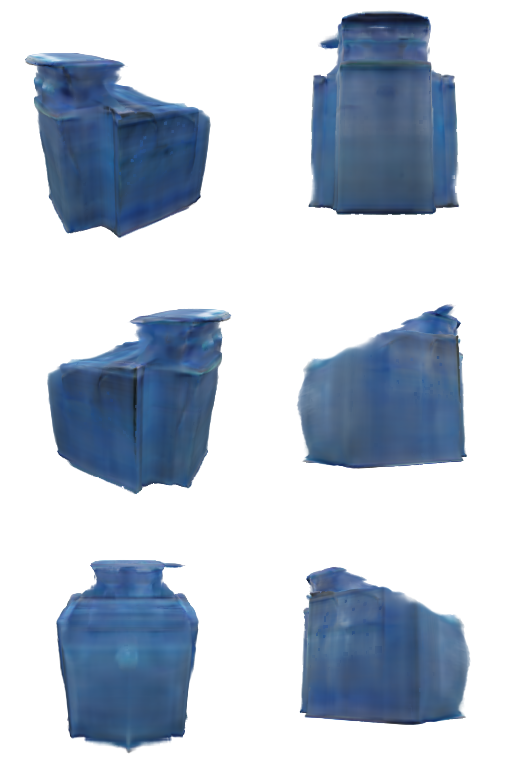}   &
        \includegraphics[width=0.18\textwidth,viewport={0 0 350 350},clip]{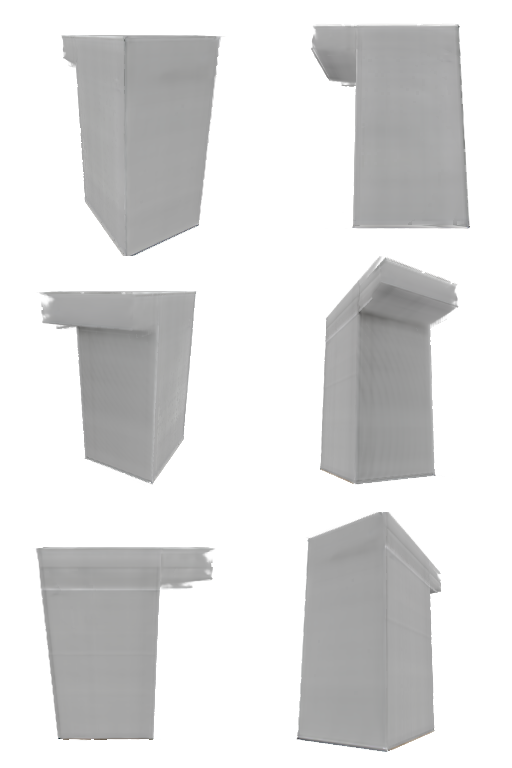}   &
        \includegraphics[width=0.18\textwidth,viewport={0 0 350 350},clip]{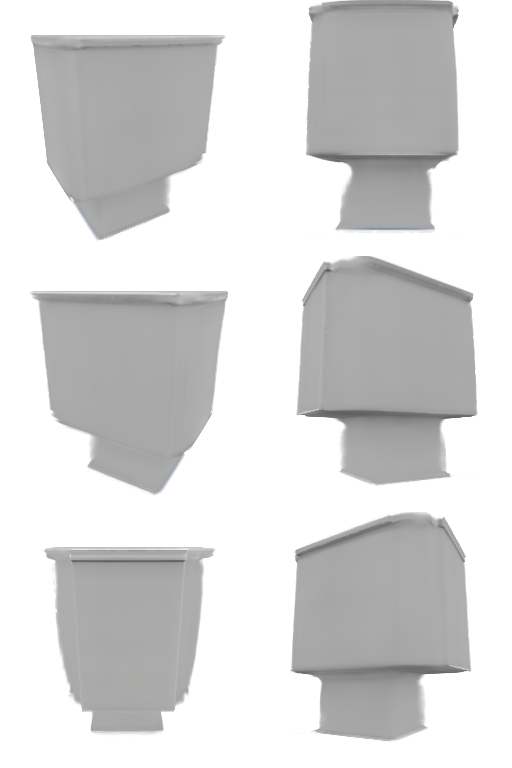}   &
        \includegraphics[width=0.18\textwidth,viewport={0 0 350 350},clip]{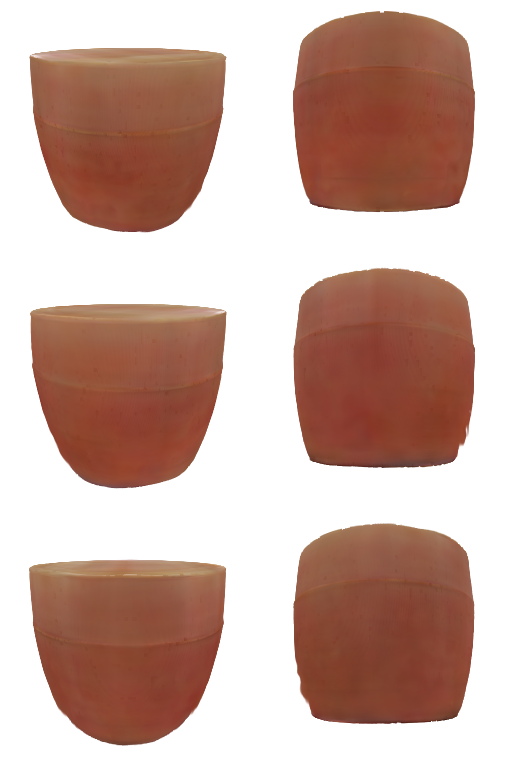}   
            \\
        \raisebox{0.5cm}{\rotatebox{90}{ \shortstack{ Direct3D \cite{liu2024direct}  \\(T=0.07)}}}  &
        \includegraphics[width=0.18\textwidth,viewport={0 0 350 350},clip]{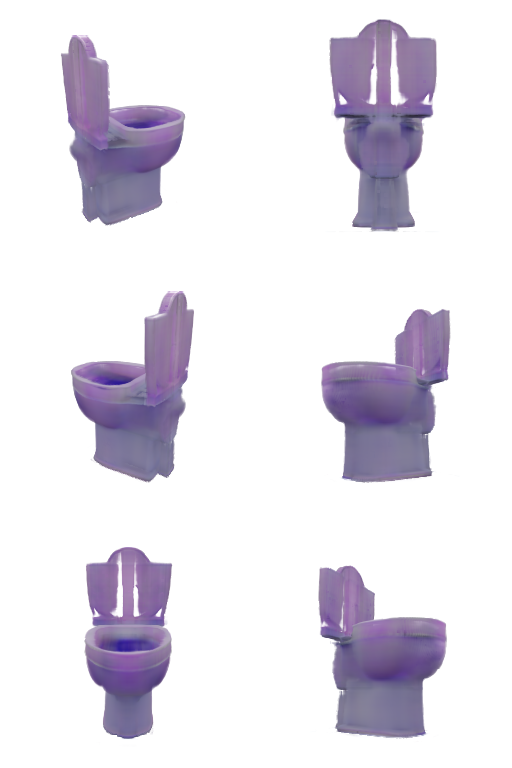}   & 
        \includegraphics[width=0.18\textwidth,viewport={0 0 350 350},clip]{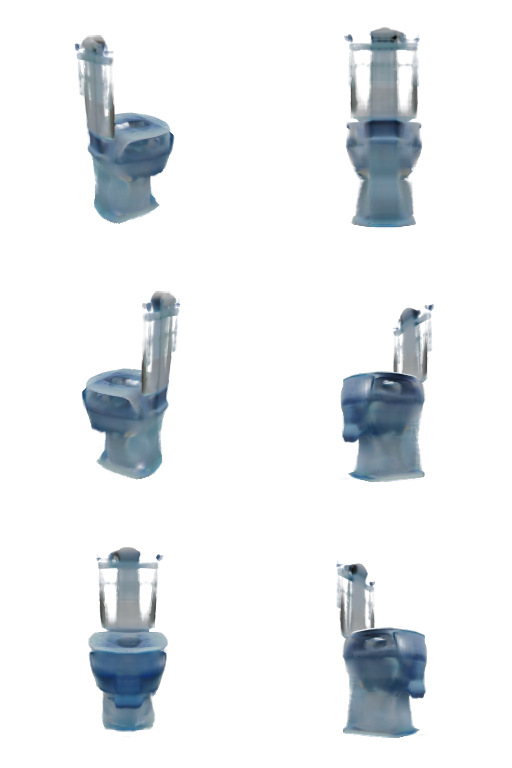}   &
        \includegraphics[width=0.18\textwidth,viewport={0 0 350 350},clip]{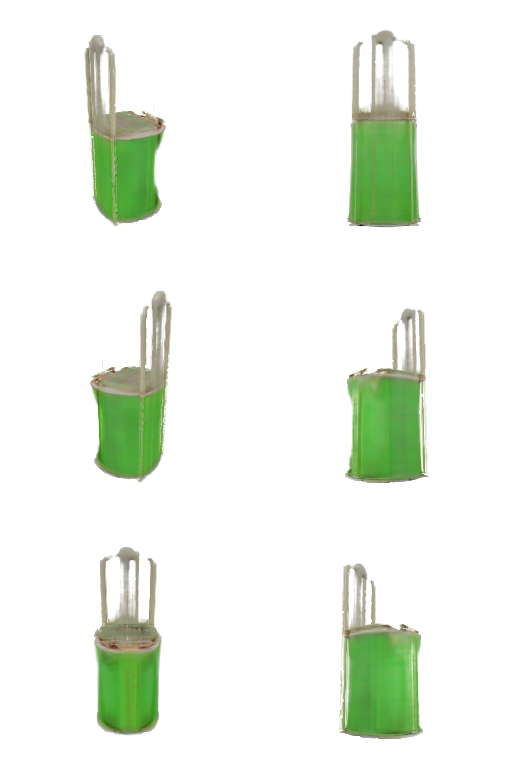}   &
        \includegraphics[width=0.18\textwidth,viewport={0 0 350 350},clip]{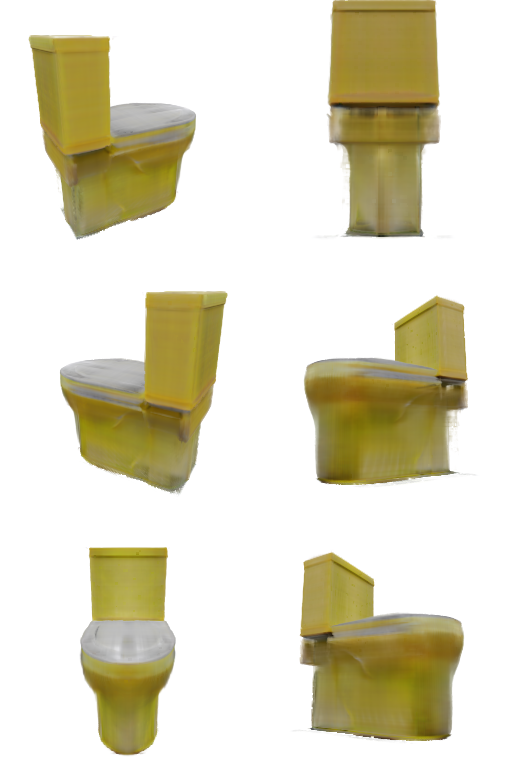}   &
        \includegraphics[width=0.18\textwidth,viewport={0 0 350 350},clip]{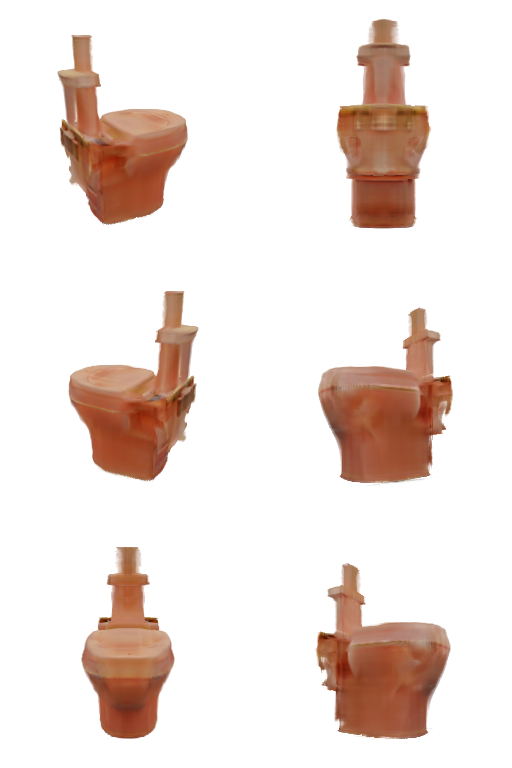}    
            \\

        \raisebox{1.cm}{\rotatebox{90}{ \textbf{Ours}}}  &
        \includegraphics[width=0.18\textwidth,viewport={0 0 512 512},clip]{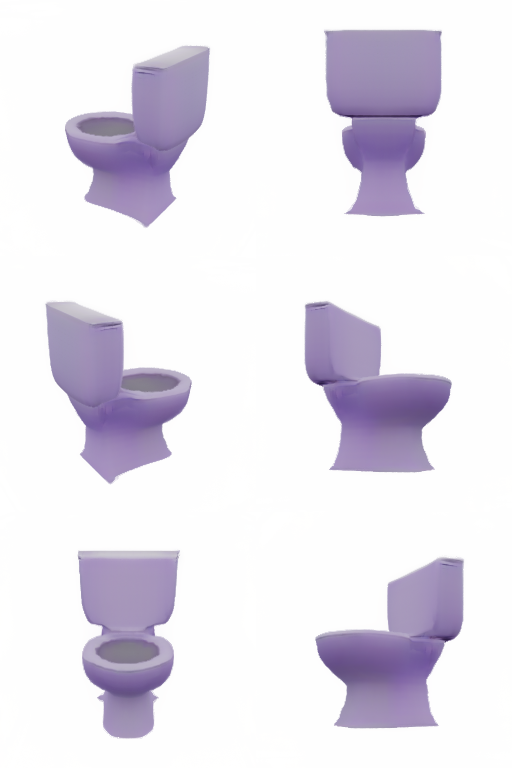}   &
        \includegraphics[width=0.18\textwidth,viewport={0 0 512 512},clip]{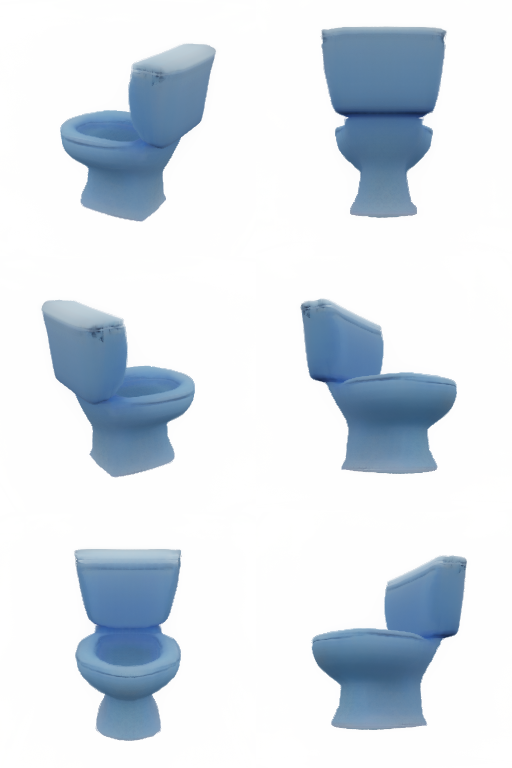}   &
        \includegraphics[width=0.18\textwidth,viewport={0 0 512 512},clip]{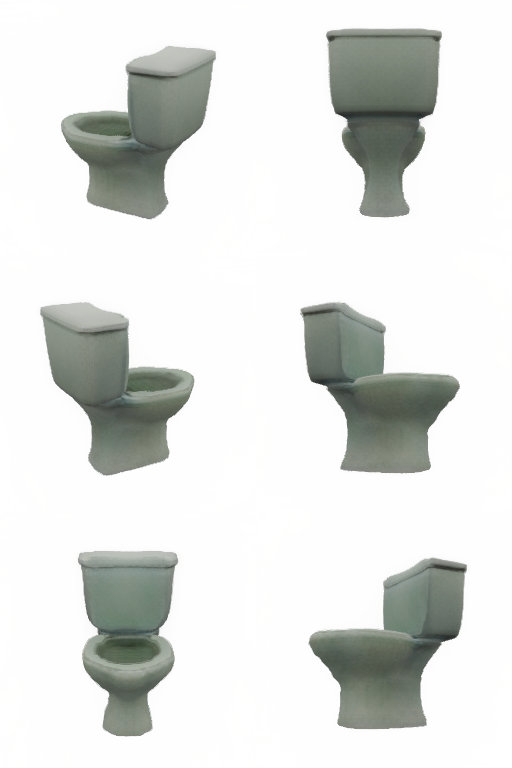}   &
        \includegraphics[width=0.18\textwidth,viewport={0 0 512 512},clip]{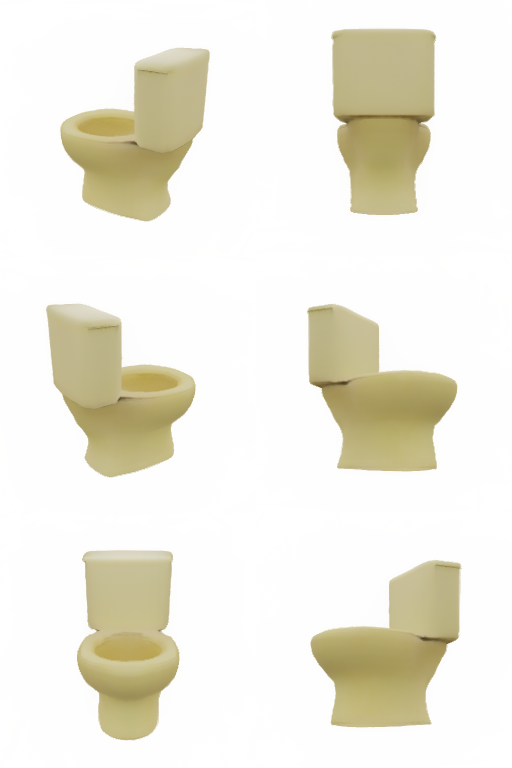}    &
        \includegraphics[width=0.18\textwidth,viewport={0 0 512 512},clip]{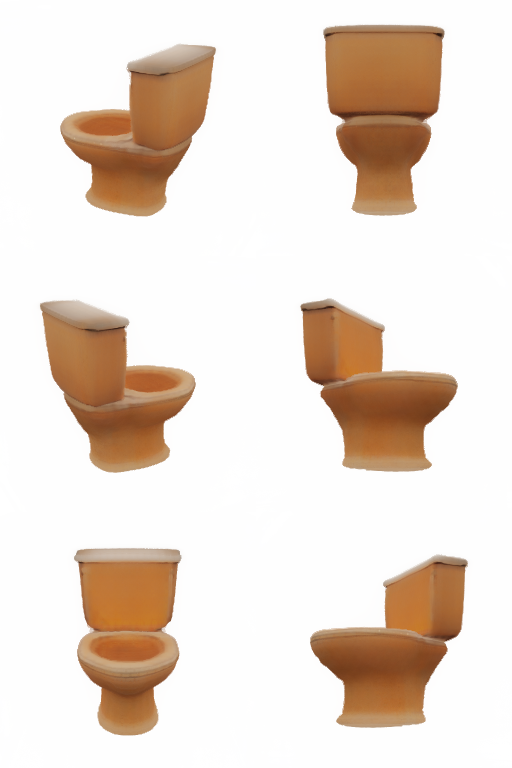}   \\

         & 
        \quad``a purple toilet'' & 
        \quad``a blue toilet'' & 
        \quad``a green toilet'' & 
        \quad``a yellow toilet''& 
        \quad``a orange toilet'' \\
    \end{tabular}
    \caption{\textbf{Varying colors with text prompts}. We asked our and baseline models to generate toilets in various unusual colors, including purple, blue, green, yellow, and orange. These colors do not appear in any of the existing toilet assets in the dataset. While the baseline methods of LN3Diff and Direct3D struggle to generate high-quality toilets respecting the color prompts, our model consistently succeeds. This phenomenon demonstrates our model's ability to generalize and understand prompts on color variations beyond the typical range.}
    \label{fig:toilet}
\end{figure*}

%% file: img/uncond_figs/uncond_figs.tex
\begin{table*}[htbp]
    \centering
    \begin{tabular}{ p{6cm} p{6cm} p{6cm}}
         \includegraphics[width=0.08\textwidth]{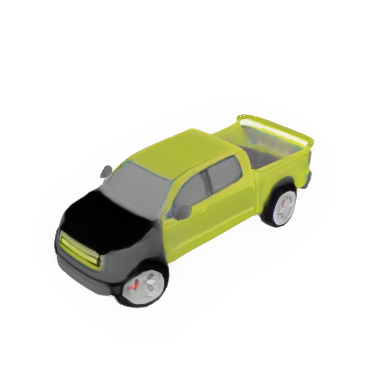} 
        \includegraphics[width=0.08\textwidth]{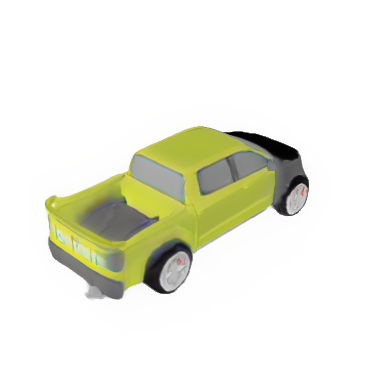}
        \includegraphics[width=0.08\textwidth]{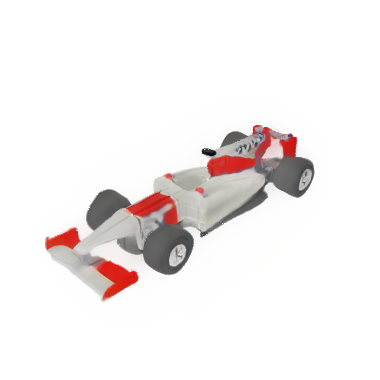}
        \includegraphics[width=0.08\textwidth]{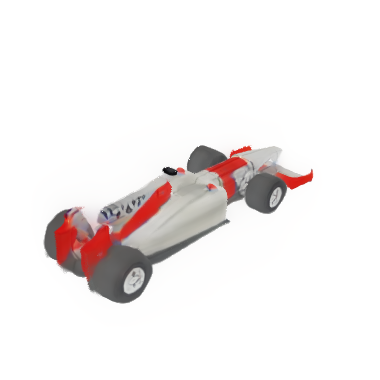} &
        \includegraphics[width=0.08\textwidth]{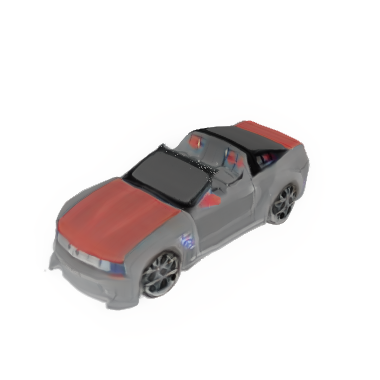} 
        \includegraphics[width=0.08\textwidth]{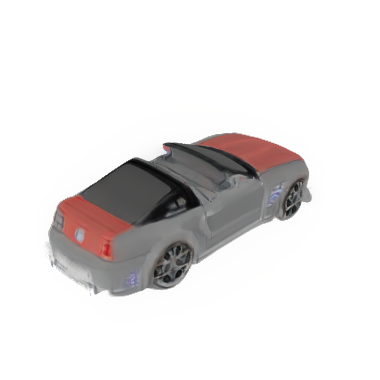}
        \includegraphics[width=0.08\textwidth]{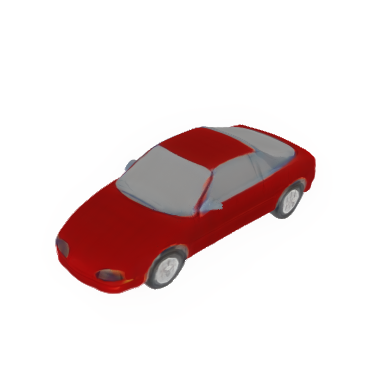}
        \includegraphics[width=0.08\textwidth]{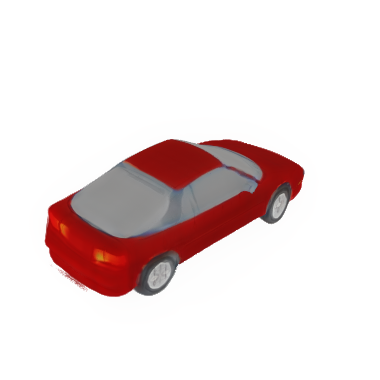}  &
        \includegraphics[width=0.08\textwidth]{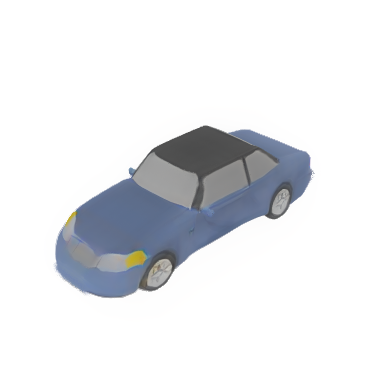} 
        \includegraphics[width=0.08\textwidth]{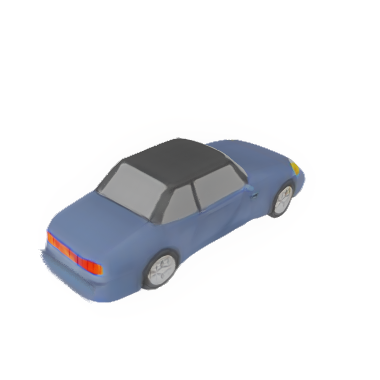}
        \includegraphics[width=0.08\textwidth]{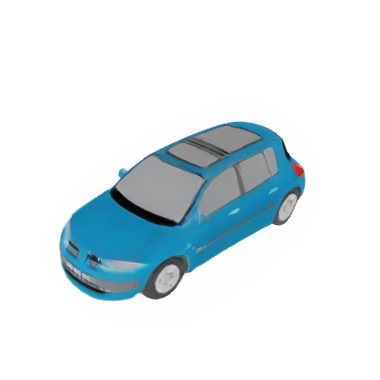}
        \includegraphics[width=0.08\textwidth]{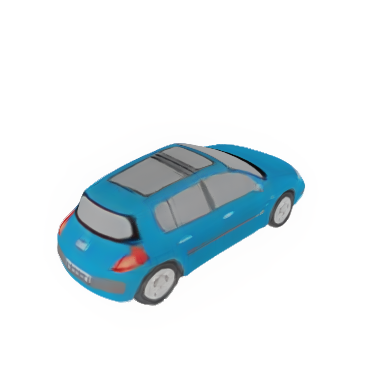} \\
        
        \includegraphics[width=0.08\textwidth]{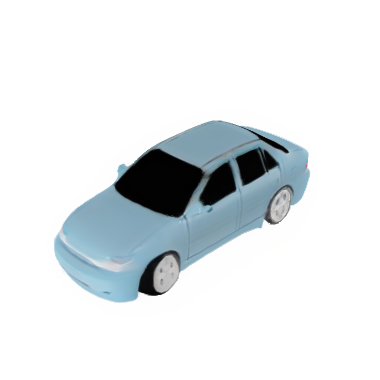} 
        \includegraphics[width=0.08\textwidth]{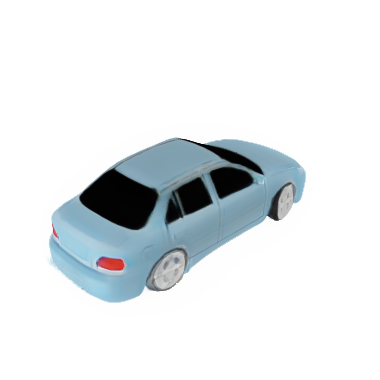}
        \includegraphics[width=0.08\textwidth]{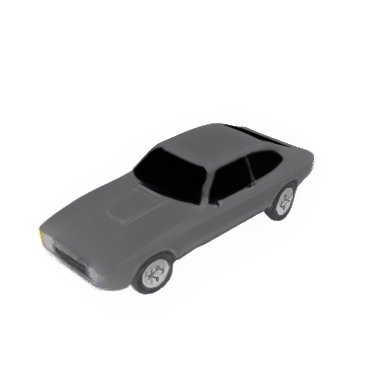}
        \includegraphics[width=0.08\textwidth]{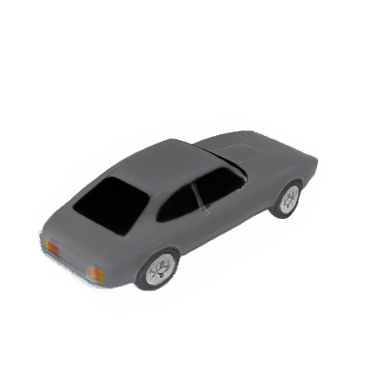} &
        \includegraphics[width=0.08\textwidth]{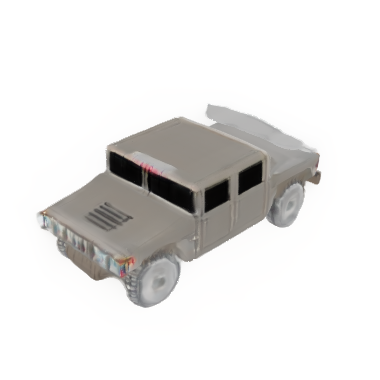} 
        \includegraphics[width=0.08\textwidth]{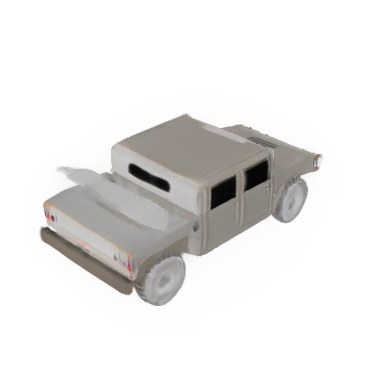}
        \includegraphics[width=0.08\textwidth]{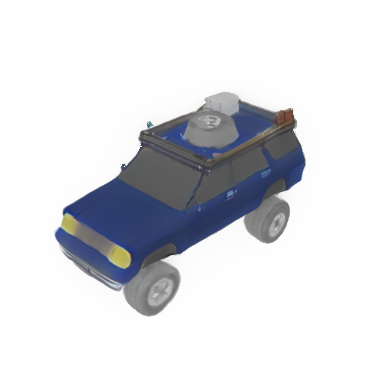}
        \includegraphics[width=0.08\textwidth]{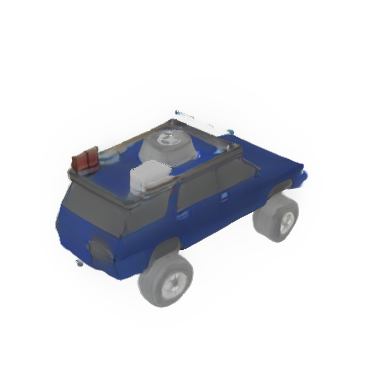}  &
        \includegraphics[width=0.08\textwidth]{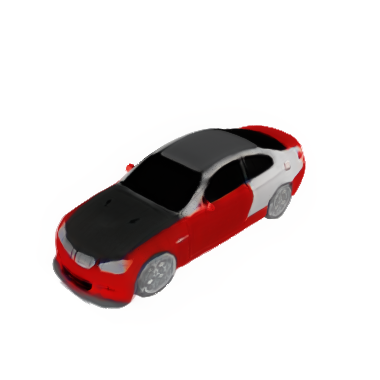} 
        \includegraphics[width=0.08\textwidth]{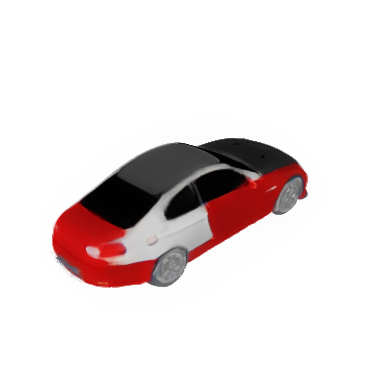}
        \includegraphics[width=0.08\textwidth]{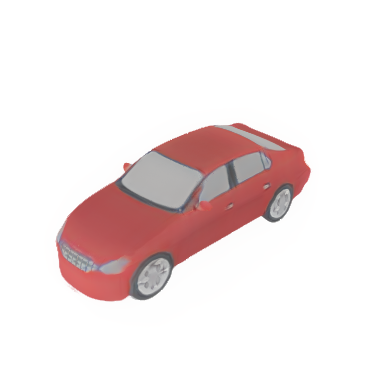}
        \includegraphics[width=0.08\textwidth]{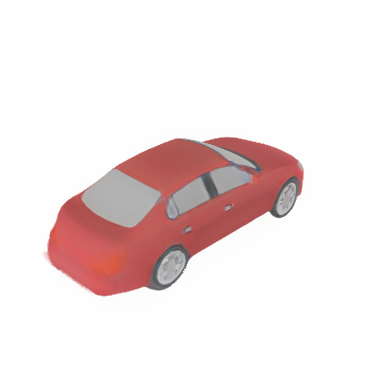} \\
        
        \includegraphics[width=0.08\textwidth]{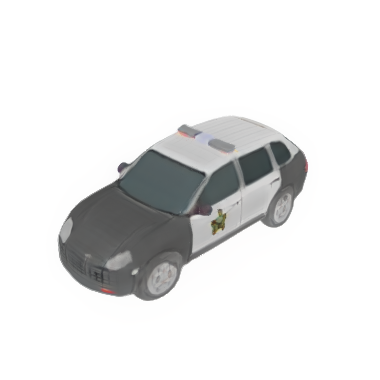} 
        \includegraphics[width=0.08\textwidth]{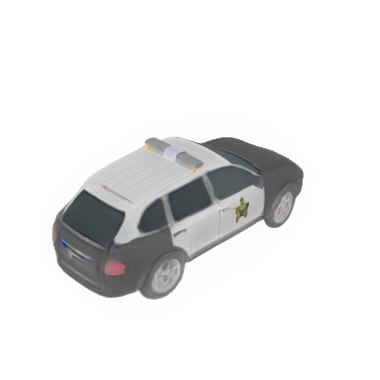}
        \includegraphics[width=0.08\textwidth]{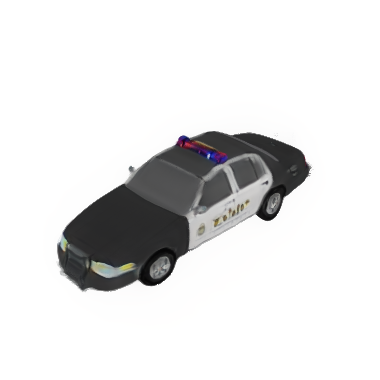}
        \includegraphics[width=0.08\textwidth]{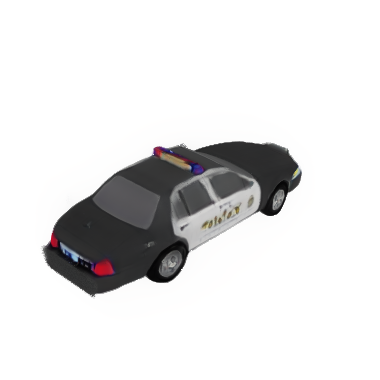}  &
        \includegraphics[width=0.08\textwidth]{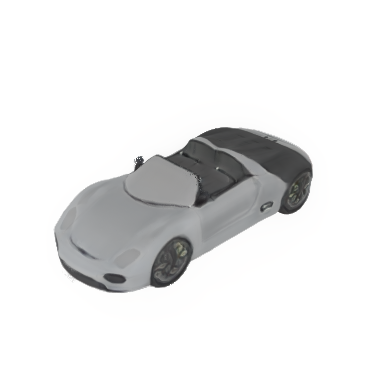} 
        \includegraphics[width=0.08\textwidth]{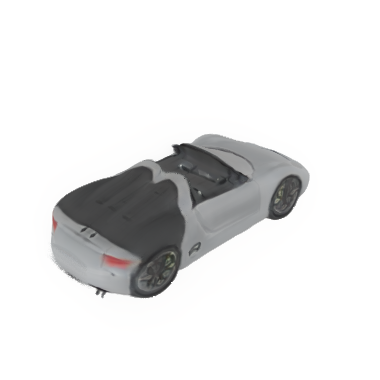}
        \includegraphics[width=0.08\textwidth]{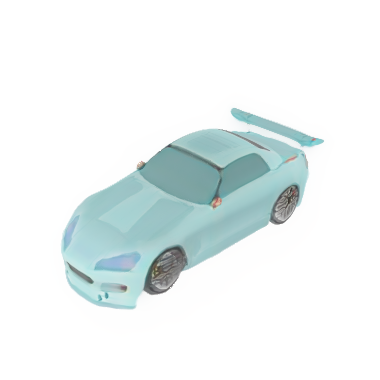}
        \includegraphics[width=0.08\textwidth]{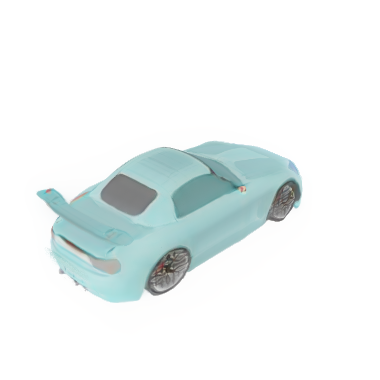}   &
        \includegraphics[width=0.08\textwidth]{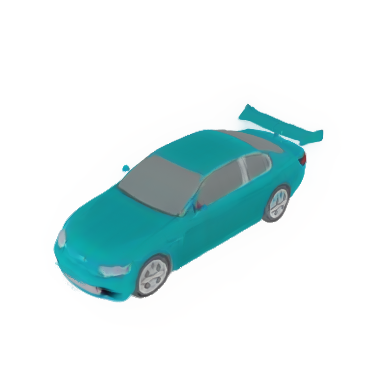} 
        \includegraphics[width=0.08\textwidth]{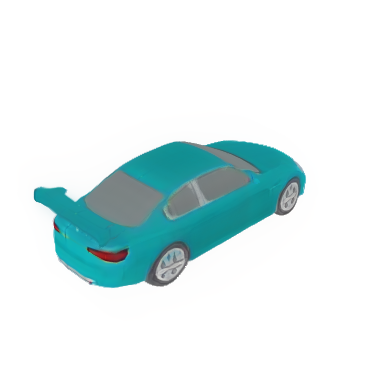}
        \includegraphics[width=0.08\textwidth]{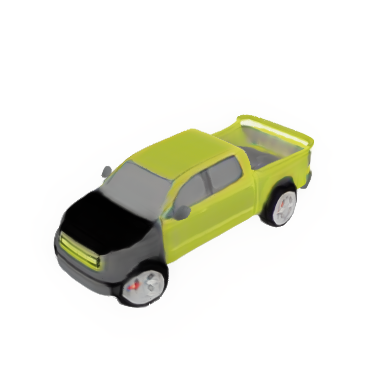}
        \includegraphics[width=0.08\textwidth]{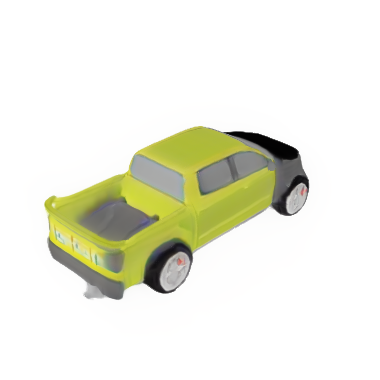}  \\
        
        \includegraphics[width=0.08\textwidth]{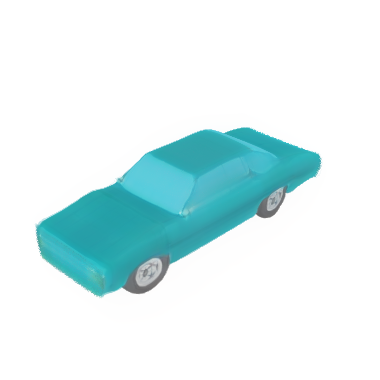} 
        \includegraphics[width=0.08\textwidth]{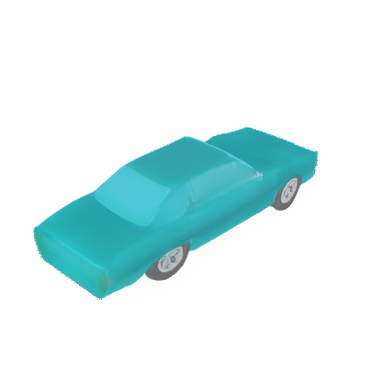}
        \includegraphics[width=0.08\textwidth]{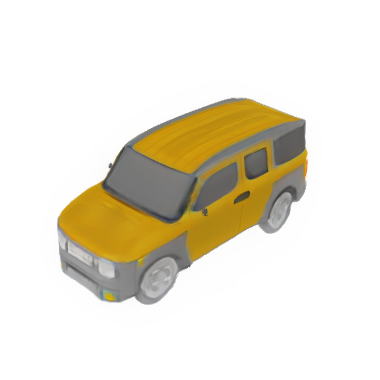}
        \includegraphics[width=0.08\textwidth]{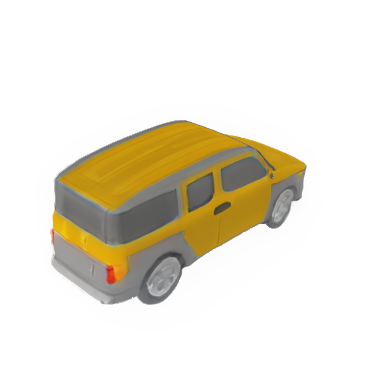}  &
        \includegraphics[width=0.08\textwidth]{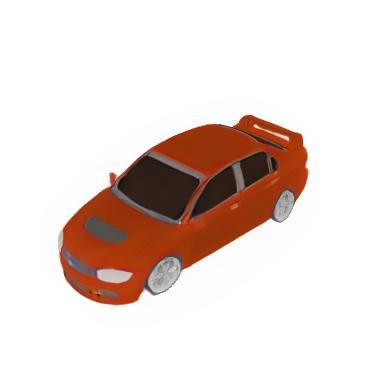} 
        \includegraphics[width=0.08\textwidth]{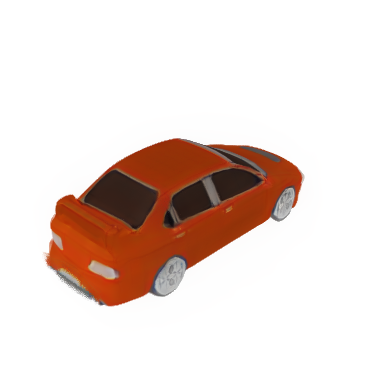}
        \includegraphics[width=0.08\textwidth]{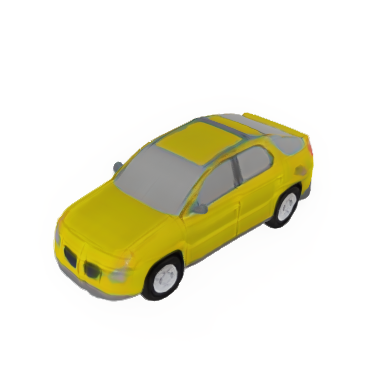}
        \includegraphics[width=0.08\textwidth]{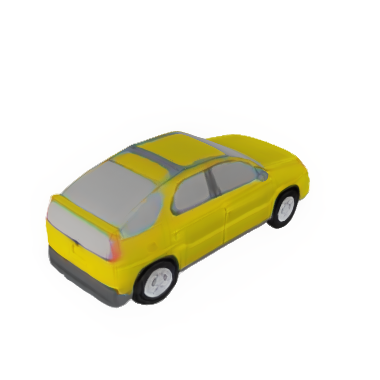}   &
        \includegraphics[width=0.08\textwidth]{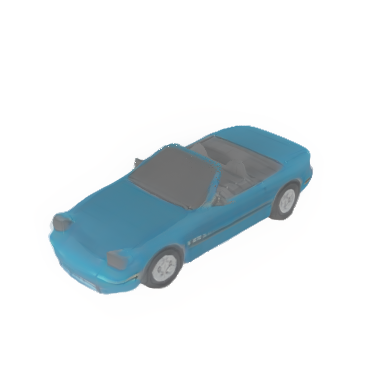} 
        \includegraphics[width=0.08\textwidth]{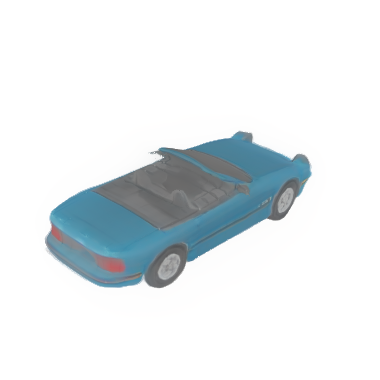}
        \includegraphics[width=0.08\textwidth]{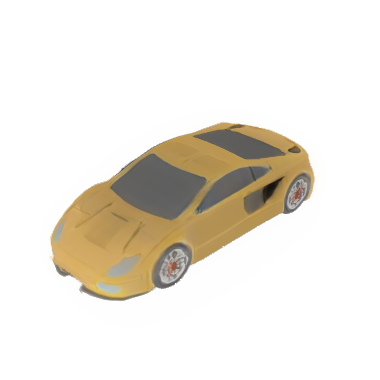}
        \includegraphics[width=0.08\textwidth]{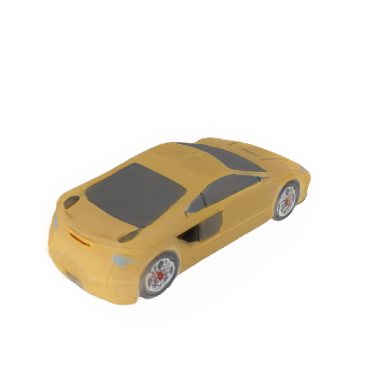} \\

         & \centering (a) ShapeNet Car & \\
        
        \includegraphics[width=0.08\textwidth]{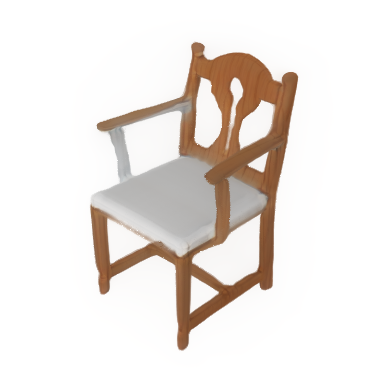} 
        \includegraphics[width=0.08\textwidth]{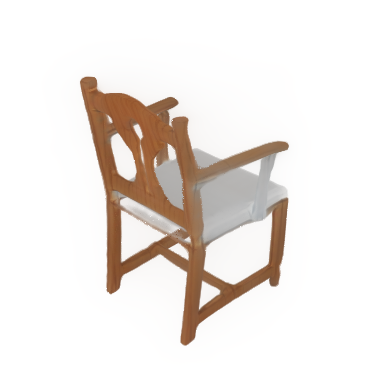} 
        \includegraphics[width=0.08\textwidth]{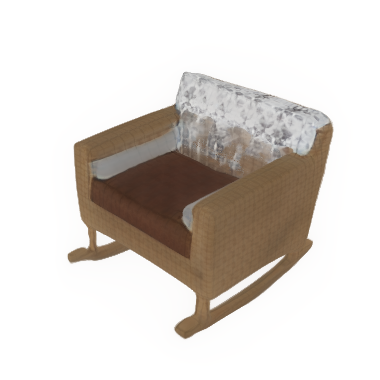} 
        \includegraphics[width=0.08\textwidth]{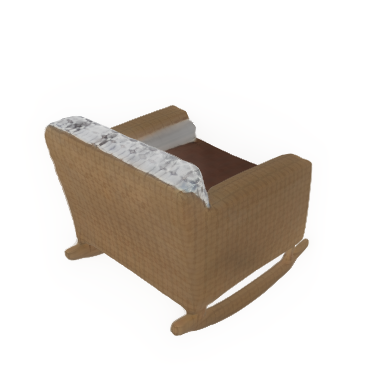}   &
        \includegraphics[width=0.08\textwidth]{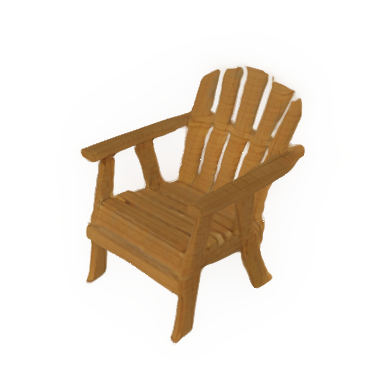} 
        \includegraphics[width=0.08\textwidth]{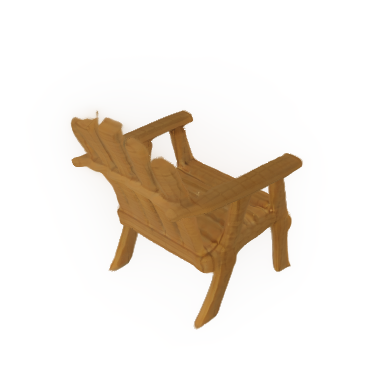} 
        \includegraphics[width=0.08\textwidth]{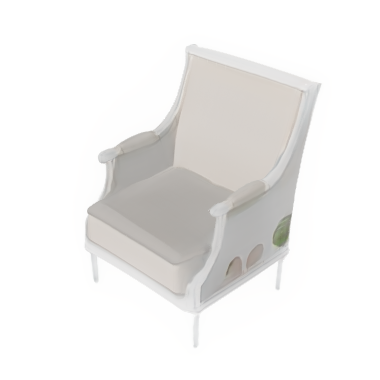} 
        \includegraphics[width=0.08\textwidth]{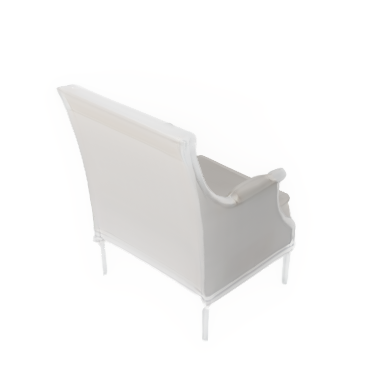}    &
        \includegraphics[width=0.08\textwidth]{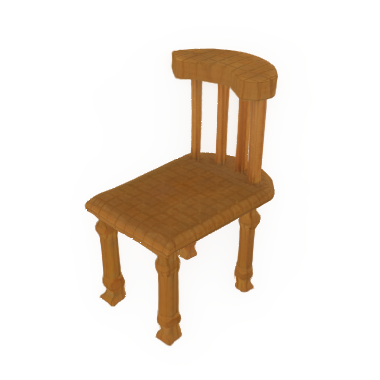} 
        \includegraphics[width=0.08\textwidth]{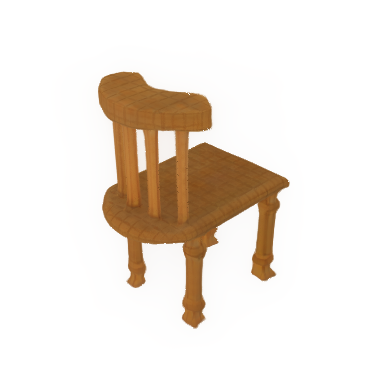} 
        \includegraphics[width=0.08\textwidth]{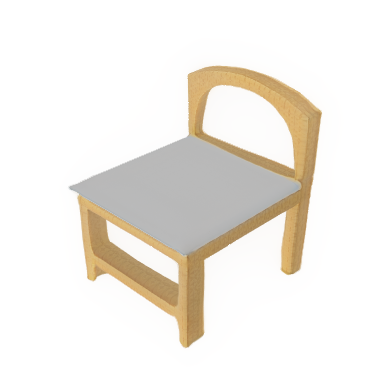} 
        \includegraphics[width=0.08\textwidth]{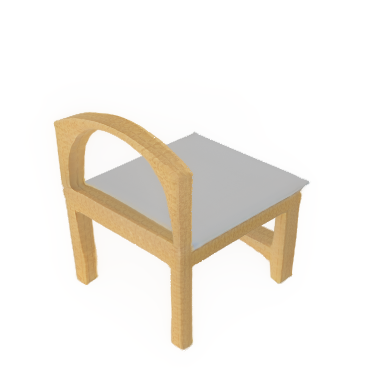} \\

        \includegraphics[width=0.08\textwidth]{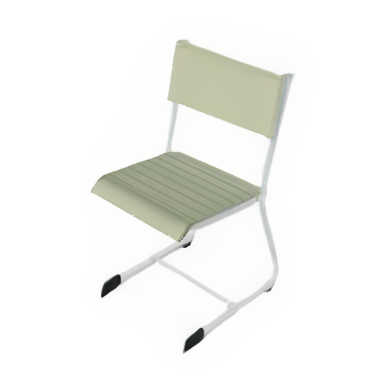} 
        \includegraphics[width=0.08\textwidth]{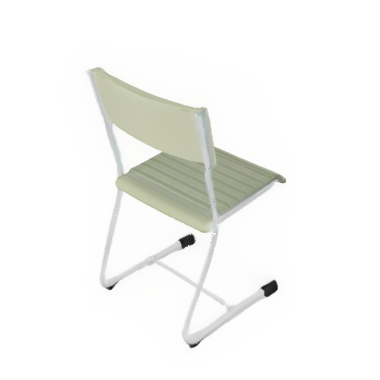} 
        \includegraphics[width=0.08\textwidth]{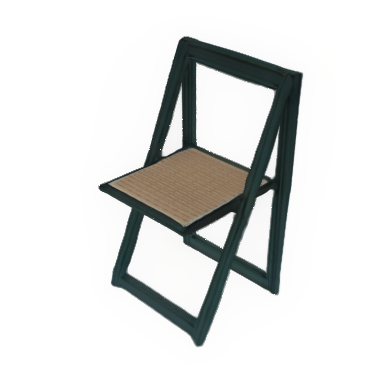} 
        \includegraphics[width=0.08\textwidth]{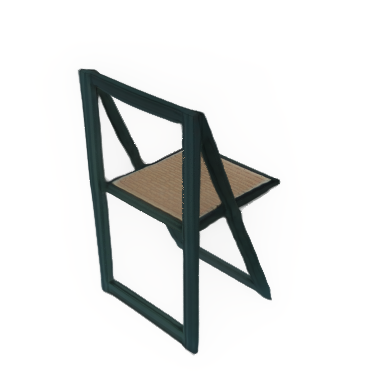}   &
        \includegraphics[width=0.08\textwidth]{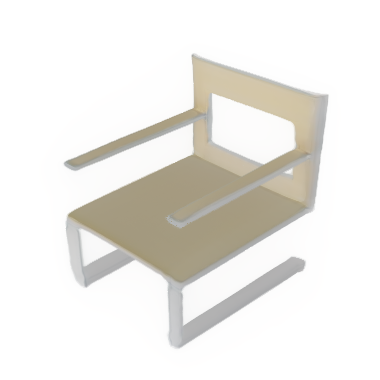} 
        \includegraphics[width=0.08\textwidth]{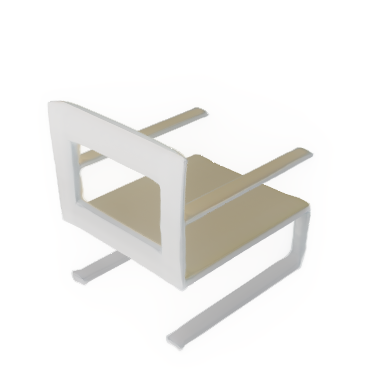} 
        \includegraphics[width=0.08\textwidth]{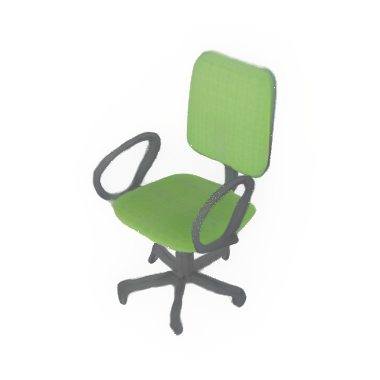} 
        \includegraphics[width=0.08\textwidth]{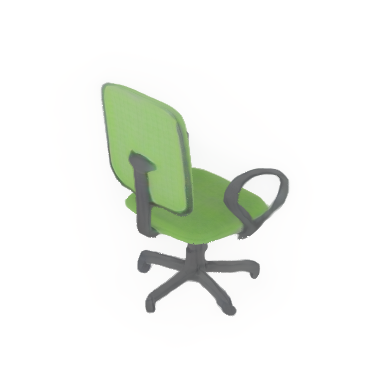}    &
        \includegraphics[width=0.08\textwidth]{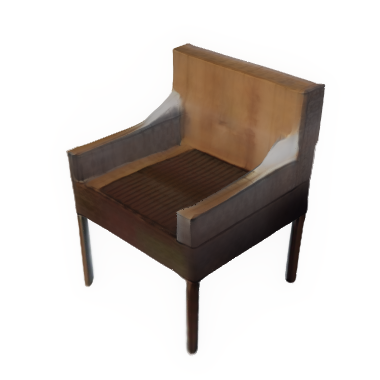} 
        \includegraphics[width=0.08\textwidth]{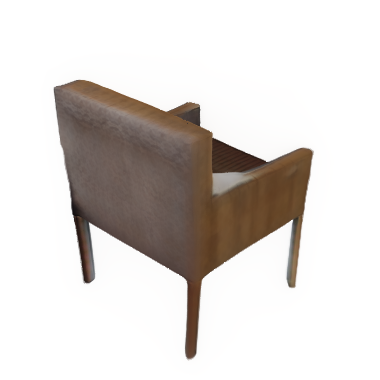} 
        \includegraphics[width=0.08\textwidth]{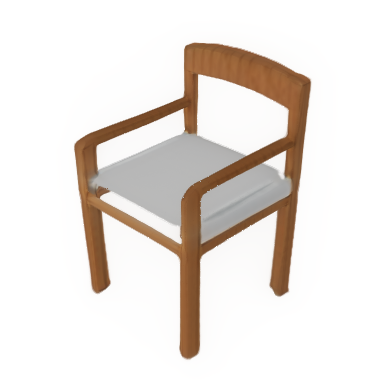} 
        \includegraphics[width=0.08\textwidth]{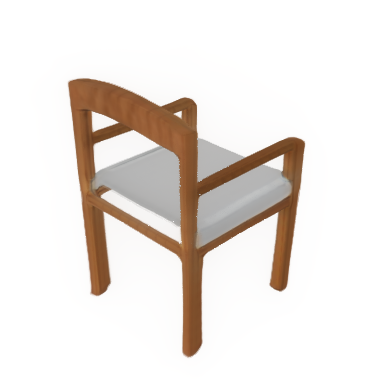} \\

        \includegraphics[width=0.08\textwidth]{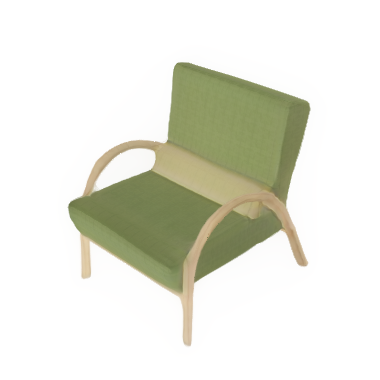} 
        \includegraphics[width=0.08\textwidth]{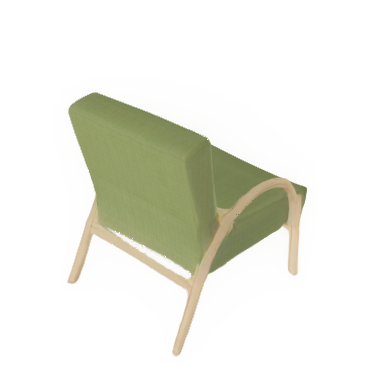} 
        \includegraphics[width=0.08\textwidth]{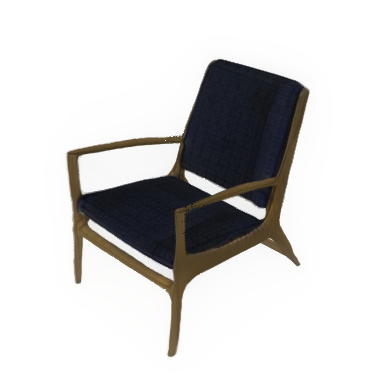} 
        \includegraphics[width=0.08\textwidth]{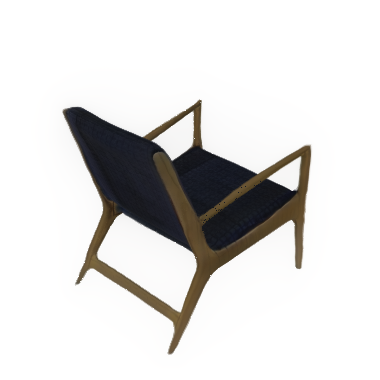}   &
        \includegraphics[width=0.08\textwidth]{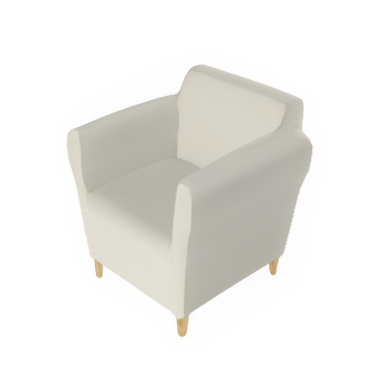} 
        \includegraphics[width=0.08\textwidth]{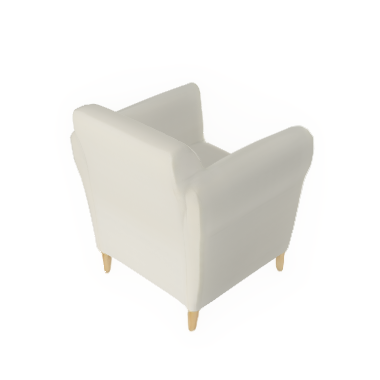} 
        \includegraphics[width=0.08\textwidth]{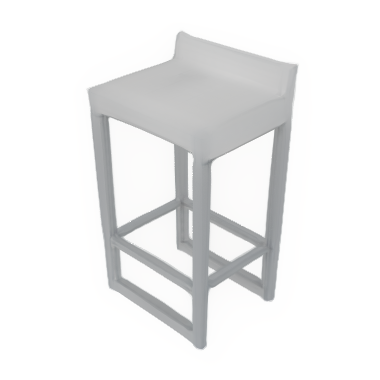} 
        \includegraphics[width=0.08\textwidth]{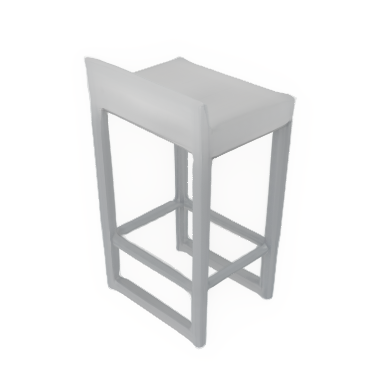}    &
        \includegraphics[width=0.08\textwidth]{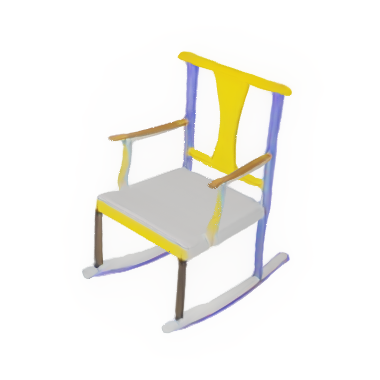} 
        \includegraphics[width=0.08\textwidth]{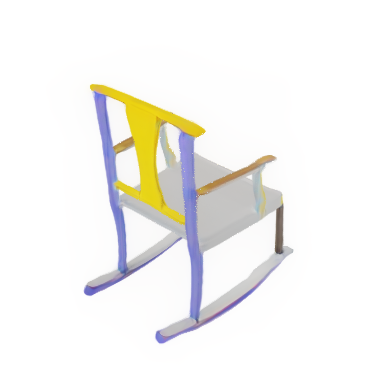} 
        \includegraphics[width=0.08\textwidth]{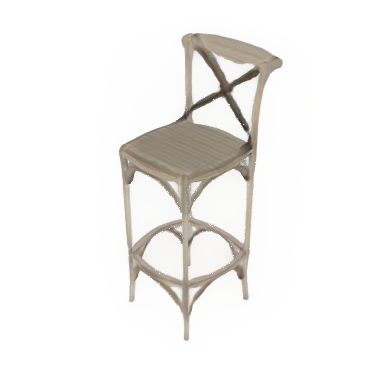} 
        \includegraphics[width=0.08\textwidth]{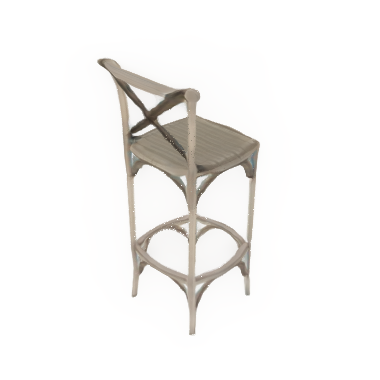} \\

        \includegraphics[width=0.08\textwidth]{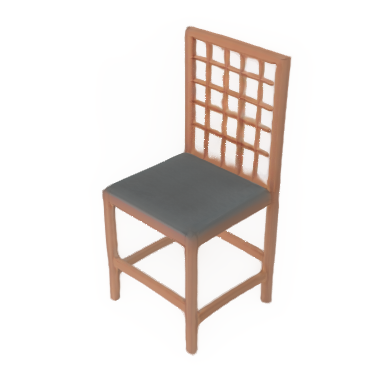} 
        \includegraphics[width=0.08\textwidth]{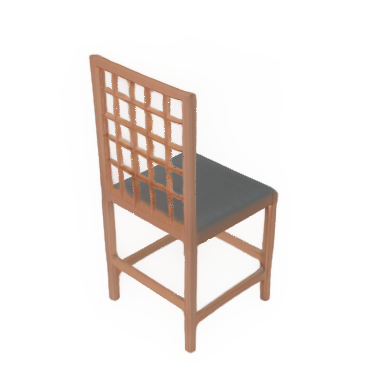} 
        \includegraphics[width=0.08\textwidth]{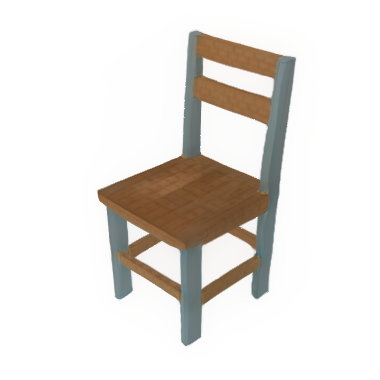} 
        \includegraphics[width=0.08\textwidth]{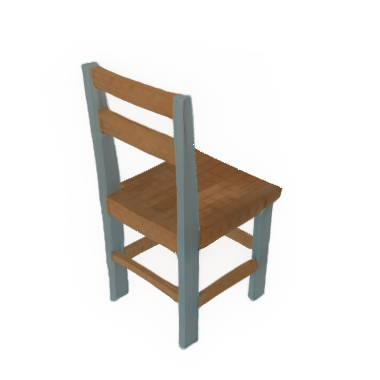}   &
        \includegraphics[width=0.08\textwidth]{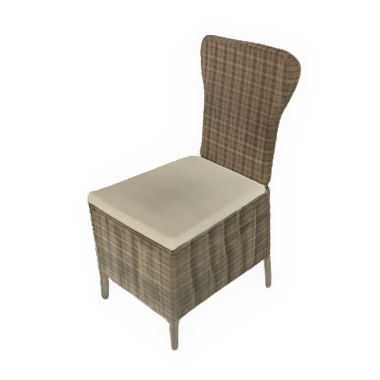} 
        \includegraphics[width=0.08\textwidth]{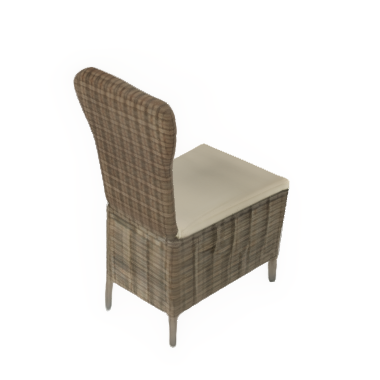} 
        \includegraphics[width=0.08\textwidth]{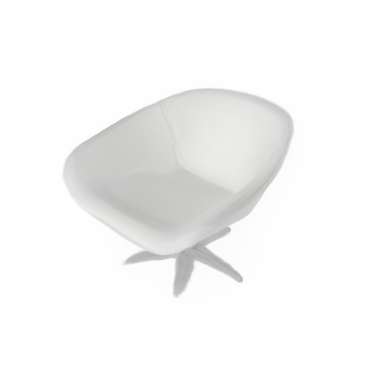} 
        \includegraphics[width=0.08\textwidth]{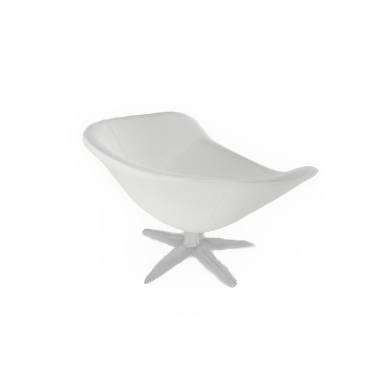}    &
        \includegraphics[width=0.08\textwidth]{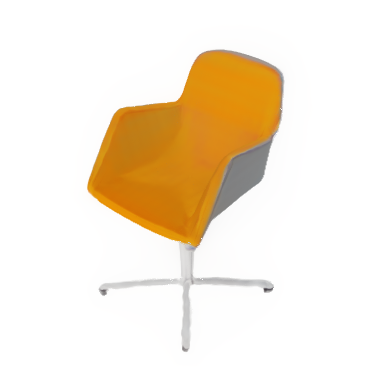} 
        \includegraphics[width=0.08\textwidth]{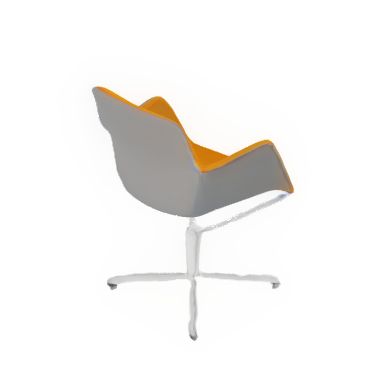} 
        \includegraphics[width=0.08\textwidth]{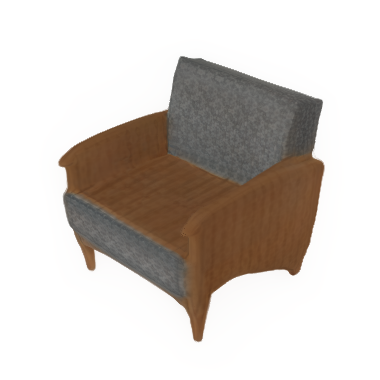} 
        \includegraphics[width=0.08\textwidth]{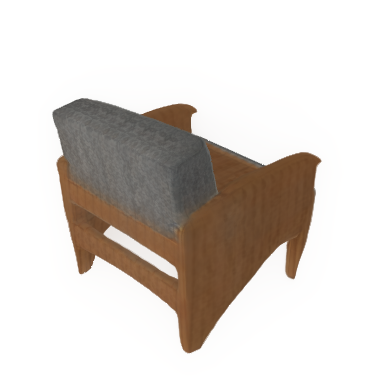}  \\
         & \centering (b) ShapeNet Chair &


    \end{tabular}
    \caption{\textbf{Uncondtional ShapeNet generation examples.} We showcase unconditional samples generated by TriFlow trained on ShapeNet Car and Chair, respectively. Better viewed digitally.}
    \label{fig:uncond_figs}
\end{table*}